\documentclass[twocolumn]{svjour3}          %
\smartqed  %
\usepackage{graphicx}

\usepackage{amssymb}
\usepackage{amsmath}
\usepackage{multirow}
\usepackage{pdfpages}
\usepackage{microtype}

\usepackage[backref=page]{hyperref}

\definecolor{ForestGreen}{RGB}{27,132,27}

\usepackage{xcolor} %
\hypersetup{
	colorlinks,
	linkcolor={red!50!black},
	citecolor={blue!50!black},
	urlcolor={blue!80!black}
}

\usepackage{natbib}
\usepackage{grffile}
\usepackage{comment}

\usepackage[letterpaper, left=0.83in, right=0.83in, top=0.81in, bottom=0.71in]{geometry}

\usepackage{tikz}
\usetikzlibrary{positioning,calc}
\usetikzlibrary{decorations,decorations.markings}
\usetikzlibrary{fit}
\usetikzlibrary{shapes,arrows,shadows}
\usetikzlibrary{decorations.pathreplacing}
\usetikzlibrary{arrows.meta}

\definecolor{mygreen}{RGB}{20, 99, 4}

\newcommand{\PPP}{{\hspace{-.049em}\raisebox{.454ex}{\tiny\bf ++}}}
\newcommand{\PP}{{\texttt{++}}}

\renewcommand{\hat}{\widehat}
\renewcommand{\tilde}{\widetilde}
\newcommand{\new}[1]{#1}
\newcommand{\newer}[1]{#1}

\authorrunning{Tristan Aumentado-Armstrong et al.}

\begin{document}

\title{%
Disentangling Geometric Deformation Spaces in Generative Latent Shape Models
}

\author{Tristan Aumentado-Armstrong$^{1,2,3}$ \and 
        Stavros Tsogkas$^{1,2}$ 		\and
        Sven Dickinson$^{1,2,3}$        \and 
        Allan Jepson$^{1,2}$%
}

\institute{
        Tristan Aumentado-Armstrong$^{1,2,3}$ \at
        \email{taumen@cs.utoronto.ca}
        \and
        Stavros Tsogkas$^{1,2}$ \at
        \email{tsogkas@cs.toronto.edu}
        \and
        Sven Dickinson$^{1,2,3}$ \at
        \email{sven@cs.toronto.edu}
        \and
        Allan Jepson$^{1,2}$ \at
        \email{jepson@cs.toronto.edu}
        \and
        $^1$ University of Toronto \\
        $^2$ Samsung Toronto AI Research Center \\
        $^3$ Vector Institute for Artificial Intelligence\\
        \\
        Disclaimer: Tristan Aumentado-Armstrong and Stavros Tsogkas
        contributed to this article in their personal capacity as PhD student and 
        Adjunct Professor at the University of Toronto, respectively.
        Sven Dickinson and Allan Jepson contributed to 
        this article in their personal capacity as Professors at the University of Toronto. 
        The views expressed (or the conclusions reached) by the authors are their own and do 
        not necessarily represent the views of Samsung Research America, Inc. \\
        \\
        \today
}

\maketitle

\begin{abstract}
A complete representation of 3D objects requires characterizing the space of deformations in an interpretable manner, 
	from articulations of a single instance to changes in shape across categories. 
In this work, we improve on a prior generative model of geometric disentanglement for 3D shapes,
	wherein the space of object geometry is factorized into rigid orientation, non-rigid pose, and intrinsic shape.
The resulting model can be trained from raw 3D shapes, 
	without correspondences, labels, or even rigid alignment, 
	using a combination of classical spectral geometry and 
		probabilistic disentanglement of a
		structured latent representation space.
Our improvements include more sophisticated handling of rotational invariance and the use of a diffeomorphic flow network to bridge latent and spectral space.
The geometric structuring of the latent space imparts an interpretable characterization of the deformation space of an object.
Furthermore, it enables tasks like pose transfer and pose-aware retrieval 
	without requiring supervision.
We evaluate our model on its generative modelling, representation learning, and disentanglement performance,
showing improved rotation invariance and intrinsic-extrinsic factorization quality over the prior model. 
\keywords{3D Shape \and Generative Models \and Disentanglement \and Articulation \and Deformation \and Representation Learning}
\end{abstract}

\begin{figure}
	\centering
	\includegraphics[width=0.499\textwidth]{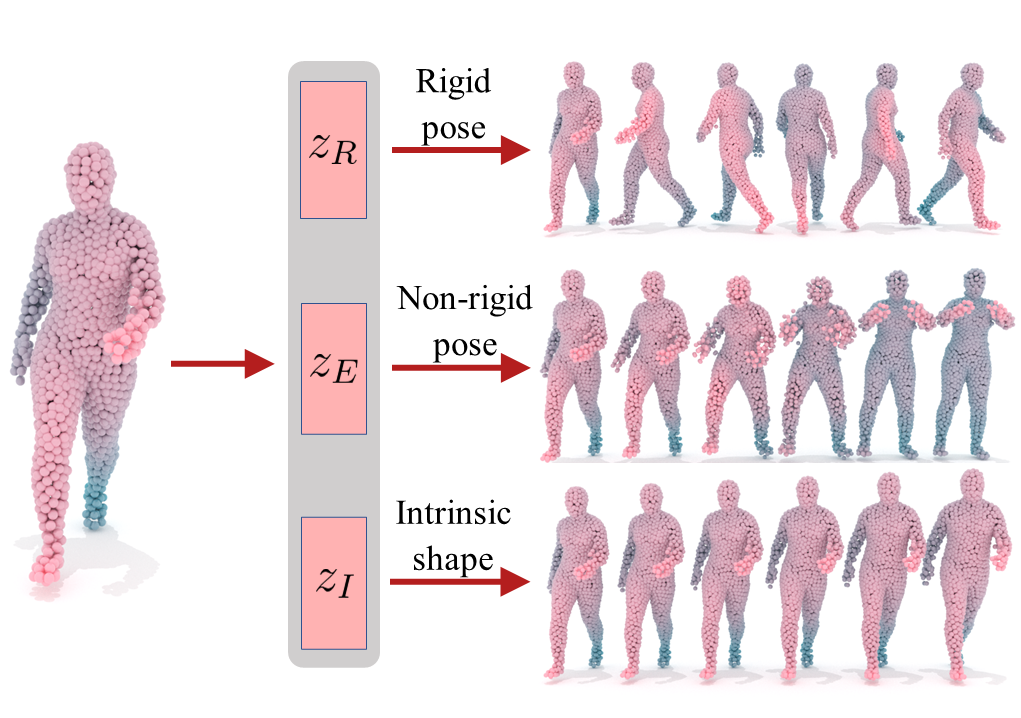}
	\caption{Depiction of overall framework goal. We factorize the latent deformation space of a given 3D object into rigid pose $z_R$, extrinsic non-rigid pose $z_E$, and intrinsic shape $z_I$, without supervision. }
	\label{fig:teaser} %
\end{figure}

\section{Introduction}
\label{sec:intro}

A major goal of representation learning is to discover and separate the underlying explanatory factors that give rise to some set of data \citep{bengio2013representation}.
For many objects, such as 3D shapes of biological entities, 
	structuring their representation within a learned model means understanding the different modes of their deformation spaces.
For instance, 
	rotating a chair does not affect its category, nor does articulated deformation of a cat alter its identity.
In general, 
	different geometric deformations may be semantically distinct,
	e.g., shape style \citep{marin2020instant}, intrinsic versus extrinsic alterations \citep{corman2017functional}, or geometric texture details \citep{berkiten2017learning}.
In other words, 
	for many objects,
	we can naturally factorize the associated deformation space,
	based on geometric characteristics.
	
Such a disentanglement can provide a useful structuring of the 3D shape representation.
For example, 
	in a vision context, 
	one could constrain inference of a 3D model from a motion sequence to change in pose, 
	but not intrinsic shape.
Or, in the context of graphics, separating shape and pose allows for tasks such as deformation transfer or shape interpolation.

\tikzset{%
	block/.style    = {draw, thick, rectangle, minimum height = 1.9em,
		minimum width = 1.9em},
	fm/.style      = {draw, circle, node distance = 1.7cm}, %
	fm2/.style      = {draw, circle, node distance = 1.7cm}, %
	input/.style    = {coordinate}, %
	output/.style   = {coordinate}, %
	fitting node/.style={
		inner sep=0pt,
		fill=none,
		draw=none,
		reset transform,
		fit={(\pgf@pathminx,\pgf@pathminy) (\pgf@pathmaxx,\pgf@pathmaxy)}
	},
	reset transform/.code={\pgftransformreset}
}

\begin{figure}
	\centering
	\begin{tikzpicture}[auto, thick, node distance=1.3cm, >={Triangle[scale width=0.97]}] %
		\draw 
		node at (0,0)[block,name=p,fill=cyan!30] (p) {$P$}
		node [fm,right of=p,fill=green!30] (x) {$x_c$}
		node [fm2,above of=x,node distance=0.95cm,fill=blue!30] (r) {$q$}
		node [fm,right of=x,fill=red!30] (ze) {$z_E$}
		node [fm,below of=ze,node distance=0.95cm,fill=red!30] (zi) {$z_I$}
		node [fm,right of=r,fill=red!30] (zr) {$z_R$}
		node [fm,right of=zr,fill=blue!30] (rhat) {$\widehat{q}$}
		node [fm2,right of=ze,fill=green!30] (xhat) {$\widehat{x}_c$}
		node [block,right of=zi,node distance=1.7cm,fill=orange!30] (lambda) {${\lambda}$}
		node [above of=zr, yshift=5mm, node distance=0.1cm](vt) {\footnotesize VAE Encoding}
		node [block,right of=xhat,fill=cyan!30,node distance=1.75cm] (phat) {$\widehat{P}$}
		node [above of=p, yshift=-1.5mm] (up1) {\includegraphics[width=0.032\textwidth]{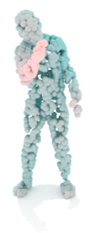}}
		node [above of=phat, yshift=-1.5mm] (up2) {\includegraphics[width=0.032\textwidth]{oth/p2.png}}
		node [below of=phat, yshift=4.5mm] (down2) {\includegraphics[width=0.079\textwidth]{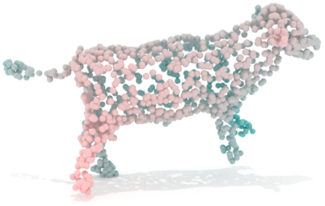}}
		node [below of=p, yshift=4.5mm] (down1) {\includegraphics[width=0.079\textwidth]{oth/p3.png}};
		\draw[->,dashed](p) -- (r);
		\draw[->,dashed](p) -- (x);
		\draw[->](x) -- (ze);
		\draw[->](x) -- (zi);
		\draw[->](r) -- (zr);
		\draw[->](zr) -- (rhat);
		\draw[->,dashed](rhat) -- (phat);
		\draw[->,dashed](xhat) -- (phat);
		\draw[->](ze) -- (xhat);
		\draw[->](zi) -- (xhat);
		\draw[<->](zi) -- (lambda);
		\draw (2.6, -1.39) rectangle (4.2, 1.4);
		\draw (1.0, -0.52) rectangle (2.4, 1.4);
		\draw node at (1.7,-0.9)[align=center] (aelabel) {\footnotesize AE \\ Encoding};
	\end{tikzpicture}
	\caption{
		A schematic overview of the combined two-level architecture used as the generative model.
		A point cloud $P$ is first encoded into $(q,x_c)$ by a \emph{deterministic} AE, 
		where
		$q$ is the quaternion representing the rotation (rigid pose) of the shape, 
		and $x_c$ the compressed representation of the input $P$, 
		in its \textit{canonical} orientation.
		$(q,x_c)$ is then further compressed into a latent representation $z=(z_R,z_E,z_I)$ of a VAE.
		The hierarchical latent variable $z$ has disentangled subgroups in red 
		(representing rotation, extrinsics, and intrinsics, respectively).
		The intrinsic latent subgroup $z_I$ is used to contain the information in the LBO spectrum ${\lambda}$ using an invertible mapping.
		Both the extrinsic $z_E$ and intrinsic $z_I$ are utilized to compute the shape $\hat{x}_c$ in the AE's latent space.
		The latent rotation $z_R$ is used to predict the quaternion $\hat{q}$.
		Finally, the decoded representation $(\hat{q},\hat{x}_c)$ is used to reconstruct the original point cloud $\hat{P}$.
		The deterministic AE mappings are shown as dashed lines; 
		VAE mappings are represented by solid lines.
	}
	\label{fig:desc}
\end{figure}

In this work,
	we consider a purely geometric decomposition of object deformations, 
	separating the space into rigid orientation, non-rigid pose, and shape.
Our method is based on methods from spectral geometry,
	utilizing the isometry invariance of the Laplace-Beltrami operator spectrum (LBOS).
The LBOS characterizes the intrinsic geometry of the shape; 
	in contrast, 
	we refer to the space of non-rigid isometric deformations of the shape as its extrinsic geometry,
	in a manner similar to
	\citet{corman2017functional}.
This decomposition is performed in the latent space of a generative model, 
	using information-theoretic methods for disentangling random variables,
	resulting in three latent vectors for rigid orientation, pose, and shape.
We apply our model to several tasks requiring this factorized structure,
	including pose-aware retrieval and pose-versus-shape interpolation 
	(for which pose transfer is a special case).
See Fig.\ \ref{fig:desc} for an overview of our approach.

We focus on minimizing the supervision required for our model, 
	eschewing requirements for identical meshing, correspondence, or labels. 
Thus, our method is orthogonal to advances in neural architectures, 
	as it can be applied to any encoder or decoder model.
For the same reason, it is also agnostic to the 3D modality 
	(e.g., meshes, voxels, or implicit fields). 
\new{We include experiments on meshes and point clouds, to showcase the versatility of our method with respect to shape modality,
	but we choose to focus on the latter,
	as they are a common data type in computer vision}\footnote{\new{However, we note that, by default, we use spectra derived from meshes, unless otherwise specified (but see \S\ref{sec:srobust}).}}.

\new{
Our method builds on a prior model \citep{aumentado2019geometric}, 
	the geometrically disentangled VAE (GDVAE),
	with two major algorithmic improvements:
	(1) we enhance the ability of the network to factorize rotation,
	and
	(2) we replace a simple spectrum regressor with a diffeomorphic flow network.
For the first point,
	we investigate two representation learning approaches 
	that allow the model to discern a canonical rigid orientation,
	with or without assuming aligned training data.	
The latter change not only guarantees that spectral information is preserved by the mapping (due to the invertibility requirement), but 
it can be readily applied to generative modelling (due to the tractability of the likelihood calculation) 
and 
it permits shape-from-spectrum computations that prevent contaminating learned latent intrinsics with extrinsic information. 
This allows us to define a better training procedure, 
		in which we use a shape-from-spectrum starting point, 
		instead of the initial input shape,
		thus ensuring that the latent intrinsics cannot access extrinsics.
These two improvements result in superior disentanglement quality, compared to the prior GDVAE model.
}		

\section{Background}
\label{sec:relwork}

\subsection{Rotation Invariant Shape Representation}
\label{sec:relwork:rot}

Invariance to rotation is generally a desirable property of shape representations,
	since many tasks (such as categorization or retrieval) tend to consider orientation a nuisance variable.
Hence, there is a significant body of work on how to learn such rigid invariance. 

Classical research includes many types of geometric features, directly computed from input shapes, 
	that are rotation invariant (e.g., \citet{guo20143d}), 
	such as structural indexing \citep{stein1992structural},
			signature of histogram orientations \citep{tombari2010unique},
			spin images
			\citep{johnson1999using},
			and point signatures \citep{chua1997point}.
More recently,
	SRINet \citep{sun2019srinet}, 
	ClusterNet \citep{chen2019clusternet},
	and
	RIConv \citep{zhang-riconv-3dv19}
	design rotation invariant hand-crafted features that can be extracted from point clouds (PCs),
	for use in learning algorithms.

Separately, rotation \emph{equivariance} has been achieved in voxel shapes 
using group convolutions \citep{worrall2018cubenet} and spherical correlations \citep{cohen2018spherical},
which can be utilized to obtain invariance.
PRIN \citep{you2018pointwise} computes rotation invariant features for point clouds, 
but requires the application of convolutions on spherical voxel grids.
SPHNet \citep{poulenard2019effective} attains rotation invariance without voxelization, by extending feature signals 
defined on a shape into $\mathbb{R}^3$, and then using a specific non-linear transform of the signal, convolved with a spherical harmonic kernel.
\new{
Additional network architectures have been applied to modelling equivariances, including 
tensor field networks \citep{thomas2018tensor,fuchs2020se}, 
graph-theoretic methods \citep{kondor2018covariant} and  
quaternion-based approaches \citep{zhao2020quaternion,zhang2020quaternion}.
See also \citet{dym2020universality} for additional discussions and theoretical analysis.
}

Other works focus on changing the input and/or utilizing other representation learning techniques, 
	which are more closely related to our work.
The PCA-RI model \citep{xiao2020endowing} 
	achieves rotation invariance by transforming each shape into an intrinsic reference frame, 
	defined by its principal components, handling frame ambiguity (due to eigenvector signs) by duplicating the input.
Info3D \citep{sanghi2020info3d} uses techniques from unsupervised contrastive learning 
	to encourage rotation invariance in the representation, 
	including the ability to handle unaligned data. 
\citet{li2019discrete} 
attain equivariance 
	by rotating each input point cloud by a discrete rotation group.
Similar to this, 
	an approximately rotation invariant encoder can be defined
	by feeding in randomly rotated copies of the input 
	\citep{sanghitowards}. 
We build on this latter approach to define one version of our 3D autoencoder (AE).
For our other approach, 
	we utilize Feature Transform Layers (FTLs) \citep{worrall2017interpretable}, 
	which allow us to make latent space rotations equivalent to 3D data space rotations.
In both cases, rather than removing rigid transforms from the embedding, 
	we attempt to factorize such transforms out, as part of the deformation space of the object.

More specifically, 
	we consider two general approaches to learning rotation invariant representations, 
	building on related work as noted above.
Both methods are 
	modality agnostic (e.g., not requiring spherical voxelization),
	architecture independent (e.g., not necessitating particular types of convolution),
	able to avoid information loss in feature extraction,
	and do not increase the cost of a forward pass (e.g., no duplication of inputs).
In this sense, 
	our method is largely orthogonal to architectural improvements 
	for PC processing, as well as the aforementioned approaches to rotation invariance.
Indeed, they can be readily applied to other 3D shape modalities.
This is because our approaches modify only the latent representation and loss calculation procedure, 
	allowing the use of arbitrary features as input, including rotation invariant ones.
	Nevertheless, we show that, despite obtaining features from a simple PointNet \citep{qi2017pointnet},
	we can still approximately attain rotation invariance without architectural alterations.
Finally, the utility of much of the related work above for generative modelling and/or autoencoding is unclear; 
	hence, we choose to use simpler architectures already known to work for these purposes 
	\citep{achlioptas2018learning,aumentado2019geometric}.

\subsection{Shape Analysis via Spectral Geometry}
\label{sec:relwork:spec}

\begin{figure}
	\centering
		\includegraphics[width=0.479\textwidth]{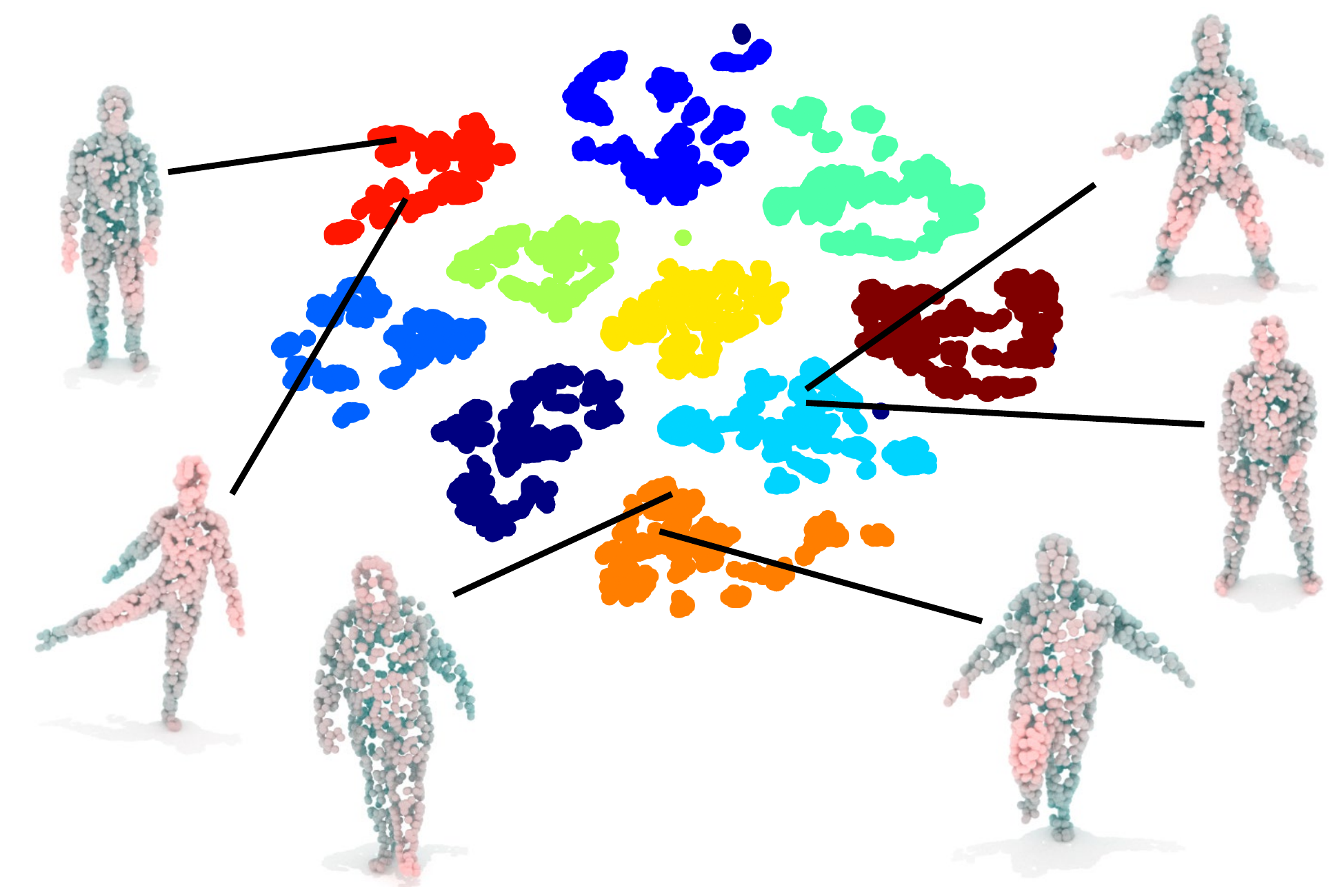}	
	\caption{Pose versus shape factorization via LBOS.
	A t-SNE \citep{maaten2008visualizing} plot of LBOSs from the Dyna dataset \citep{dyna}, 
	with illustrative accompanying point cloud representations 
	for several spectra. 
	Different body shapes are mapped close together, regardless of articulated pose.
}
\label{fig:lbotsne} %
\end{figure}

\newcommand{\lbow}{0.0517}
\begin{figure}
	\centering
	\begin{tikzpicture}[auto, thick, node distance=1.3cm, >=triangle 45] 
		\draw node at (0,0)[] (img) {   
		  \includegraphics[draft=false,width=0.39\textwidth]{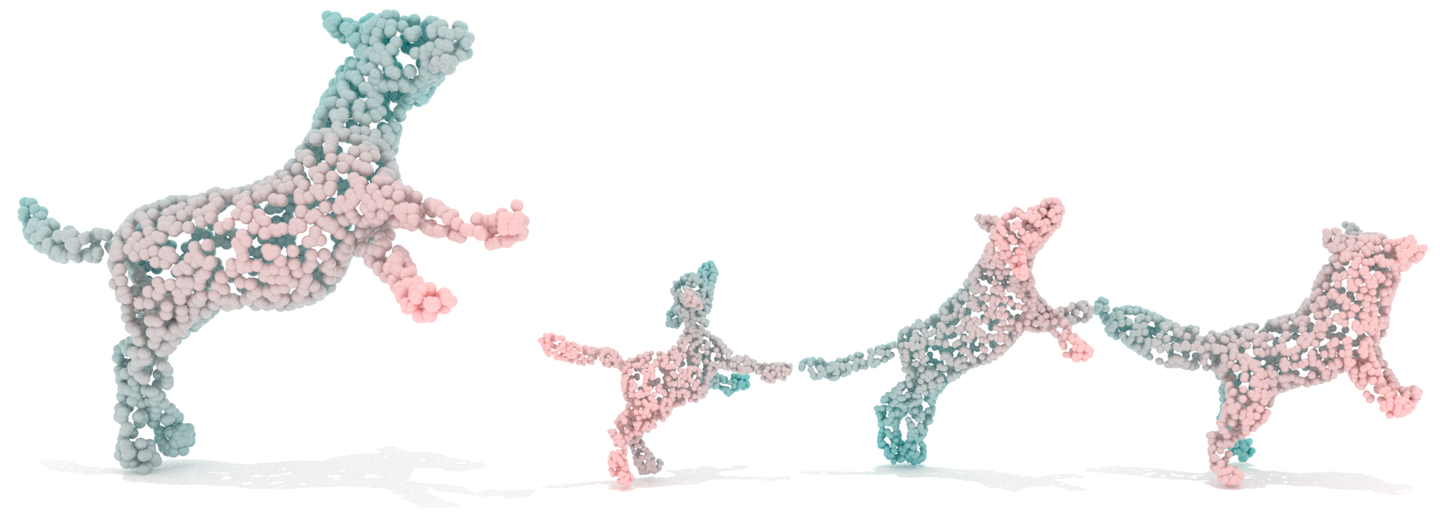} 
  		   };
  	    \draw node [right of=img, node distance=4.15cm] (graph) {
  	      \includegraphics[draft=false,width=0.045\textwidth]{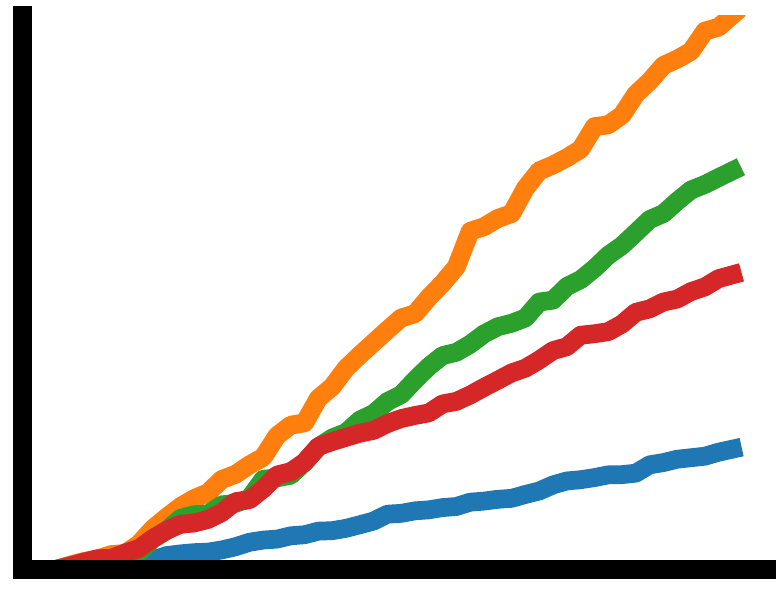} };
		\draw node at (0,0)[below of=img,node distance=1.4cm,] (label) {Changing Intrinsics};
		\path let \p1 = (img.west), \p2 = (label.west) in node (left) at (\x1,\y2) { };
		\draw[->](label.west) -- (left.center);
		\path let \p1 = (img.east), \p2 = (label.east) in node (right) at (\x1,\y2) { };
		\draw[->](label.east) -- (right.center);
		\draw node[above left=0.1mm and 0.1mm of graph,xshift=4.6mm,yshift=-2.5mm] (specsym) {$\lambda_i$};
		\draw node[below right=0.1mm and 0.1mm of graph,xshift=-2.1mm,yshift=3.8mm] (indsym) {$i$};
	\end{tikzpicture}
	\begin{tikzpicture}[auto, thick, node distance=1.3cm, >=triangle 45] 
		\draw node at (0,0)[] (img) {\includegraphics[width=0.39\textwidth,draft=false]{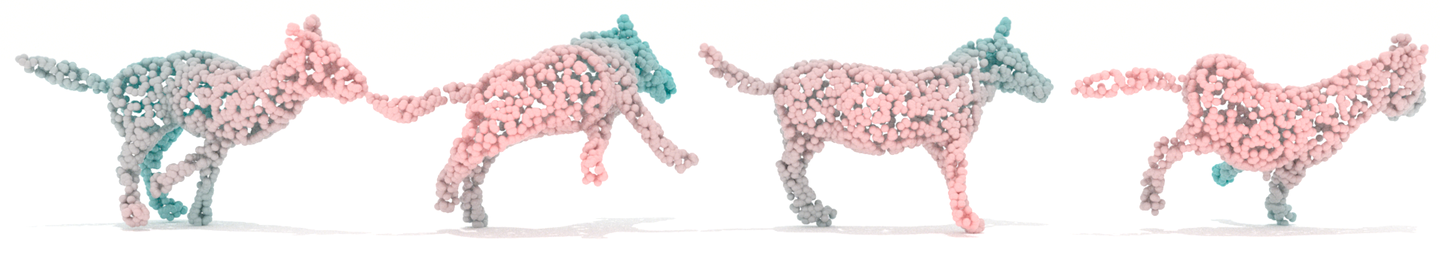}};
		\draw node [right of=img, node distance=4.15cm] (graph) {
			\includegraphics[width=0.045\textwidth,draft=false]{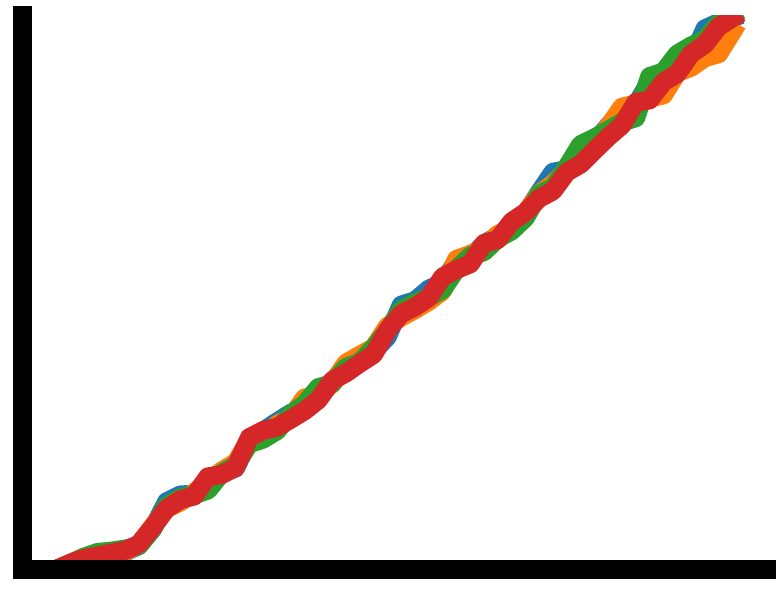} };
		\draw node at (0,0)[below of=img,node distance=0.97cm] (label) {Changing Extrinsics};
		\path let \p1 = (img.west), \p2 = (label.west) in node (left) at (\x1,\y2) { };
		\draw[->](label.west) -- (left.center);
		\path let \p1 = (img.east), \p2 = (label.east) in node (right) at (\x1,\y2) { };
		\draw[->](label.east) -- (right.center);
		\draw node[above left=0.1mm and 0.1mm of graph,xshift=4.6mm,yshift=-2.5mm] (specsym) {$\lambda_i$};
		\draw node[below right=0.1mm and 0.1mm of graph,xshift=-2.1mm,yshift=3.8mm] (indsym) {$i$};
	\end{tikzpicture}
	\caption{Visual explanation of the use of spectral geometry in characterizing 
				intrinsic versus extrinsic shape.
			We display two rows of animals, showing different \textit{intrinsics} (first row) and \textit{extrinsics} (second row), across the columns per inset.
			The plots show the LBOS $\lambda$ across shapes (indices $i$ range from 1 to 50); notice the lack of variability as extrinsics change.
	}
	\label{fig:int_vs_ext} %
\end{figure}

Any 3D surface can be viewed as a 2D Riemannian manifold $(\mathcal{M},g)$, 
	with metric tensor $g$, 
	which allows the application of differential geometry to shape analysis
	in computer graphics and vision.
One major technique in this area is the use of \textit{spectral geometry},
	which is mainly concerned with the Laplace-Beltrami Operator (LBO), $\Delta_g$, 
	and its associated spectrum 
	(i.e., the eigenvalues $\lambda$ of $-\Delta_g \phi_i = \lambda_i \phi_i$) 
	for shape processing \citep{patane2016star}.
Use of the spectrum generalizes classical Fourier analysis on Euclidean domains to manifolds,
	transferring concepts from signal processing to transforms of non-Euclidean geometry itself~\citep{taubin1995signal}.
The LBO spectrum (LBOS) characterizes the {intrinsic} properties of a manifold
	\citep{levy2006laplace,rustamov2007laplace,vallet2008spectral},
	sufficiently matching human intuition 
	on the meaning of ``shape'', 
	to the extent it is considered as its ``DNA'' 
	\citep{reuter2006laplace}.
Mathematically, \textit{intrinsic} properties of a shape are those that depend only on its metric tensor, independent of its embedding \citep{corman2017functional};
	this includes, for example, geodesic distances and the LBOS.
Among the most useful advantages of intrinsic shape properties is \textit{isometry invariance}, 
	meaning intrinsics do not change in response to alterations that do not affect the metric.
This includes rigid transforms, as well as certain non-rigid deformations, such as biological articulations (approximately).
Algorithms relying on shape intrinsics are therefore able to ignore such deformations 
	(e.g., recognize a person regardless of articulated pose).
We show some examples of the intrinsic-extrinsic geometric decomposition provided by the LBOS 
in Figures \ref{fig:lbotsne} and \ref{fig:int_vs_ext}.
We remark that we also refer to extrinsic shape as non-rigid pose, since this is the most intuitive interpretation for the case of approximately isometrically articulating objects, like animals.

Intrinsic spectral geometry processing has thus yielded numerous useful techniques for vision and graphics, 
	often due to its isometry invariance.
This includes semi-localized, articulation invariant feature extraction, such as the
	heat \citep{sun2009concise,gebal2009shape} and wave \citep{aubry2011wave} kernel signatures, 
	later extended to learned generalizations \citep{boscaini2015learning}.
Similar techniques can be applied to a variety of downstream tasks for 3D shapes as well, including
correspondence
\citep{rodola2017partial,ovsjanikov2012functional},
retrieval
\citep{bronstein2011shape},
segmentation
\citep{reuter2010hierarchical},
analogies
\citep{boscaini2015shape},
classification
\citep{masoumi2017spectral},
and manipulation
\citep{vallet2008spectral}.
Beyond the standard LBOS, more recent research has also explored 
	localized manifold harmonics \citep{neumann2014compressed,melzi2018localized},
	modifications of the LBO \citep{choukroun2018hamiltonian,andreux2014anisotropic}, and
	\textit{ex}trinsic spectral geometry \citep{liu2017dirac,ye2018unified,wang2017steklov}.

\newer{
While the above applications rely on the spectral intrinsics of existing shapes, 
    the inverse problem seeks to reconstruct a shape from an intrinsic operator (or function thereof), such as the LBO \citep{boscaini2015shape,chern2018shape,huang2019operatornet}.
In particular, the \textit{shape-from-spectrum} (SfS) task seeks to recover a shape from its LBOS, an instance of an ``inverse eigenvalue problem'' investigated in other fields (e.g., \citep{chu2005inverse,panine2016towards}).
This enables useful spectral-space tasks, such as shape style transfer and correspondence matching \citep{cosmo2019isospectralization,marin2021spectral}.
Fortunately, despite theoretical results suggesting such recovery is not always possible, due to the existence of non-isometric isospectral shapes (i.e., ``one cannot hear the shape of a drum'') \citep{kac1966can,gordon1992one}, 
it appears \textit{practically} possible in many circumstances \citep{cosmo2019isospectralization,panine2016towards}.
Indeed, \citet{cosmo2019isospectralization} show several applications of their approach to SfS recovery, though it is computationally costly and difficult to constrain.
More recently, \citet{rampini2021universal} utilize spectral perturbations to define universal geometric deformations, while \citet{moschella2022learning} apply a learning framework to process unions of partial shapes in the spectral domain.
Closest to our work, \citet{marin2020instant,marin2021spectral} apply a data-driven approach to the SfS problem, among other tasks.
}

In this work, we focus on utilizing the classical LBOS as a purely intrinsic characterization of the shape.
By exploiting the approximate articulation invariance conferred by its isometry invariance,
	we gain access to a signal that can separate intrinsic shape from articulated pose, 
	without supervision beyond the geometry itself.
While the LBO has been used to perform disentangled shape manipulations in the context of computer graphics and vision, 
	such as 
	    isometric shape interpolation \citep{baek2015isometric}, 
	    spectral pose transfer \citep{yin2015spectral}, 
	    \newer{and shape-from-spectrum recovery \citep{cosmo2019isospectralization}},
	we show how to do such manipulations within a generative model, 
	as a byproduct of the learned representation.

\subsection{Learning Shape-Pose Disentanglement}
\label{sec:relwork:learn}

A common task that has been tackled in the context of computer graphics is \textit{pose transfer}.
Utilizing a small set of correspondences, 
	an optimization-based approach can be applied to perform deformation transfer \citep{sumner2004deformation}.
Later work utilized the LBO eigenbases to perform pose transfer
	\citep{kovnatsky2013coupled,yin2015spectral}, 
	via exchanging low-frequency coefficients of the manifold harmonics.
In our work, we use the LBOS instead, 
	which avoids issues of basis computation and spectral compatibility %
	\citep{kovnatsky2013coupled}.
\citet{basset2020contact} consider transferring shape instead of pose;
	due to our symmetric formulation, our approach is also capable of this.
We refer to \citet{roberts2020deformation} for a survey of related work.

Recently, several works have attacked pose transfer from a machine learning point of view.
\citet{gao2018automatic} 
present a method for mesh deformation transfer
	using a cycle consistent GAN and a visual similarity metric,  
	but require retraining new models for each source and target set.
\citet{levinson2019latent} utilize a mesh VAE, 
	which relies on data having identical meshing, 
	to separate pose and shape via batching with identical pose and shape labels.
LIMP \citep{cosmo2020limp} disentangles intrinsic and extrinsic deformations in a generative model, 
	utilizing a differentiable geodesic distance regularizer; 
	identical meshing or labels are not required, though correspondence is. 
\citet{zhou2020unsupervised} devise a method for separating intrinsics and extrinsics 
	using corresponding meshes known only to have the same shape but different pose,
	and applying a powerful as-rigid-as-possible geometric prior.
Similarly, \citet{fumero2021learning} make use of data pairs with shared transforms to obtain a general disentanglement mechanism. 
\citet{su2021learning} also use identity-based semantic supervision, but with an adversarial mechanism on point clouds.	
\newer{
Finally, \citet{marin2020instant} consider learning a bijective mapping of the LBOS as well, 
	examining its use in the context of neural networks for several tasks,
	including spectrum estimation from point clouds and shape style transfer;
	however, they do not focus on deformation space factorization 
		or generative representation learning.
Followup work \citep{marin2021spectral} investigates shape-from-spectrum tasks, as well as shape-pose disentanglement via optimization.
}

In our work, we focus on learning a generative representation that 
	factorizes the latent deformation space 
	into intrinsic shape and extrinsic pose, without supervision.
We do not require labels (e.g., identity, pose, or shape), identical meshing, correspondence, 
	or even rigid alignment  
	-- only the raw geometry, which we use to compute the LBOS.
Rather than targeting pose transfer specifically, 
	in our model,
	the ability to transfer articulation arises naturally from the learned representation.
In particular, we build on the GDVAE model \citep{aumentado2019geometric}, 
	which disentangles shape and pose into two continuous and independent latent factors.
Our method, 
	which we refer to as the GDVAE\PP\  model,
	includes adding a bijective mapping from an LBO spectrum to the space of latent intrinsics,
	and defining a new training scheme based on this function.
We show that the resulting model is significantly improved in terms of disentanglement.

\section{Autoencoder Model}
\label{sec:ae}

Our model consists of two components: 
	an autoencoder (AE) on the 3D shape data and a variational autoencoder (VAE)
	defined on the latent space of the AE.
We show an overview of the complete framework in Fig.\ \ref{fig:desc}.

In this work, the AE is used to map a 3D point cloud (PC) to a latent vector, 
	and then decode it back to a reconstruction of the original input.
In contrast to the AE used in the prior GDVAE model \citep{aumentado2019geometric}, 
	we specifically consider the rotational invariance properties of the AE architecture.

\paragraph{Notation.}
We assume our input is a PC 
$P\in\mathbb{R}^{N_p \times 3}$, which we want to reconstruct as 
$ \widehat{P} \in \mathbb{R}^{N_p \times 3} $.
To do so, we encode $P$ into a rigid rotation, 
    represented as a quaternion $q\in\mathbb{R}^4$,
    and canonically oriented non-rigid latent shape,
    $x_c \in \mathbb{R}^{n}$,
    using learned mappings $E_r$ and $E_x$.
We can also obtain a canonical PC, $P_c = D(x_c) \in \mathbb{R}^{N_p \times 3}$, via a decoder $D$.
The details for obtaining this rigid versus non-rigid factorization are given below.

\subsection{Autoencoder Architecture}
\label{sec:ae:arch}

We consider two possible AE architectures on PCs.
	Both models attempt to regress a rotation matrix 
		and a rotation invariant latent shape representation 
		from an input.
	The first type, which we denote ``standard'' (STD), uses a straightforward reconstruction loss, 
		but also includes a random rotation before attempting to encode the shape, 
		inspired by prior work 
		\citep{sanghitowards,li2019discrete}.
	The second type relies on feature transform layers (FTLs) \citep{worrall2017interpretable} 
		to learn a latent vector space that transforms covariantly with the 3D data space under rotation,
		thus allowing the model to learn how to ``derotate'' to a canonical representation 
		(denoted ``FTL-based'').
		
Implementation-wise, we use PointNet \citep{qi2017pointnet} to encode an input point cloud, $P$ (which allows us to handle dynamic PC sizes), and fully connected layers (with batch normalization and ReLU) for all other learned mappings, unless otherwise specified. 
See Appendix \S\ref{app:aearch} for details.

\tikzset{->-/.style={decoration={
			markings,
			mark=at position #1 with {\arrow{>}}},postaction={decorate}}}

\subsubsection{Standard Architecture}
\label{sec:ae:std}

\begin{figure}
	\centering
	\begin{tikzpicture}[auto, thick, node distance=1.5cm, >={Triangle[scale width=0.97]}] 
		\draw node at (0,0)[block,name=p,fill=cyan!30] (p) {$P$};
		\draw node [fm,right of=p,fill=green!30,xshift=7mm] (x) {$x_c$};
		\node (rr) at ($(p)!0.4!(x)$) [fill=yellow!30, draw] {$\widetilde{R}$};
		\draw node [block,right of=x,fill=cyan!30,xshift=0mm] (pc) {$P_c$};
		\draw node [fm2,above right of=p,node distance=1.4cm,fill=blue!30] (r) {$q$};
		\path let \p1 = (r), \p2 = (pc) in node (t1) at (\x2,\y1) { };
		\node (t2) at ($(t1)!0.5!(pc.north)$) [] {};
		\draw node [block,right of=t2,fill=cyan!30,xshift=3mm] (phat) {$\hat{P}$};
		\draw[-](p) -- (rr);
		\draw[->](rr) -- node [midway,above,align=center,xshift=-0.5mm,yshift=-0mm] {$E_x$}  (x);
		\draw[->](p) -- node [midway,above,align=center,xshift=-3mm,yshift=-1.5mm] {$E_r$} (r);
		\draw[->](x) -- node [midway,above,align=center,xshift=-0mm,yshift=-0mm] {$D$} (pc);
		\draw[->-=0.45](pc) -- (t1.center);
		\draw[->-=0.5](r) -- (t1.center);
		\draw[->](t2.center) -- node [midway,above,align=center,xshift=-0mm,yshift=-0mm] {$R(q)$}  (phat);
	\end{tikzpicture}
	\caption{STD AE architecture. 
		Our standard AE encodes an input PC $P$ into a rigid pose component (quaternion $q$) 
		and canonically oriented shape embedding $x_c$. 
		Before passing $P$ to the shape encoder $E_x$, 
		a random rotation $\widetilde{R}$ is sampled and applied.
		The decoder $D$ then generates the canonical PC $P_c$, 
		which is rotated by $R(q)$ into the final reconstruction $\hat{P}$.
	}
	\label{fig:stdaearch} %
\end{figure}

Let $P$ be an input PC, that has potentially undergone an arbitrary rotation.
We learn two mappings as our encoder, $E_r$ and $E_x$, 
	which map $P$ to a quaternion $q = E_r(P)$ and a latent shape embedding $x_c = E_x(P)$.
Our decoder $D$ generates a canonically oriented PC $P_c = D(x_c)$, 
	which can be rotated to match the input via $\widehat{P} = P_c R(q)$, 
	where $R(q)$ is the parameter-less conversion from quaternion to rotation matrix.
Inspired by \citep{sanghitowards,li2019discrete}),
	we insert an additional layer before $E_x$ that randomly rotates $P$ 
	(i.e., $x_c = E_x(P\widetilde{R})$, $\widetilde{R}$ being a random sample),
	to further encourage learning rotation invariant features.
	We only do this for the standard architecture, shown in Fig.~\ref{fig:stdaearch}.

\subsubsection{FTL-based Architecture}
\label{sec:ae:ftl}

We also consider a slightly more complex architecture with a latent space designed for interpretability under rotation transformations, using a Feature Transform Layer (FTL) \citep{worrall2017interpretable}. 
\new{Several methods have utilized latent-space rigid transforms for mapping 3D data between views \citep{rhodin2018unsupervised,rhodin2019neural,chen2019monocular,chen2019weakly}.}
Our design is in particular inspired by prior work that extracts \textit{canonical representations} in the context of 3D human pose using FTLs  \citep{remelli2020lightweight}. 
Nevertheless, the architecture components of the FTL-AE are nearly the same as those of the STD-AE.

\paragraph{Rotational Feature Transform Layers.}
The main idea behind FTLs is to view 
	a latent vector $x \in \mathbb{R}^n$ 
	as an ordered set of subvectors 
	$U(x) = (u(x)_1,\ldots,u(x)_{N_s})\in\mathbb{R}^{N_s\times 3}$, 
	where $N_s = n/3$ and $u(x)_i\in\mathbb{R}^{3}$, 
	by simply folding it into a matrix.
Consider rotating a point cloud $P\in\mathbb{R}^{N_p\times 3}$ by a 3D rotation operation $R\in\mathbb{R}^{3\times 3}$, to get a new shape $PR$.
	By folding, one can analogously perform this rigid transformation 
		on a ``latent point cloud'', as $U(x)R$.
Ideally, applying $R$ to $P$ or $U(x)$ has the \textit{same effect} 
	(i.e., rotates the underlying shape in the same way), 
	resulting in an interpretable latent space, with respect to rotation.
We define the rotational feature transform layer $F(R,x) = U^{-1}( U(x) R )$ 
	as a latent rotation $R$ of the subvectors of $x$, 
	where the inverse $U^{-1}$ ``unfolds'' the ordered set of subvectors 
	into a single vector-valued latent variable again 
	(as opposed to the ``folding'' operator $U$). 
We will use the FTL mapping $F$ to enforce a rotation equivariant structure onto the latent space, 
	thus allowing us to ``derotate'' the shape embedding to some canonical rigid pose.
We depict the desired duality over rotations in Fig.\ \ref{fig:commute}.

\begin{figure}
	\centering
	  \begin{tikzpicture}[auto, thick, node distance=1.5cm, >={Triangle[scale width=0.97]}] 
		\draw node at (0,0)[name=p] (p) {$P$};
		\draw node [right of = p,node distance=1.9cm] (pr) {$PR$};
		\draw node [below of = p] (ux) {$U(x)$};
		\draw node [below of = pr] (uxr) {$U(x)R$};
		\draw[->](p) -- node [midway,above,align=center,xshift=-0.0mm,yshift=-0mm] {$R$} (pr);
		\draw[->](p) -- node [midway,left,align=center,xshift=-0.0mm,yshift=-0mm] {$U\circ E_x$} (ux);
		\draw[->](ux) -- node [midway,above,align=center,xshift=-0.0mm,yshift=-0mm] {$R$} (uxr);
		\draw[->](uxr) -- node [midway,right,align=center,xshift=-0.0mm,yshift=-0mm] {$D\circ U^{-1}$} (pr);
	  \end{tikzpicture}%
	  \caption{Desired commutativity structure of FTL-based architecture. Ideally, latent rotations should have the same effect as in the data space.}
  	  \label{fig:commute} %
\end{figure}

\paragraph{Architectural Details.}

\begin{figure}
	\centering
	\begin{tikzpicture}[auto, thick, node distance=1.5cm, >={Triangle[scale width=0.97]}] %
		\draw node at (0,0)[block,name=p,fill=cyan!30] (p) {$P$};
		\draw node [fm,right of=p,fill=green!30,xshift=21mm] (x) {$x_c$};
		\node (rr) at ($(p.center)!0.5!(x.center)$) [fm,fill=green!30, draw] {$\widetilde{x}$};
		\node (w) at ($(rr.center)!0.35!(x.center)$) [fill,circle,inner sep=0.5mm,minimum size=0.5mm] { };
		\draw node [block,right of=x,fill=cyan!30,xshift=0mm] (pc) {$P_c$};
		\draw node [fm2,above right of=p,node distance=1.5cm,fill=blue!30] (r) {$q$};
		\path let \p1 = (r), \p2 = (pc) in node (t1) at (\x2,\y1) { };
		\path let \p1 = (w), \p2 = (r) in node (tw) at (\x1,\y2) { };
		\node (t2) at ($(t1)!0.5!(pc.north)$) [] {};
		\draw node [block,right of=t2,fill=cyan!30,xshift=1mm] (phat) {$\hat{P}$};
		\draw[->](p) -- node [midway,above,align=center,xshift=-0.0mm,yshift=-0mm] {$\widetilde{E}_x$}  (rr);
		\draw[->](rr) -- node [pos=0.6,above,align=center,xshift=-0.5mm,yshift=-0mm] {$F$} (x);
		\draw[->](p) -- node [midway,above,align=center,xshift=-3mm,yshift=-1.5mm] {$E_r$} (r);
		\draw[->](x) -- node [midway,above,align=center,xshift=-0mm,yshift=-0mm] {$D$} (pc);
		\draw[->-=0.45](pc) -- (t1.center);
		\draw[->-=0.67](r) -- (t1.center);
		\draw[->-=0.55](tw.center) -- (w.center);
		\draw[->](t2.center) -- node [midway,above,align=center,xshift=-0mm,yshift=-0mm] {$R(q)$}  (phat);
	\end{tikzpicture}
	\caption{FTL-based AE architecture. 
		An input PC, $P$, is encoded into a quaternion $q$ and a \textit{pose-aware} embedding $\widetilde{x}$, representing the rotated (rather than the canonical) shape. The rotational FTL $F$ is then used to de-rotate $\widetilde{x}$ to obtain the \textit{canonically} oriented shape $x_c = E_x(P) = F(R(q), \widetilde{x})$.
		Finally, a reconstruction of the input, $\hat{P}$, is produced by rotating the decoded canonical PC $P_c = D(x_c)$ using the predicted rigid pose $R(q)$.
	}
	\label{fig:ftlaearch} %
\end{figure}

Utilizing similar notation to \S\ref{sec:ae:std}, we first encode $q = E_r(P)$, as before, and convert it to a predicted rotation $\widehat{R} = R(q)$. 
We then compute a \textit{non}-canonical latent shape $\widetilde{x} = \widetilde{E}_x(P)$, 
	which encodes the \textit{rotated} shape $P$. 
We then use the FTL to obtain the canonical latent shape via
	$x_c = F(\widehat{R}, \widetilde{x})$, which can be decoded via $P_c = D(x_c)$ with shared parameters.
As before, we obtain the final reconstruction via $\widehat{P} = P_c \widehat{R}$.
For notational consistency, 
we write $E_x(P) = x_c = F(\widehat{R}, \widetilde{E}_x(P))$.
See Fig.\ \ref{fig:ftlaearch} for a visual depiction.

This FTL-based architecture provides greater interpretability in terms of the effect of a rigid transform on the representation; 
	rather than trying to remove the dependence on rotation, we attempt to explicitly characterize it.
Rotations in the 3D \textit{data} space should thus 
have an identical effect
on the resulting \textit{latent} space representation (and vice versa).

\subsection{Autoencoder Loss Objective}
\label{sec:ae:loss}

The overall loss function for the AE can be written 
\begin{equation}
	\mathfrak{L}_\text{AE} = \mathcal{L}_c + \mathcal{L}_R + \mathcal{L}_P + \mathcal{L}_x,
\end{equation}
where the terms control representational consistency $\mathcal{L}_c$, rotation prediction $\mathcal{L}_R$, reconstruction $\mathcal{L}_P$ and regularization $\mathcal{L}_x$. 
These terms will be different depending on whether one uses the STD-AE (\S\ref{sec:ae:std}) or FTL-AE (\S\ref{sec:ae:ftl}).

\subsubsection{Standard Loss Objective}
\label{sec:ae:loss:std}

\paragraph{Reconstruction Loss.}
The reconstruction loss term for the STD architecture is
$	\mathcal{L}_P =  \gamma_{P} D_P(P,P_c)$,	
where $D_P$ includes a Chamfer distance and an approximate Hausdorff loss, 
similar to prior work  \citep{aumentado2019geometric,chen2019unpaired}:
\begin{equation}
    \label{eq:aereconloss}
	D_P(P_1,P_2) = \alpha_C d_C(P_1,P_2) + \alpha_H d_H(P_1,P_2),
\end{equation}
in which the squared $L_2$ distance is used between matched points.

\paragraph{Cross-Rotational Consistency Loss}
is a simple loss designed to promote consistency of the latent representation across rotations of the input, 
i.e., encourage rotation invariance.
First, we split each batch into $N_R$ copies of the same PCs; we then apply a different random rotation to each copy. 
Letting $x_{c,i}$ be the embedding of $P$ after having undergone the $i$th rotation, the loss is then
\begin{equation}
	\mathcal{L}_c = \frac{\gamma_c}{M_c} \sum_{i=1}^{N_R} \sum_{j > i} ||x_{c,i} - x_{c,j}||_2^2,
\end{equation}
where $M_c$ is the number of pairwise distances.
Note that, unlike combining features across rotated copies~\citep{xiao2020endowing,li2019discrete}, 
	this approach does not increase the computational cost of a forward pass for a single input.

\paragraph{Rotation Loss}
depends on whether we assume the data is rigidly aligned or unaligned, 
	i.e., whether we have rotational supervision or not.
In the supervised case, where the canonical rigid pose is shared across data examples,
we simply predict the real rotation for every example:
$	\mathcal{L}_R = \gamma_R d_R(R,\widehat{R})$,
where 
\begin{equation}
	d_R(R_k,R_\ell) = \frac{1}{\pi} \arccos\left( \frac{ \text{tr}(R_k R_\ell^T) - 1}{2}\right) 
	\label{eq:rotdist}
\end{equation}
is the geodesic distance on $SO(3)$ \citep{huynh2009metrics}.

In the \textit{un}supervised case, we enforce a consistency loss across rotational predictions, which does not rely on a ground truth $P$ being canonical across multiple shapes. 
Instead, it only asks that the predicted rotations of an object have the same \textit{relative} difference as the original rotations of the input 
(which should be true regardless of whether $P$ was originally canonically oriented). 
Consider rotations of a PC, $P_i = P_0 R_0 R_i$, 
    where our data follows $P = P_0 R_0$, 
    in which $P_0$ (the ground truth PC in canonical orientation) 
    and $R_0$ (the rotation of the ground truth datum) 
    are both unknown.
Our predictions are $\widehat{P}_i = P_c \widehat{R}_i$, 
    so for any two rotations of a single observed PC 
    (e.g., $P_i$ and $P_j$),
    we want 
    $\widehat{R}_i \approx R_0 R_i$
    and
    $\widehat{R}_j \approx R_0 R_j$, meaning 
    we want to predict $R_k$ \emph{relative to} $R_0$.
Combining these equations means we want 
$R_i^T \widehat{R}_i \approx R_j^T \widehat{R}_j$ for each such pair.
Formally, we write this constraint as
\begin{equation}
	\mathcal{L}_R = \frac{\gamma_r}{M_c} \sum_{i=1}^{N_R} \sum_{j > i}
						d_R( R_i^T \hat{R}_i, R_j^T \hat{R}_j ),			
\end{equation}
where $R_k$ and $\widehat{R}_k$ are the true and predicted rotations for the $k$th copy in the duplicated batch, respectively.
As noted above, in the unsupervised case, we do not necessarily wish to regress $R_i$ as $\hat{R}_i$, because the initial (derotated) input $P$ is not assumed to be in the canonical orientation of $P_c$. 

\paragraph{Regularization Loss.}
The primary purpose of the AE is to provide a space with reduced complexity and dimensionality, 
for training the generative VAE model. 
Following work on learning probabilistic samplers 
with latent-space generative autoencoders \citep{ghosh2019variational}, 
we apply a small weight decay and latent radius loss:
$	\mathcal{L}_x = \gamma_w L_2(\Theta) + \gamma_d ||x_c||_2^2$,
where $L_2(\Theta)$ is an $L_2$ weight decay on the network parameters $\Theta$.

\subsubsection{FTL-based Loss Objective}
\label{sec:ae:loss:ftl}

Similar to the network functions, the FTL-AE objective terms, as well as the training regime, are largely reused from the STD-AE.
The only major difference is that we compute reconstruction losses for both 
	the instance representation, $\widetilde{x}$, and the canonical representation, $x_c$:
\begin{equation}
	\mathcal{L}_P = \gamma_{\widetilde{P}} D_P(P_r, \widetilde{P}) + \gamma_{P} D_P(P,P_c).
\end{equation}
Here, the decoder output $\widetilde{P} = D(\widetilde{x})$ 
is encouraged to be similar to the rotated input $P_r$.

We note that the penalty $\mathcal{L}_c$, enforcing consistency of the canonical latent shape vectors $x_{c,j}$ in the FTL architecture, ties the non-canonical embeddings, $\widetilde{x}_{j}$, through an FTL operation (across rotated inputs), as follows:
\begin{align}
	|| x_{c,i} - x_{c,j} ||_2^2
	&= || ( U(\widetilde{x}_{i}) R_i -  U(\widetilde{x}_{j}) R_j ) R_i^T ||_F^2 \\
	&= || U(\widetilde{x}_{i}) -  U(\widetilde{x}_{j}) R_j R_i^T ||_F^2,
\end{align}
where we have used $U(x_{c,k}) = U(\widetilde{x}_{k}) R_k$, and the orthogonality of $R_k$ implies $||R_k v||_2^2 = ||v||_2^2$ for any $v\in \mathbb{R}^3$.

\section{Latent Variational Autoencoder Model}
\label{sec:vae}

\subsection{Overview}

Our goal is to define a disentangled generative model of 3D shapes, using a VAE. 
The model should be capable of encoding for representation inference, 
	decoding random noise for novel sample generation, and
	allowing factorized latent control of intrinsic shape and extrinsic non-rigid pose. %
The latter decomposition is made possible by use of the LBO spectrum,
	which allows us to separate non-rigid deformations into intrinsic shape and \new{extrinsic (articulated or non-rigid) pose} 
	(see \S\ref{sec:relwork:spec}).

Following \citet{ghosh2019variational} and \citet{achlioptas2018learning}, 
	we use the AE latent space  to define our generative model and disentangled representation learning.
This allows us to train with much larger batch sizes (useful for information-theoretic objectives based on estimating marginal distribution properties from samples), and generally obtain better computational efficiency.
See Fig.\ \ref{fig:desc} for a pictographic overview.

Compared to our prior GDVAE model \citep{aumentado2019geometric}, %
	we replace a simple predictor of the LBOS from the latent intrinsic shape 
	with a diffeomorphic mapping between the two quantities. 
	This allows us to use the spectrum directly in training (see \S\ref{sec:trainregimes}) and increase the dimensionality of the latent intrinsics, improving representation performance.

\subsection{Model Architecture}

\subsubsection{Hierarchically factorized VAE}

\label{sec:hfvae}

The core of our VAE model is the Hierarchically factorized VAE (HFVAE) model \citep{esmaeili2018structured},
	which permits penalization of mutual information between sets of vector-valued random variables.
This allows us to enforce the latent intrinsics to be separate from the latent extrinsics, specifically.

\sloppy Let $(q,x_c)$ be an encoded input from the AE.
We define
$z_R\sim \mathcal{N}(\mu_R(q), \Sigma_R(q))$,
$z_E\sim \mathcal{N}(\mu_E(x_c), \Sigma_E(x_c))$, and
$z_I\sim \mathcal{N}(\mu_I(x_c), \Sigma_I(x_c))$ 
to be the latent encodings of the rotation, extrinsic shape, and intrinsic shape, respectively, 
sampled from their variational latent posteriors.
Our decoder is deterministic: 
$\widehat{q} = D_q(z_R)$ and $\widehat{x}_c = D_x(z_E, z_I)$.
All three variables use isotropic Gaussians as latent priors.
See Appendix \S\ref{app:vaearch} for further details.

\subsubsection{Normalizing Flow for Spectrum Encoding}
\label{sec:normflowforspectrum}
In order to encourage $z_I$ to hold only shape intrinsics, we utilize the LBOS.
In particular, we define an \textit{invertible} mapping between $\lambda$ and $\mu_I$.
Let $\widetilde{\mu}_I = f_\lambda(\lambda)$ be the latent encoding of a \emph{real} spectrum (i.e., computed from a shape), $\lambda$,
and $\widehat{\lambda} = g_\lambda(\mu_I)$ be the predicted spectrum, 
	with $g_\lambda = f_\lambda^{-1}$.
We implement $f_\lambda$ as a \textit{normalizing flow} network \citep{papamakarios2019normalizing,kobyzev2020normalizing},
defining a bijective mapping between $z_I$-space and the space of spectra.
For VAE calculations, we use
$\widetilde{z}_I \sim \mathcal{N}(\widetilde{\mu}_I,\widetilde{\Sigma}_I(\lambda))$.

\new{
Briefly, flow networks are specialized neural modules with two general properties: 
(1) being a diffeomorphic mapping,
and
(2) having a simple analytic Jacobian determinant.
These properties allow tractable exact likelihood computations through the network, via the probability chain rule through each layer \citep{papamakarios2019normalizing}.
Many architectures have been proposed with these functional properties
(e.g., \citep{kingma2016improved,kingma2018glow,dinh2016density,dinh2014nice})
and they have been applied to generative modelling tasks in both 2D and 3D
\citep{kingma2018glow,yang2019pointflow}, 
as the tractable exact likelihood allows for stable training of the distribution matching loss to the prior, at the cost of requiring the dimensions of the input and output space to match and restricting the class of allowed neural architectures.
}

\new{
Using a flow mapping ensures that $f_\lambda(\lambda)$ can hold complete information about $\lambda$, since the learned network is guaranteed to be diffeomorphic (i.e., it is invertible and differentiable in either direction).
Unlike \citet{aumentado2019geometric},
	this approach also allows various ``shape-from-spectrum'' applications \citep{marin2020instant}, 
	\newer{which we explore in \S\ref{sec:meshexps:coma}}.
Thus, the flow network confers an additional benefit, which is the presence of a mapping from $\lambda$-space to $z_I$-space,
which allows us to define a novel training regime that prevents encouraging the network to store extrinsic information in the $z_I$-space for reconstruction, by instead using $\widetilde{\mu}_I$ for reconstruction and pushing $\mu_I$ to match it
(see \S\ref{sec:trainregimes}).
Finally, it has the benefit of being specifically designed for likelihood-based generative modelling,
	hence its training procedure synergizes well with the HFVAE. %
In particular, since we want the latent intrinsic space $z_I$ to conform to a Gaussian prior (which we enforce with the HFVAE prior-matching losses), 
we also wish to ensure anything mapped from $\lambda$-space to there
does as well.
Fortunately, the tractable likelihood of flow networks allows us to directly optimize a prior-matching likelihood, which is not an upper-bound (unlike for VAEs).
See \S\ref{sec:flowlikeloss} for details.
}

\subsection{VAE Loss Function}
\label{sec:vae:loss}

The VAE model is trained with the following objective:
\begin{equation}
   \mathfrak{L}_\text{VAE} =  
	  \mathcal{L}_\text{HF} 
	+ \mathcal{L}_\lambda      
	+ \mathcal{L}_F
	+ \mathcal{L}_D,
\end{equation}
where $\mathcal{L}_\text{HF}$ is the hierarchically factorized VAE loss \citep{esmaeili2018structured}, 
$\mathcal{L}_\lambda$ measures the likelihood defined by the spectral flow network between spectra $\lambda$ and latent intrinsics $z_I$, 
$\mathcal{L}_F$ is a consistency loss between the VAE (mapping between $x_c$ and $z$ space) and the flow network,
and
$\mathcal{L}_D$ is an additional disentanglement penalty.
We next define the component loss functions used in this complete objective in detail.
Note that we assemble two versions of this loss, 
	expounded in \S\ref{sec:foloss} and \S\ref{sec:pploss},
	which differ in whether to use the latent intrinsics derived 
	from $x_c$ or $\lambda$.

\subsubsection{HFVAE Loss $\mathcal{L}_{\text{HF}}$}
\label{sec:vae:loss:hfvae}

Recall that our latent space $z = (z_R,z_E,z_I)$ is \textit{structured}, 
	in that we can partition it into three sub-vectors.
Our goals are to 
(1) push $z$ to follow an isotropic Gaussian latent prior and 
(2) force each component group $z_g$, with $g\in\{R,E,I\}$, to be independent from the other two groups, in an information-theoretic sense.
Specifically, we use \textit{total correlation} (TC), 
	a measure of multivariate mutual information, 
	between latent groups to optimize disentanglement \citep{watanabe1960information}.

Prior work on structured disentanglement \citep{esmaeili2018structured} has shown that the VAE objective can be decomposed in a hierarchical fashion via
\begin{align}
	\mathcal{L}_{\text{HF}}[z] 
	= \; & \omega_R L_R + 
	\beta_1 \sum_g \mathcal{I}_\text{TC}[z_g] \; +
	\nonumber \\
	&\beta_2 \sum_{d,g} 
	\mathcal{D}_\text{KL}[q_\phi(z_{g,d}) \mid\mid P(z_{g,d})] \; + \nonumber \\
	&\beta_3 \mathcal{I}[x_c,z] +
	\beta_4 \mathcal{I}_\text{TC}[z]
\end{align}
where 
$L_R$ denotes the reconstruction loss, %
the $\beta_1$ term controls the \textit{intra}-group total correlation,
the $\beta_2$ term penalizes the dimension-wise KL-divergence from the latent prior, 
the $\beta_3$ term controls the mutual information between $x_c$ and $z$, and
the $\beta_4$ term controls the \textit{inter}-group total correlation.
The latter term, $\mathcal{I}_\text{TC}(z)$, is the most important for our application,
as it encourages statistical independence between latent intrinsics and extrinsics -- this is our \textit{disentanglement} objective.

Recall that the VAE input is a 
quaternion $q$ and 
canonical shape vector $x_c$,
while the output are the regressions 
$\widehat{q}$ and $\widehat{x}_c$.
The reconstruction loss, $L_R$, is written as follows:
\begin{equation}
	L_R((q,x_c),(\widehat{q},\widehat{x}_c)) = 
			\frac{|| x_c - \widehat{x}_c ||_2^2}{ n\,\mathbb{E}[|| x_c ||_2^2]} 
		  + \omega_q d_q(q,\widehat{q}),
\end{equation}
where 
$n = \text{dim}(x_c)$, 
the expected norm $\mathbb{E}[|| x_c ||_2^2]$ normalizes for differing AEs (making hyper-parameter setting across models easier), 
and $d_q(q_1, q_2) = 1 - |q_1\cdot q_2|$ is a distance metric on rotations, through unit quaternions $q_1, q_2$ \citep{huynh2009metrics}. 

\subsubsection{Flow Likelihood Loss $\mathcal{L}_\lambda$}
\label{sec:flowlikeloss}
Since $f_\lambda$ is a normalizing flow network and we want to enforce $\widetilde{z}_I$ to follow the Gaussian latent prior, 
	we can simply use the standard likelihood objective \citep{papamakarios2019normalizing,kobyzev2020normalizing}: 
	\begin{equation}
	\label{eq:plambda}
		P_\lambda(\lambda) = P_{z_I}(f_\lambda(\lambda)) \,
		\left|\det \mathcal{J}[f_\lambda](\lambda)\right|,
	\end{equation}
where $P_{z_I}$ represents the density of an isotropic Gaussian (latent prior of $z_I$)
and $\mathcal{J}[f]$ is the Jacobian of $f$.
We use a weighted log-likelihood as the final loss:
$\mathcal{L}_\lambda = - \omega_p \log P_\lambda(\lambda) $.
This loss enforces $\widetilde{z}_I$ to follow the latent prior,
	as in most flow-based generative models.
\new{While it is similar to the HFVAE loss on $z_I$, 
	it is an exact likelihood \citep{papamakarios2019normalizing,kobyzev2020normalizing}, 
	rather than a lower bound.
As discussed in \S\ref{sec:normflowforspectrum},
this is intuitively possible due to the use of a diffeomorphic transform, constrained to have an computationally tractable Jacobian determinant.
}

\subsubsection{Spectral Intrinsics Consistency Loss $\mathcal{L}_F$}
\label{sec:spectralconsis}
We also want the VAE encoder to be consistent with the spectral flow network,
    so we apply a loss between the spectral and latent intrinsic space outputs:
\begin{equation}
	\mathcal{L}_F =  \omega_I || \mu_I - \widetilde{\mu}_I ||_2^2 +
		\omega_\lambda d_\lambda(\lambda, \widehat{\lambda}).
\end{equation}
$\widetilde{\mu}_I = f_\lambda(\lambda)$, $\widehat{\lambda} = g_\lambda(\mu_I)$, and $d_\lambda$ 
is a weighted distance between spectra \citep{aumentado2019geometric},
\begin{equation}
	d_\lambda(\lambda, \widehat{\lambda}) = \frac{1}{N_\lambda} \sum_{n = 1}^{N_\lambda}
											\frac{ |\lambda_n - \widehat{\lambda}_n| }{n},
	\label{eq:specdist}
\end{equation}
where $N_\lambda$ is the number of elements used in the spectrum.
This formulation is inspired by Weyl's estimate 
\citep{weyl1911asymptotische,reuter2006laplace}, 
which posits approximately linear eigenvalue growth asymptotically.
The motivation is to avoid overweighting the higher elements of the spectrum
	(corresponding to higher geometric frequencies and thus noisier, small-scale shape details).
See also \citet{cosmo2019isospectralization}.
Note that this does not assume a particular structure for the LBO, nor for the growth of its eigenvalues; 
rather, it is a heuristic for reducing the effect of the monotonic growth  of $\lambda$ (i.e., non-linear growth will simply change the relative importance of the frequencies in the loss).

\subsubsection{Additional Disentanglement Losses $\mathcal{L}_D$}
\label{sec:additionaldisentloss}

Following \citet{aumentado2019geometric}, we utilize two additional losses to promote disentanglement.
The first is motivated by  \citet{kumar2017variational}, 
	penalizing the covariance between latent groups:
\begin{equation}
	\mathcal{L}_\Sigma = \sum_{g\ne \widetilde{g}} \sum_{i,j} \left|\widehat{\Sigma}[\mu_g, \mu_{\widetilde{g}}]_{i,j}\right| ,
\end{equation}
where $\widehat{\Sigma}$ is the empirical covariance matrix between latent vectors, computed per batch, and $g,\widetilde{g}\in\{R,E,I\}$.
The second takes advantage of the differentiable nature of the networks involved, 
	directly penalizing the rate of change in the intrinsics as the extrinsics are varied (and vice versa).
This is implemented as a penalty on the Jacobian between latent groups
\begin{equation}
\mathcal{L}_\mathcal{J} = \left| \left| \frac{\partial \widehat{\mu}_E }{ \partial \mu_I } \right| \right|_F^2
+ \left| \left| \frac{\partial \widehat{\mu}_I }{ \partial \mu_E } \right| \right|_F^2,
\end{equation}
where $\widehat{\mu}_g = \mu_g(\widehat{x}_c)$ is the re-encoding of the reconstructed shape from the AE, 
	$\widehat{x}_c = D_x(z_E, z_I)$, 
such that 
$\frac{ \partial \widehat{\mu}_g }{ \partial {\mu}_{\widetilde{g}} } 
= \frac{\partial \widehat{\mu}_g }{\partial \widehat{x}_c} 
\frac{\partial \widehat{x}_c }{\partial \mu_{\widetilde{g}} } $ 
for $g\ne \widetilde{g}$ and $\mu_g$ is the approximate posterior mean from which $z_g$ is sampled.
Hence, the final loss term is given by
$	\mathcal{L}_D = 
	\omega_\Sigma \mathcal{L}_\Sigma + \omega_\mathcal{J} \mathcal{L}_\mathcal{J}$.
	
\subsection{Training Regimes}

\label{sec:trainregimes}

We consider two methods of training, \new{which differ in the manner in which the latent variables are obtained at training time}.
The first is similar to the original Geometrically Disentangled VAE (GDVAE) model, 
	where $z_I$ is used for reconstruction and predicting the spectrum. 
	This is the ``flow-only'' (FO) model.
The second takes advantage of the shape-from-spectrum capabilities of the bijective flow mapping,
	using $\widetilde{\mu}_I = f_\lambda(\lambda)$ for reconstruction 
	(which does not depend on $x_c$),
	and encouraging $\mu_I(x_c)$ to be close to $\widetilde{\mu}_I$.
We refer to these models as 
	\textbf{GDVAE-FO} and \textbf{GDVAE\PP}, 
	respectively.
Notice that the latter approach more stringently separates extrinsics and intrinsics,
	as the decoder has more limited access 
	to extrinsics from  
	$f_\lambda(\lambda)$, as opposed to using $x_c$.
We visualize the two pathways in Fig.\ \ref{fig:vaecompare}.
\new{Notice that the two training regimes do not differ in their architecture, hyper-parameters, and structure of the forward pass at inference, but only in the structure of the forward pass at training time.}

\subsubsection{GDVAE-FO Loss}
\label{sec:foloss}

The ``flow-only'' model is most similar to the prior GDVAE model \citep{aumentado2019geometric}.
We want the encoded intrinsic shape vector $\mu_I(x_c)$ to hold as much information as possible about the spectrum.
This is accomplished through the diffeomorphic mapping to $\lambda$ and the spectral losses in $\mathcal{L}_F$.
In other words, 
we reconstruct via $\widehat{x}_c = D_x(z_E, z_I)$ and $\widehat{\lambda} = g_\lambda(\mu_I)$. 
The disentanglement losses $\mathcal{L}_\text{HF}$ and $\mathcal{L}_D$ are computed with $\mu_I$.

\begin{figure}
	\centering
	\begin{tikzpicture}[auto, thick, node distance=1.3cm, >={Triangle[scale width=0.97]}] %
		\draw 
		node at (0,0)[fm,fill=green!30] (x) {$x_c$}
		node [fm,above right of=x,fill=red!30] (ze) {$z_E$}
		node [fm,right of=x,node distance=1.15cm,fill=red!30] (zi) {$z_I$}
		node [fm2,right of=ze,fill=green!30,node distance=1.25cm] (xhat) {$\widehat{x}_c$}
		node [block,right of=zi,node distance=1.3cm,fill=orange!30] (lambdahat) {$\widehat{\lambda}$}
		node [fm,right of=xhat,node distance=1.37cm,fill=red!30] (zitilde) {$\widetilde{z}_I$}		
		node [block,right of=zitilde,node distance=1.37cm,fill=orange!30] (lambda) {${\lambda}$}
		;
		\draw[->](x) -- (ze);
		\draw[->](x) -- (zi);
		\draw[->](ze) -- (xhat);
		\draw[blue,->](zi) -- (xhat);
		\draw[->](zi) --  node [midway,below,align=center,xshift=-0.0mm,yshift=-0mm,text=black,] {$g_\lambda$}  (lambdahat);
		\draw[red, ->](lambda) --  node [midway,below,align=center,xshift=-0.0mm,yshift=-0mm,text=black,] {$f_\lambda$}  (zitilde);
		\draw[red, ->](zitilde) -- (xhat);
	\end{tikzpicture}
	\caption{
		Diagram of VAE mappings, 
		depicting the ability to use latent intrinsics derived from $x_c$ versus $\lambda$.
		Black lines indicate mappings always run in training. 
		The {\color{blue}blue arrow} is used when predicting the latent reconstruction from $x_c$ rather than $\lambda$, 
		which is useful at inference time (when $\lambda$ may not be known) and for the GDVAE-FO training scheme (see \S\ref{sec:trainregimes}).
		The {\color{red}red arrows} depict using latent intrinsics directly procured from the LBOS $\lambda$, 
		as in the GDVAE\PPP \ training scheme.
	}
	\label{fig:vaecompare}
\end{figure}

\subsubsection[GDVAE++ Loss]{GDVAE$\,$\PP\  Loss}
\label{sec:pploss}

For the GDVAE$\,$\PP \  loss, we use the known spectrum to compute the output latent shape.
The idea during training is to enforce the latent intrinsics used for reconstruction 
(in this case, $\widetilde{z}_I$)
to \textit{only} hold intrinsic geometry (using $f_\lambda(\lambda)$), 
and push $z_I$ (inferred from $x_c$) to be close to it.
Thus, 
$\widehat{x}_c = D_x(z_E, \widetilde{z}_I)$ is used for reconstruction, 
where 
$\widetilde{z}_I \sim \mathcal{N}(\widetilde{\mu}_I = f_\lambda(\lambda), \widetilde{\Sigma}_I(\lambda))$.
In addition, the disentanglement losses $\mathcal{L}_\text{HF}$ and $\mathcal{L}_D$ are computed with $\widetilde{\mu}_I$.
Note that this training strategy does not preclude us 
	from processing shapes \textit{without} spectra at test time, 
	which we do for our evaluations.

\section{Experimental Results}
\label{sec:results}

\subsection{Datasets}

\newcommand{\hh}{0.085}
\newcommand{\hm}{0.105}
\begin{figure*}[t]
	\centering
	\includegraphics[height=0.17\textwidth]{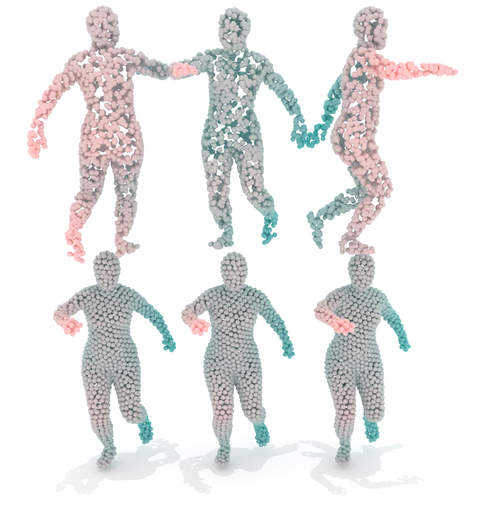}
	\includegraphics[height=0.17\textwidth]{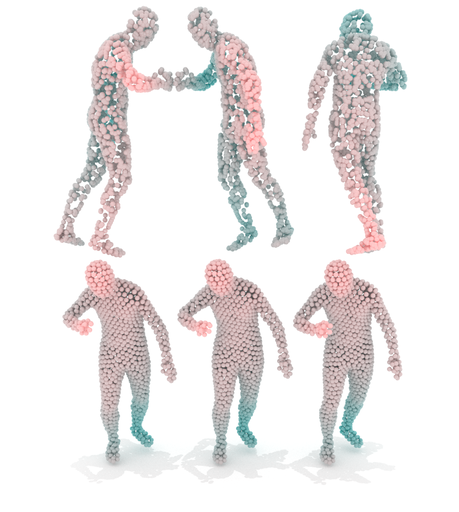}
	\includegraphics[height=0.13\textwidth]{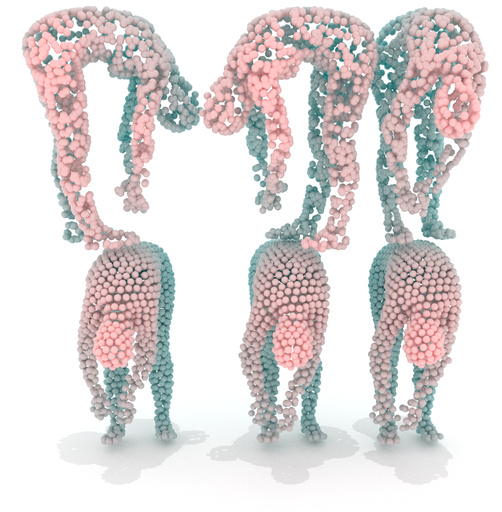}
	\includegraphics[height=0.17\textwidth]{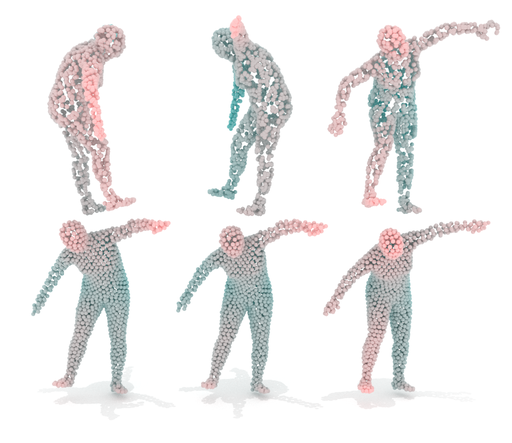}
	\includegraphics[height=0.13\textwidth]{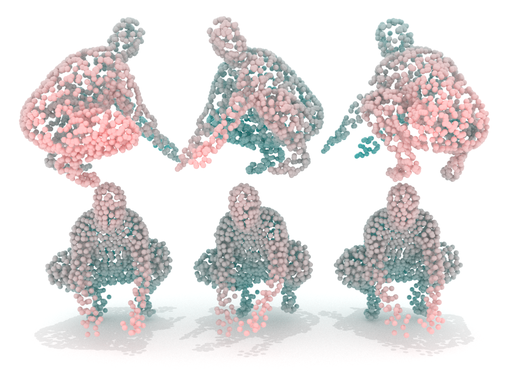}	
	\includegraphics[height=0.17\textwidth]{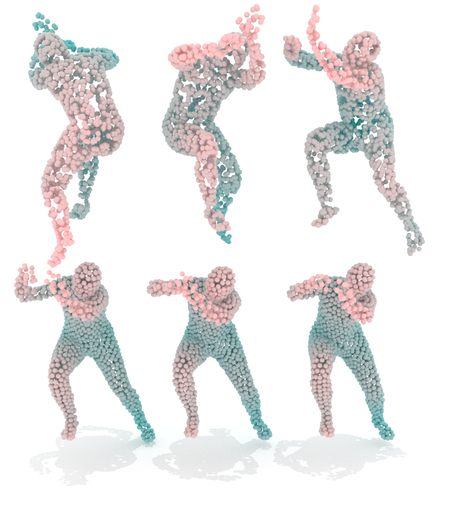} \\
	\includegraphics[height=\hh\textwidth]{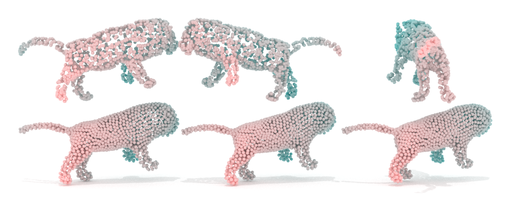}
	\includegraphics[height=\hh\textwidth]{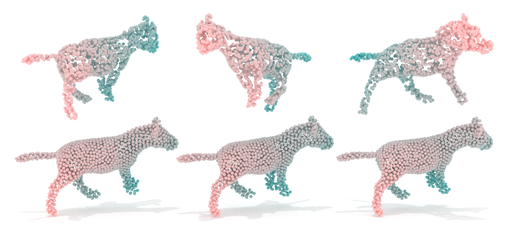}
	\includegraphics[height=\hh\textwidth]{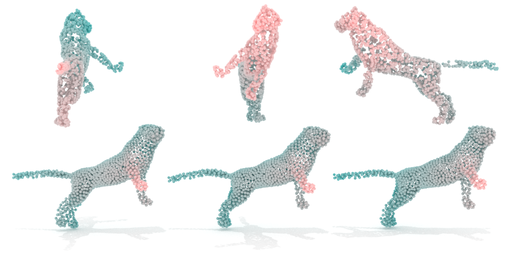}
	\includegraphics[height=\hh\textwidth]{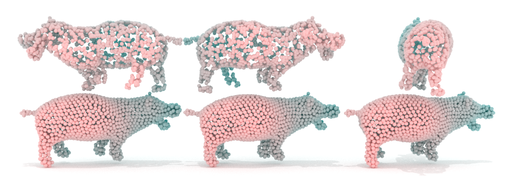}
	\includegraphics[height=\hh\textwidth]{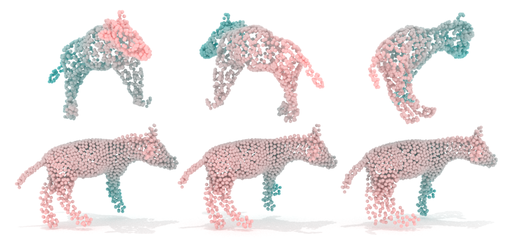} \\
	\includegraphics[height=\hm\textwidth]{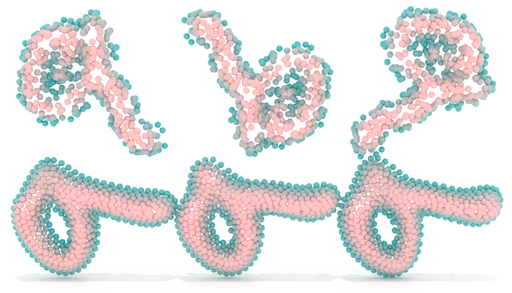}
	\includegraphics[height=\hm\textwidth]{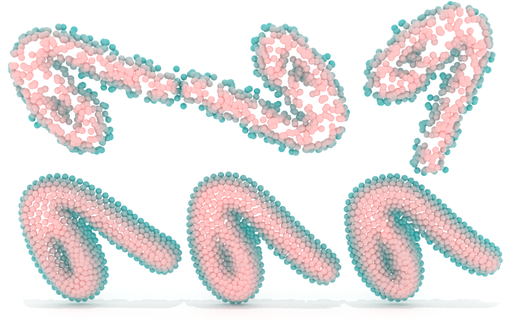}
	\includegraphics[height=\hm\textwidth]{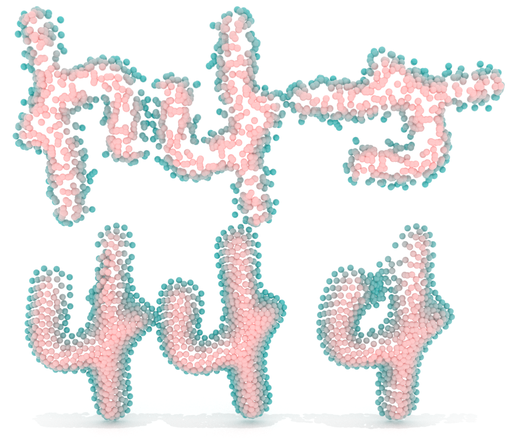}
	\includegraphics[height=\hm\textwidth]{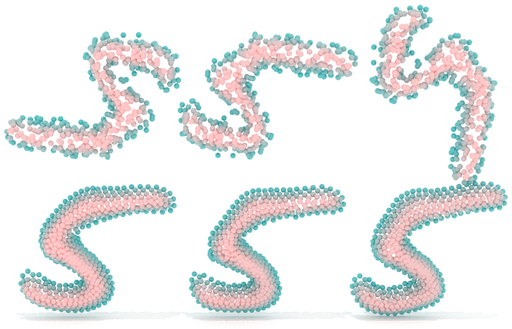}
	\includegraphics[height=\hm\textwidth]{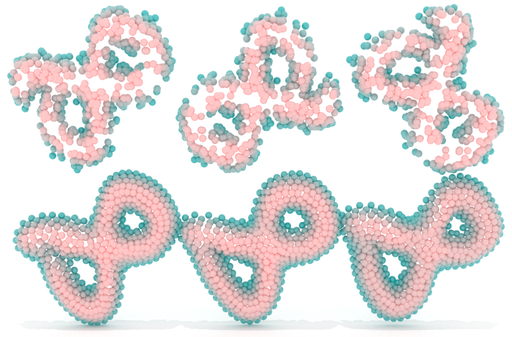}	
	\includegraphics[height=\hm\textwidth]{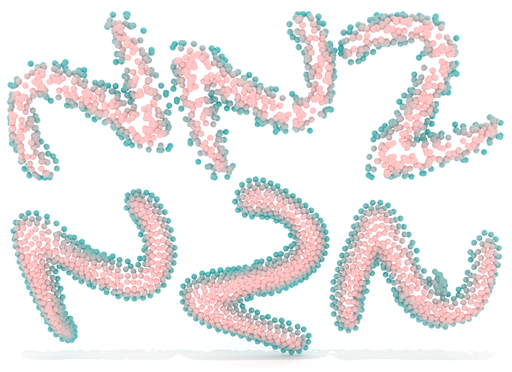} \\
	\caption{ Example AE reconstructions and rigid canonicalizations of input point clouds.
		Per inset, 
			the top row shows input point clouds $P$ under random rotations,
		while the bottom row displays the resulting canonical decodings $P_c$.
		Colors indicate depth.
		Architectures used are SMPL-FTL-S, SMAL-FTL-S, and MNIST-FTL-U, 
			respectively (see Table \ref{tab:ae}).
		Since MNIST is unsupervised, we rotate the canonical output by a constant rotation for visualization purposes.
		In particular, notice that the first two MNIST insets (``9" and ``6") are rotated to match 
			(which would be incorrect in the supervised case).
		We show some failure cases in the last column of each row:
			for SMPL, the pose (e.g., arms) is incorrectly reconstructed;
			for SMAL, the hind-legs of the third canonical PC does not match its counterparts, though the overall pose does;
			for MNIST, the canonical decoding simply fails to match across rotations.
		Note that we show additional reconstructions (through the VAE) 
		in Fig.\ \ref{fig:vaerecon}. 
	}
	\label{fig:aerecon} %
\end{figure*}

\begin{figure}
	\centering
	\includegraphics[width=0.064\textwidth]{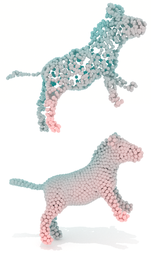}\hspace{\fill}
	\includegraphics[width=0.050\textwidth]{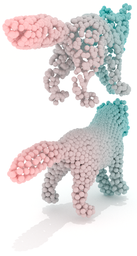}\hspace{\fill}
	\includegraphics[width=0.072\textwidth]{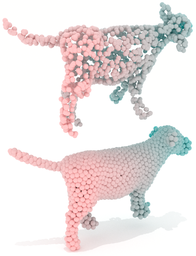}\hspace{\fill}
	\includegraphics[width=0.085\textwidth]{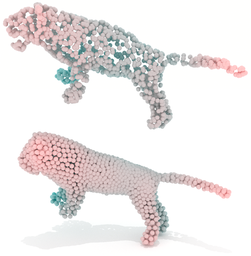}\hspace{\fill}
	\includegraphics[width=0.056\textwidth]{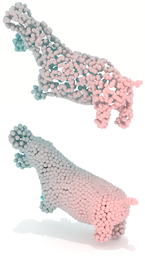}\hspace{\fill}
	\includegraphics[width=0.071\textwidth]{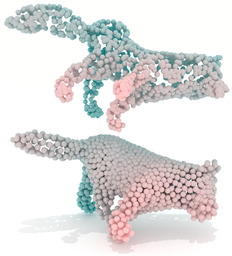}\hspace{\fill}
	\includegraphics[width=0.044\textwidth]{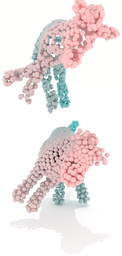} \\
	\includegraphics[width=0.043\textwidth]{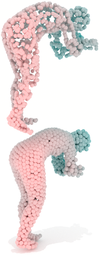}\hspace{\fill}
	\includegraphics[width=0.026\textwidth]{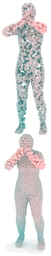}\hspace{\fill}
	\includegraphics[width=0.036\textwidth]{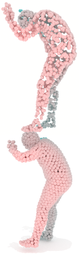}\hspace{\fill}
	\includegraphics[width=0.049\textwidth]{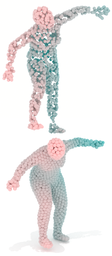}\hspace{\fill}
	\includegraphics[width=0.03\textwidth]{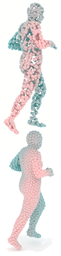}\hspace{\fill}
	\includegraphics[width=0.046\textwidth]{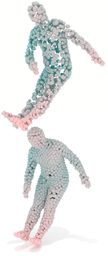}\hspace{\fill}
	\includegraphics[width=0.037\textwidth]{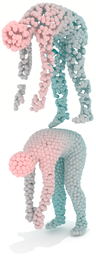}\hspace{\fill}
	\includegraphics[width=0.028\textwidth]{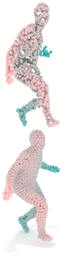}\hspace{\fill}
	\includegraphics[width=0.021\textwidth]{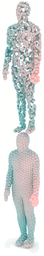} \\	
	\includegraphics[height=0.079\textwidth]{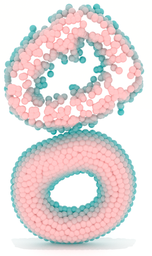}	\hspace{\fill} 	
	\includegraphics[height=0.079\textwidth]{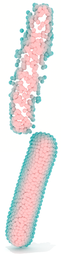} 	\hspace{\fill}
	\includegraphics[height=0.079\textwidth]{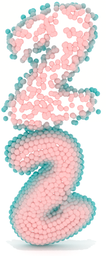} 	\hspace{\fill}	
	\includegraphics[height=0.079\textwidth]{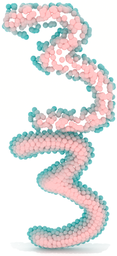} 	\hspace{\fill}	
	\includegraphics[height=0.079\textwidth]{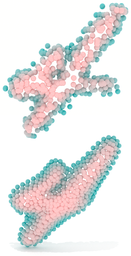} 	\hspace{\fill}	
	\includegraphics[height=0.079\textwidth]{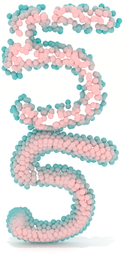}	\hspace{\fill}	
	\includegraphics[height=0.079\textwidth]{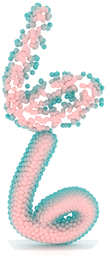} 	\hspace{\fill}%
	\includegraphics[height=0.079\textwidth]{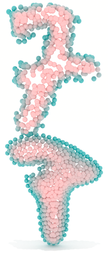} 	\hspace{\fill}
	\includegraphics[height=0.079\textwidth]{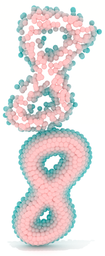} 	\hspace{\fill}
	\includegraphics[height=0.079\textwidth]{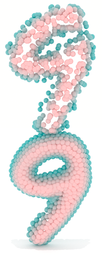} 	
	\caption{ 
		Reconstructions through the VAE.
		Odd rows are inputs; even rows are reconstructions.
		Since the model is learned in $x_c$-space, reconstruction error manifests as shapes being valid, 
		but slightly off in shape and/or pose 
		(e.g., left-most and right-most insets of top row).
		All models use the FTL-based AE.
		See Fig.\ \ref{fig:aerecon} for qualitative example reconstructions through only the AE.
		Note that inputs and outputs have the same number of points.
	}
	\label{fig:vaerecon} %
\end{figure}

We use the same datasets as in \citet{aumentado2019geometric}.
Specifically, we consider MNIST \citep{lecun1998gradient}, SMAL \citep{Zuffi:CVPR:2017}, and SMPL \citep{SMPL:2015}.
We also assemble a Human-Animal (HA) mixed dataset by combining data from SMAL and SMPL.
Note that, in all cases, we perform a scalar rescaling of the dataset such that 
	the largest bounding box length is scaled down to unit length.
This scale is the same across PCs
	(otherwise the change in scale would affect the spectrum for each shape differently).
We apply random rotations about the gravity axis (SMAL and SMPL) or the out-of-image axis (MNIST).
For rotation supervision, 
    the orientation of the raw data is treated as canonical.
\new{We also remark that we use LBOSs derived from the mesh shape, rather than PCs, unless otherwise specified.}
See Appendix \S\ref{app:datasets} for additional dataset details.

\begin{table}
	\centering
\caption{AE evaluation on held-out test data. 
	     Metrics (left to right) refer to the Chamfer distance in reconstructions and the rotational consistency measures 
	     (in 3D and $x_c$-space, respectively). 
	     HA is the humans and animals dataset (SMPL+SMAL). 
		 For each model, STD and FTL refer to the type of AE architecture, 
		 	and U and S denote the use (S) or lack of (U) rotational supervision. 
		 For the HA dataset, $a/b$ denote the values on the SMAL and SMPL test sets, respectively.
		 Also, note that, for HA, SMPL shapes are scaled with the SMAL maximum bounding box length; 
		 therefore, we scale the Chamfer distances in the evaluations to match the other SMPL models, to make them comparable.
		 $\uparrow (\downarrow)$ means the higher (lower) the better.
	 }
\label{tab:ae}       %
\setlength{\tabcolsep}{4pt}
\begin{tabular}{cc|ccc}
\hline\noalign{\smallskip}
Dataset & Model & $d_C(P,\widehat{P}) \downarrow$ & $\mathcal{C}_{\text{3D}} \downarrow$  & $\mathcal{C}_X \uparrow$ \\
\noalign{\smallskip}\hline\noalign{\smallskip}
\multirow{2}{*}{MNIST} &
        STD-U & 1.19 & 1.57  & 0.92 \\
 $\,$ & FTL-U & 0.94 & 2.73  & 0.65 \\
 \noalign{\smallskip}\hline\noalign{\smallskip}
\multirow{4}{*}{SMAL} &
 		STD-S & 0.35 & 0.03   & 0.97 \\
 $\,$ & FTL-S & 0.10 & 0.14  & 0.93 \\
 $\,$ & STD-U & 0.29 & 0.01  & 0.97 \\ 
 $\,$ & FTL-U & 0.10 & 0.21  & 0.88 \\
  \noalign{\smallskip}\hline\noalign{\smallskip}
\multirow{4}{*}{SMPL} &
 		STD-S & 0.34 & 0.03  & 0.97 \\
 $\,$ & FTL-S & 0.19 & 0.30  & 0.71 \\
 $\,$ & STD-U & 0.23 & 0.05  & 0.97 \\ 
 $\,$ & FTL-U & 0.18 & 0.45  & 0.70 \\
  \noalign{\smallskip}\hline\noalign{\smallskip}
\multirow{4}{*}{HA} &
 		STD-S 	& 0.36/0.44 & 0.03/0.05  & 0.97/0.97 \\
 $\,$ & FTL-S 	& 0.11/0.19 & 0.24/0.22  & 0.66/0.62 \\
 $\,$ & STD-U 	& 0.33/0.34 & 0.02/0.06  & 0.97/0.97 \\
 $\,$ & FTL-U 	& 0.11/0.19 & 0.20/0.20  & 0.72/0.66 \\
\noalign{\smallskip}\hline
\end{tabular}
\end{table}

\subsection{Autoencoder Results}
\label{sec:aeresults}

\begin{figure}
	\centering
	\includegraphics[width=0.237\textwidth]{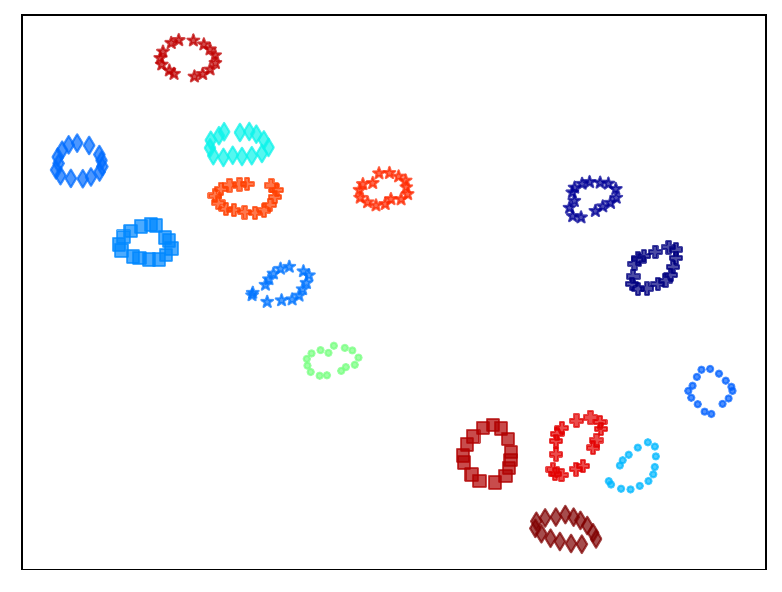}
	\includegraphics[width=0.237\textwidth]{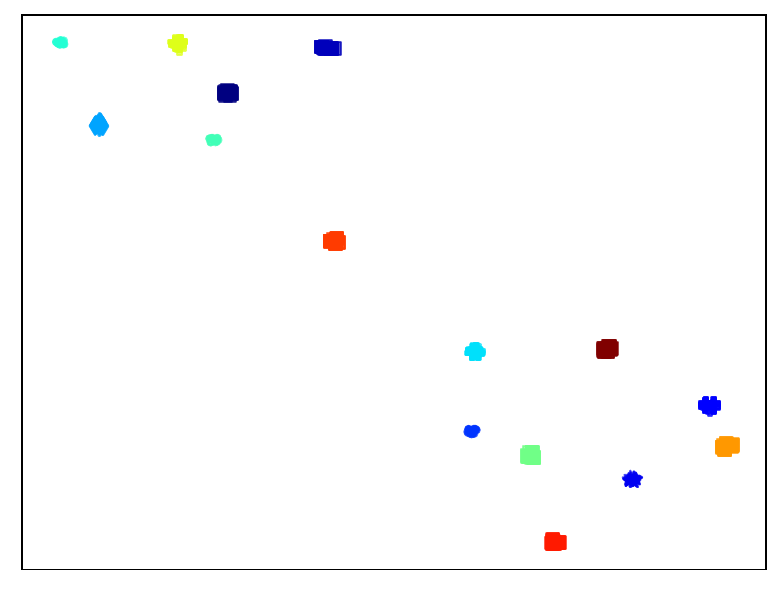}
	\caption{ 
	    Latent rotational invariance via t-SNE plots of $x_c$ vector embeddings.
		Plots are done with the HA-FTL-U (left) and HA-STD-U (right) models, respectively (see Table \ref{tab:ae}).
		Colors denote a single shape; 
		markers with the same color are rotations of that single shape.
		Marker types are only meant to help tell apart shapes with similar colors.
		Notice the non-FTL (STD) architecture gives a tighter latent invariance.
	}
	\label{fig:aetsne} %
\end{figure}

Our AE is designed to factorize out rigid pose, as well as encode a complete representation of a canonical shape.
In Fig.\ \ref{fig:aerecon}, 
	we show example reconstructions, 
	as well as the canonicalization capability of the model.
In Fig.\ \ref{fig:aetsne}, 
	we show latent embeddings of the shape representations $x_c$ across different rotations of input shapes.
The results show that the AE is not only able to accurately reconstruct the inputs, 
	but also correctly derotate the canonical PCs in 3D, 
	and that the encodings are close to being orientation invariant in the latent space.

\newer{
We consider two AE types, the STD and FTL models with their differing rotation handling techniques. We also examine two ablations: the unsupervised (U) scenario, which removes the assumption of aligned data, and the HA-trained model, which eliminates the use of specialized single models for SMAL and SMPL.
}

Quantitatively,
we evaluate our autoencoders on (1) reconstruction capability and (2) rotation invariance in their representation.
Reconstruction quality is computed with the standard Chamfer distance between the output PC and a uniform random sampling from the raw shape mesh.
We average over five randomly rotated copies of the test set.

Rotation invariance is assessed with two measures.
The first is in 3D space, and checks that canonicalizations of the same PC under different rotations are close 
	(according to the Chamfer distance between PCs):
\begin{equation}
	\mathcal{C}_{\text{3D}} = \frac{1}{M_R} \sum_{i=1}^{m_R} \sum_{j\ne i} d_C\left( D(E(P_i)), P_j \right),
\end{equation}
where $m_R$ is the number of random copies we use for evaluation and $M_R$ is the number of pairs tested.

The second measure is in the latent canonical shape space (i.e., $x_c$). 
Since latent distances are less meaningful (e.g., dimensions may have very different scales) and will differ across AEs, 
	we choose to measure performance by \textit{clustering quality}. 
Ideally, a representation that canonicalizes an input shape should map rotated copies of a given PC to the same latent encoding -- exactly fulfilling this would make it \textit{rotation invariant}.
Hence, we create rotated copies of many input shapes, encode them, 
	and then cluster in the AE embedding space. 
We expect that rotated copies of the same instance should cluster together; 
	hence, we treat instance identity as a ground truth cluster label and use 
	Adjusted Mutual Information (AMI) 
	to measure quality \citep{vinh2010information}.
An AMI of 1 indicates perfect matching of the predicted and real partitions, 
	while an AMI of 0 is the expected value of a random clustering.
We average AMIs over clusterings obtained from different random sample sizes 
(i.e., the number of unique shapes duplicated and clustered).
The resulting ``area-under-the-curve''-like latent space clustering metric for rotational invariance is denoted $\mathcal{C}_X$.
See Appendix \S\ref{app:CX} for additional details.

The original GDVAE model 
\citep{aumentado2019geometric}
was trained on limited angles of rotation about the canonical one, 
since otherwise reconstruction quality was degraded but
in this work we always consider full rotation about a single axis.
Despite the fact that two models use essentially the same  architectural components,
our AE is better able to obtain canonical orientations, while maintaining reconstruction quality.

The results in Table \ref{tab:ae} show a few patterns between the AE types\footnote{\new{We remark that these results utilize single-axis (planar) rotations; we refer the reader to Appendix \S\ref{app:fullrot} for tests with full rotations, which results in reduced rotational robustness.}}.
First, 
    we find the that the FTL-based AE has superior reconstruction quality,
    while the STD AE has much better rotation invariance.
Second,
    the difference between the unsupervised and supervised scenarios is relatively smaller, 
    with the unsupervised reconstruction quality being slightly better than the supervised, 
    whereas the supervised case has superior rotation invariance.
Finally, 
    performance on the HA dataset 
    (which is a union of the SMAL and SMPL data) 
    is only slightly degraded compared to the per-category models 
    (moreso for FTL than STD).

\subsubsection{Results Summary}
\newer{
Since the FTL-based AE maintains strong rotation invariance, 
    with superior latent interpretability and reconstruction error, 
    we suggest using it as a starting point.
We also find that rotation factorization can be done without aligned data supervision, at little cost to reconstruction or rotational invariance quality.
}

\begin{figure}
	\centering
	\includegraphics[width=0.47\textwidth,draft=false]{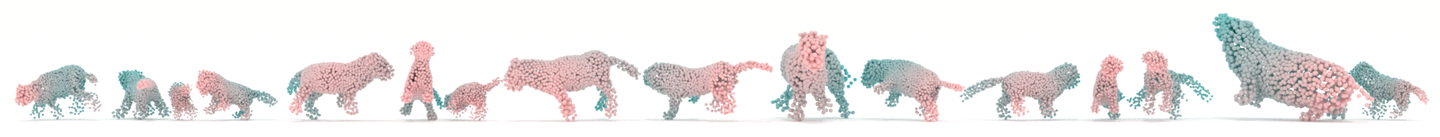}	
	\includegraphics[width=0.47\textwidth,draft=false]{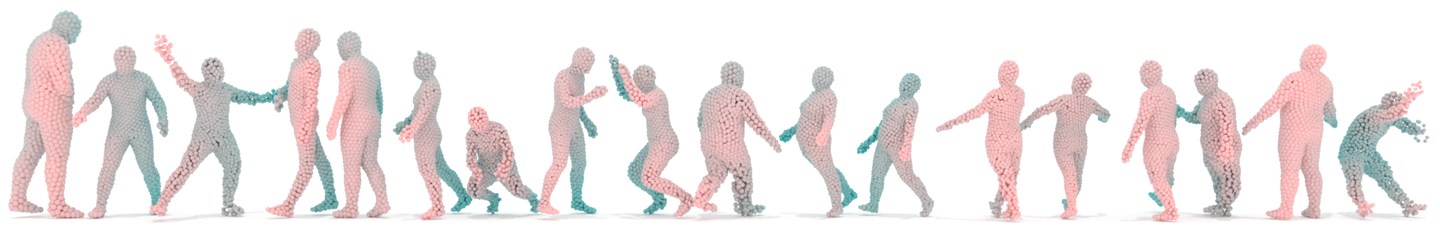}
	\includegraphics[width=0.47\textwidth,draft=false]{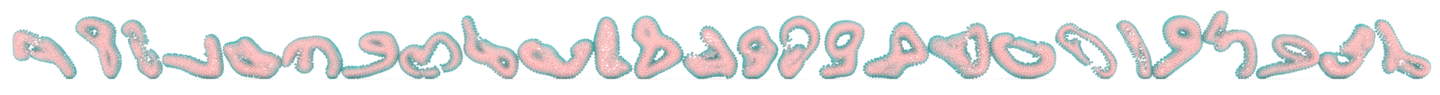}	
	\includegraphics[width=0.47\textwidth,draft=false]{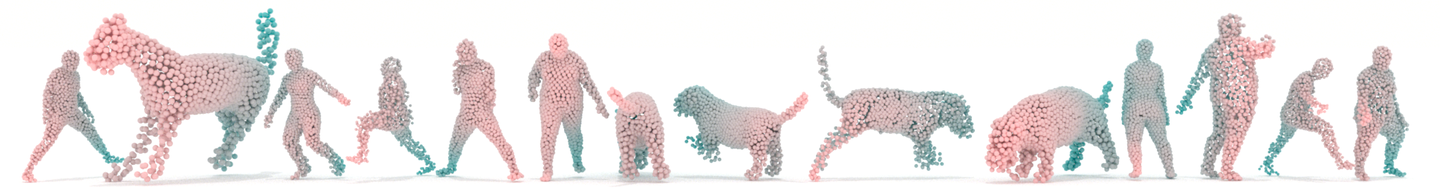}	
	\caption{ 
		Random sample generations from the VAE.
		All models use the FTL-based AE.
		Note that the MNIST model was trained on digits at all orientations %
		and thus should output samples at any orientation (as for SMAL and SMPL).
		Rows: SMAL, SMPL, MNIST, and HA.
	}
	\label{fig:vaegenerative} %
\end{figure}

\begin{figure*}
	\centering
	\includegraphics[width=0.241\textwidth]{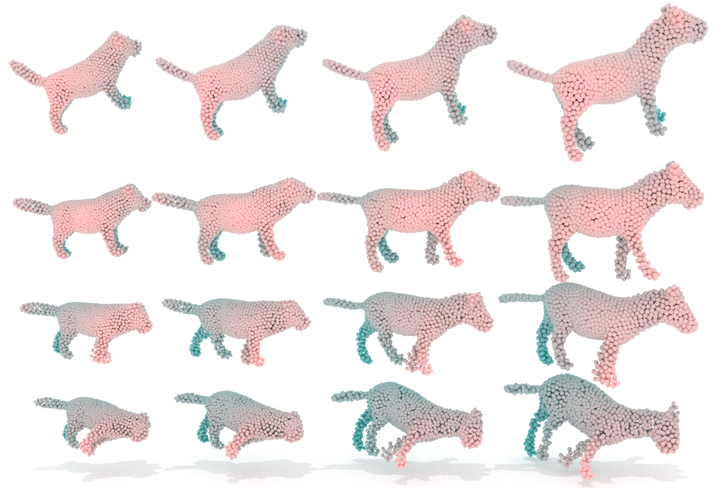}\hspace{\fill}%
	\includegraphics[width=0.241\textwidth]{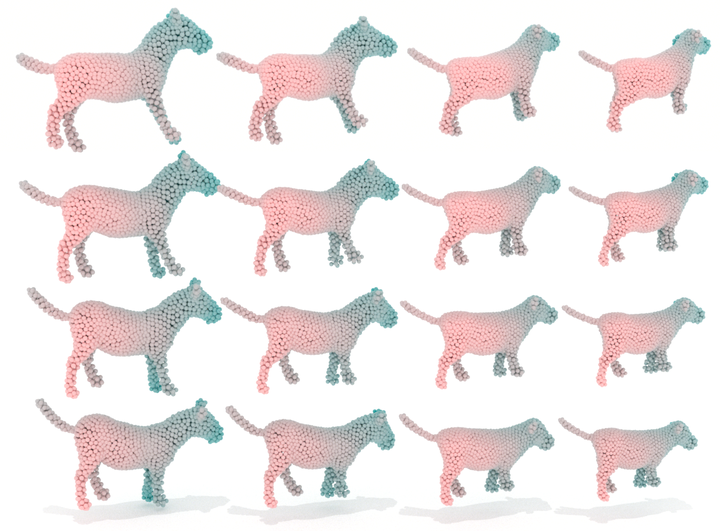}\hspace{\fill}%
	\includegraphics[width=0.241\textwidth]{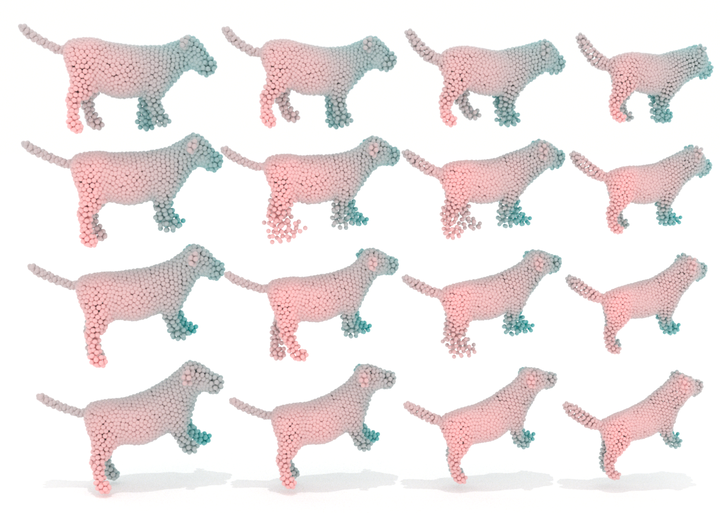}\hspace{\fill}%
	\includegraphics[width=0.241\textwidth]{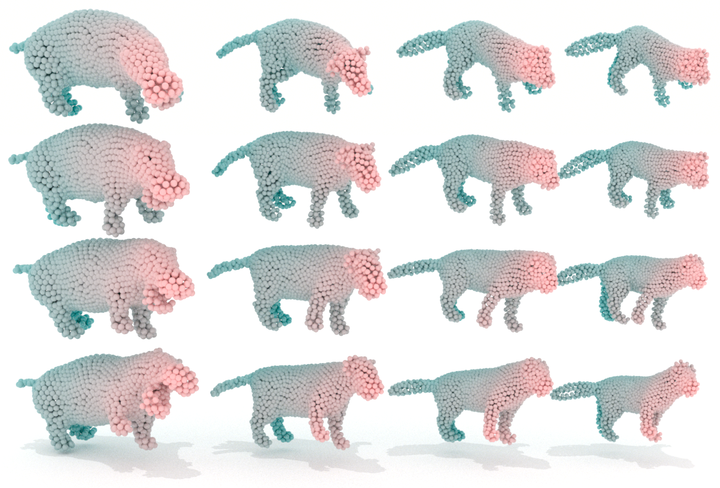}\\
	\includegraphics[height=0.36\textwidth]{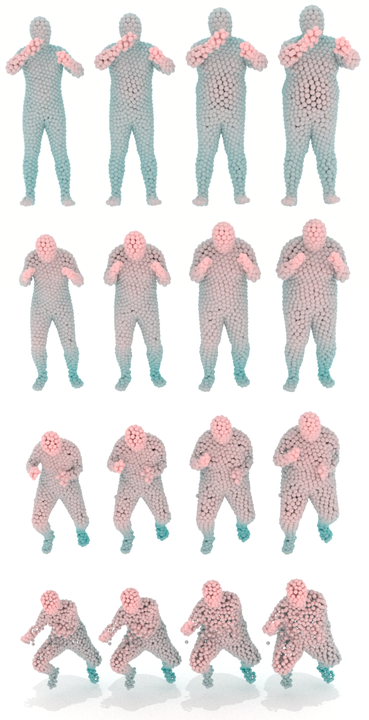}\hspace{\fill}%
	\includegraphics[height=0.36\textwidth]{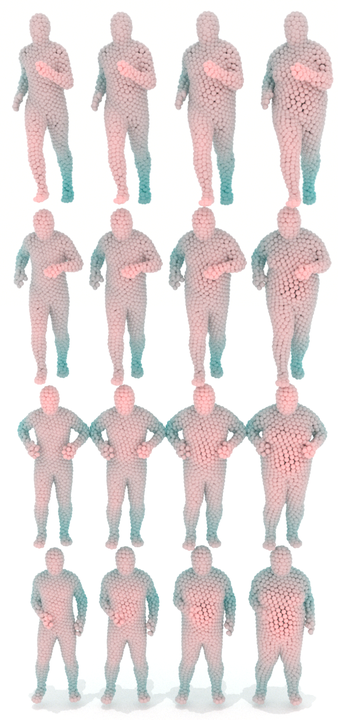}\hspace{\fill}%
	\includegraphics[height=0.36\textwidth]{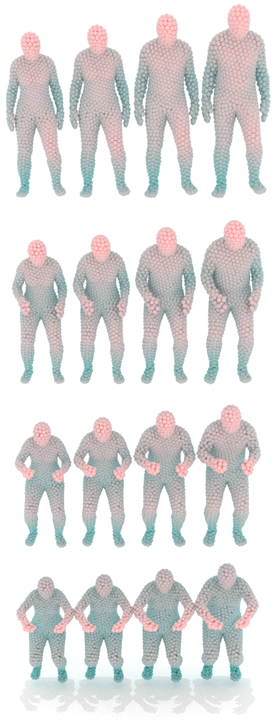}\hspace{\fill}%
	\includegraphics[height=0.36\textwidth]{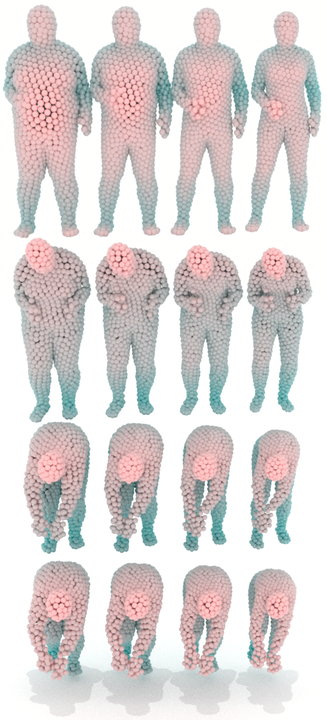}\hspace{\fill}%
	\includegraphics[height=0.36\textwidth]{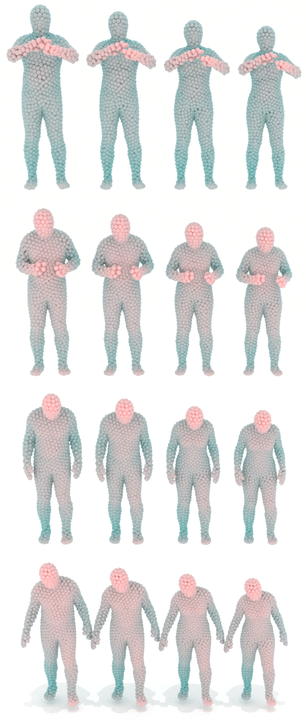}\hspace{\fill}%
	\includegraphics[height=0.36\textwidth]{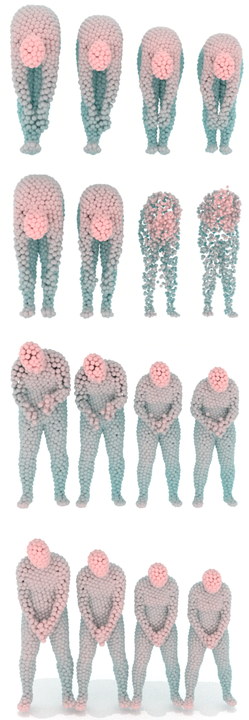}\\
	\includegraphics[height=0.19\textwidth]{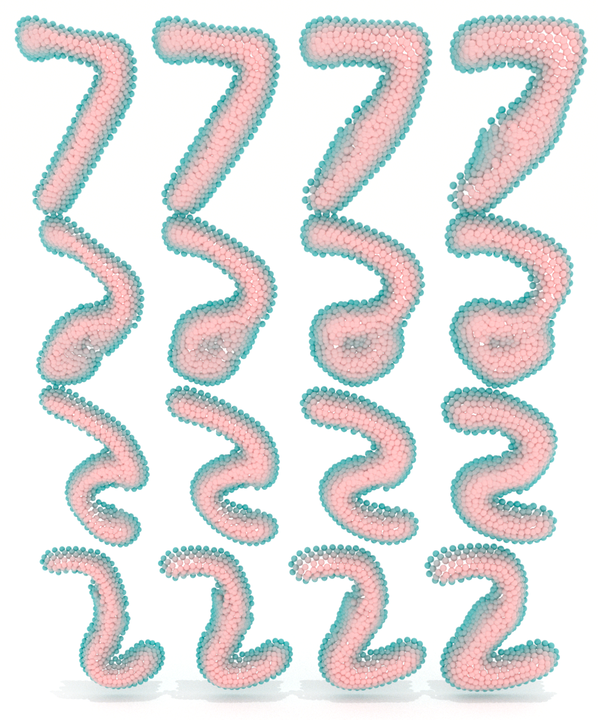}
	\includegraphics[height=0.19\textwidth]{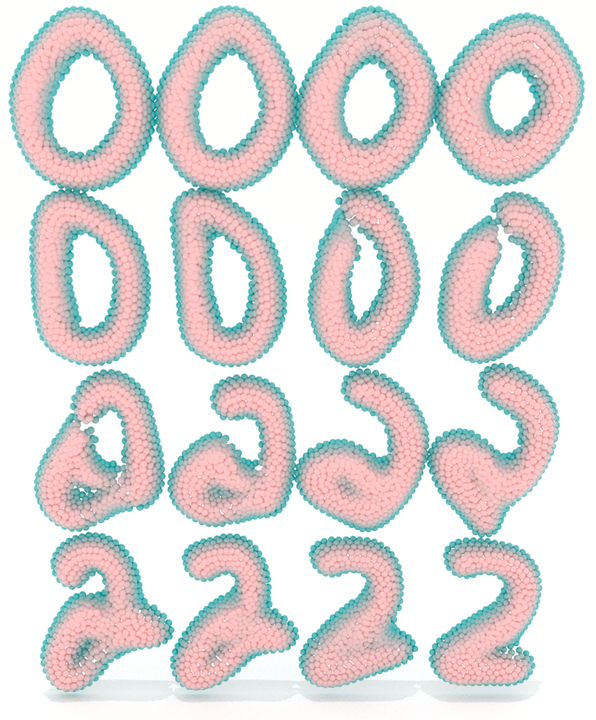}
	\includegraphics[height=0.19\textwidth]{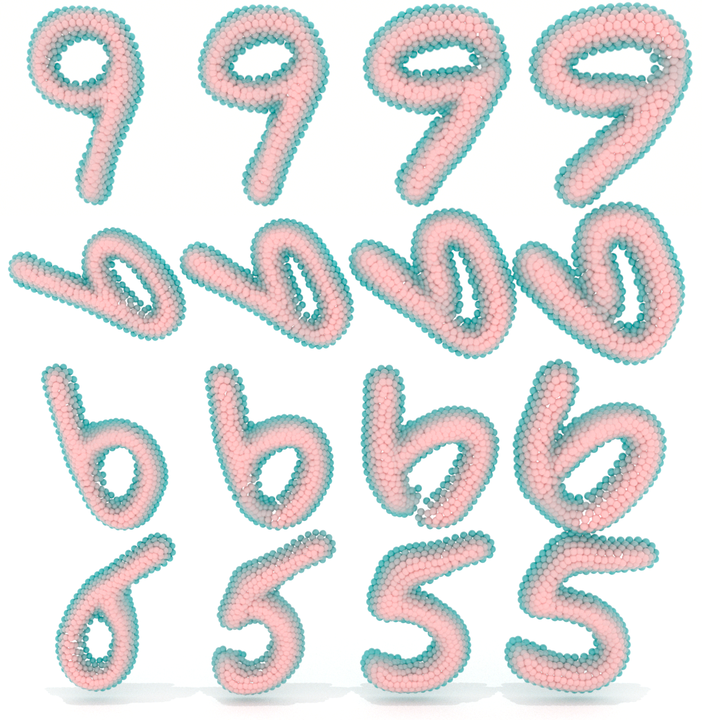}
	\includegraphics[height=0.19\textwidth]{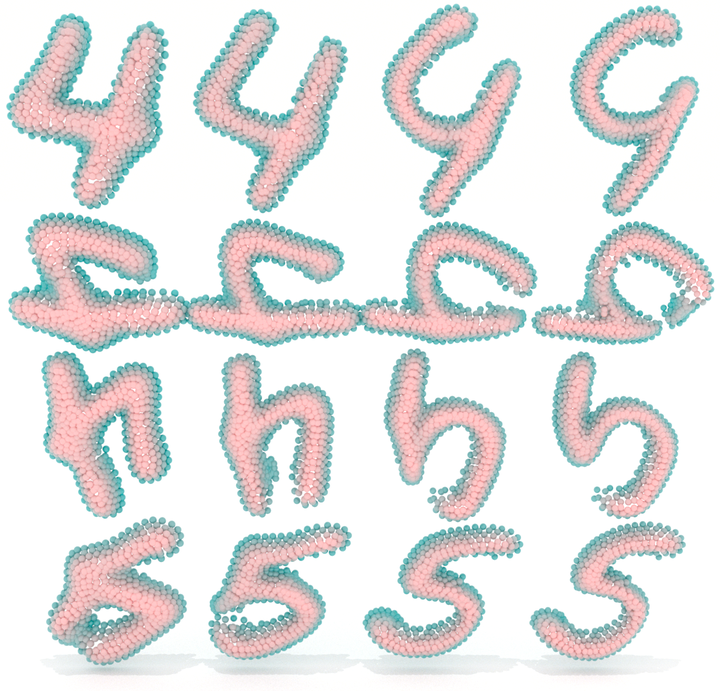}
	\includegraphics[height=0.19\textwidth]{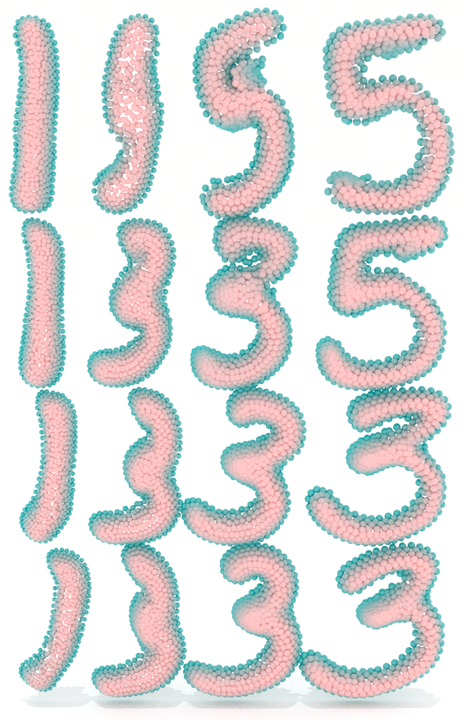}
	\includegraphics[height=0.19\textwidth]{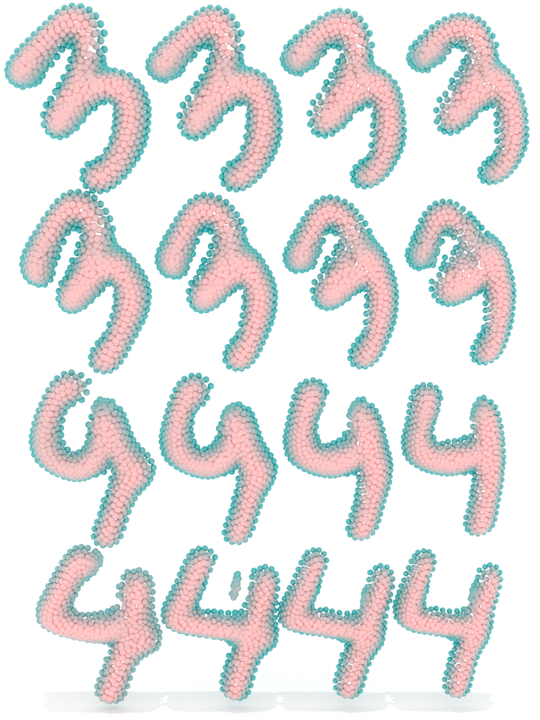}
	\caption{ 
		Intrinsic-extrinsic latent space interpolations using the disentangled VAE representation.
		In each $4\times 4$ panel,
		top-left and bottom-right shapes are reconstructions of real inputs.
		Along the horizontal axis, we interpolate along the intrinsics ($z_I$), 
		whereas we do so for the extrinsics ($z_E$) along the vertical axis.
		Notice that the bottom-left and top-right shapes are \textit{pose transfers},
		for which one of the latent factors is exchanged, while the other remains unchanged.
		SMAL and SMPL shapes are shown in the learned canonical orientation (using FTL-S);
		for MNIST only (using FTL-U), 
		we interpolate the estimated AE rotation encodings $q$ as well, 
		via slerping the quaternions \citep{shoemake1985animating} 
		between the two inputs along with the extrinsics (i.e., along the vertical axis).
	}
	\label{fig:vaeinterp} %
\end{figure*}

\begin{figure*}
	\centering
	\begin{tabular}{c|c|c}
		Query & Extrinsic Retrievals (via $z_E$) & Intrinsic Retrievals (via $z_I$) \\
		\includegraphics[width=0.06\textwidth,draft=false]{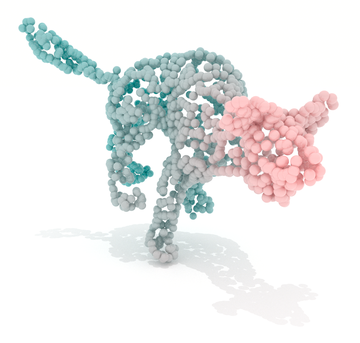}
		&
		\includegraphics[width=0.43\textwidth,draft=false]{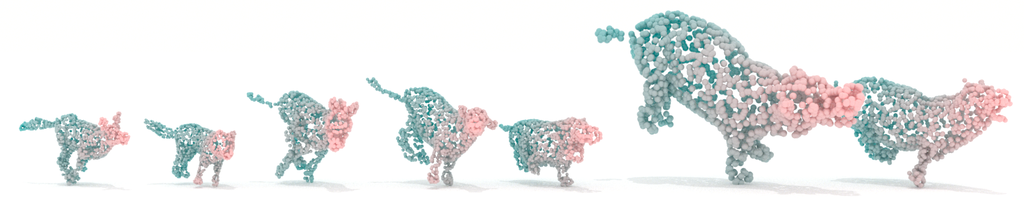}
		&
		\includegraphics[width=0.43\textwidth,draft=false]{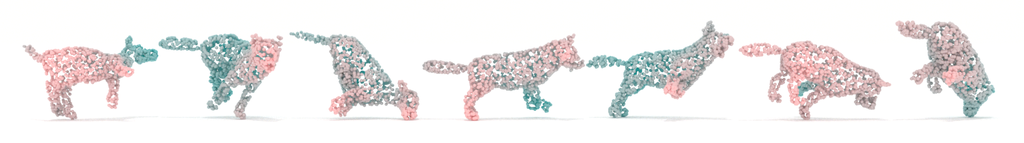} \\
		\includegraphics[width=0.06\textwidth,draft=false]{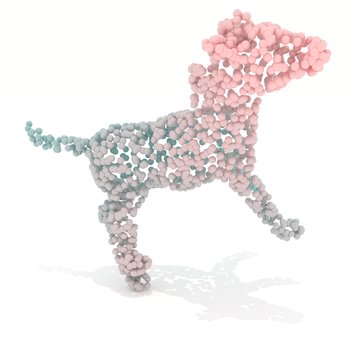}
		&
		\includegraphics[width=0.43\textwidth,draft=false]{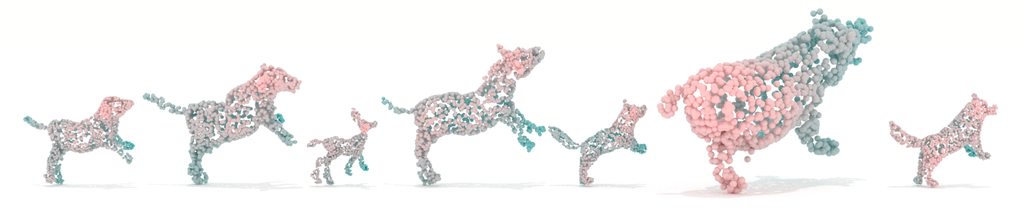}
		&
		\includegraphics[width=0.43\textwidth,draft=false]{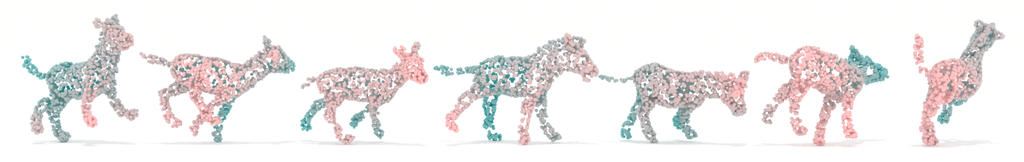} \\
		\includegraphics[width=0.045\textwidth,draft=false]{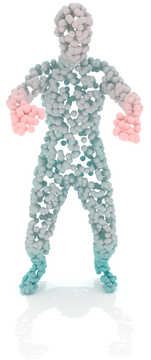}
		&
		\includegraphics[width=0.41\textwidth,draft=false]{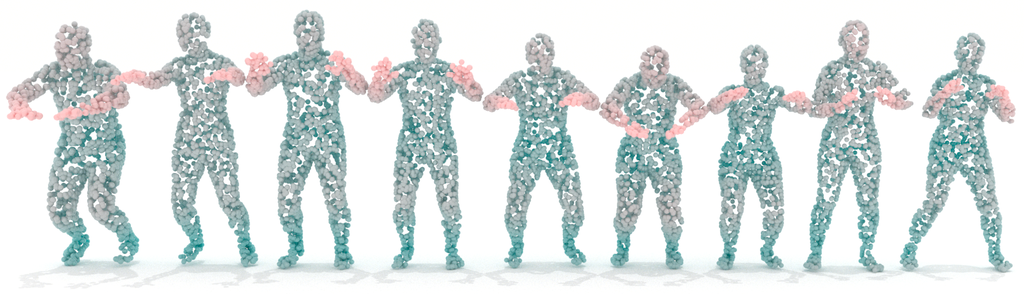}
		&
		\includegraphics[width=0.43\textwidth,draft=false]{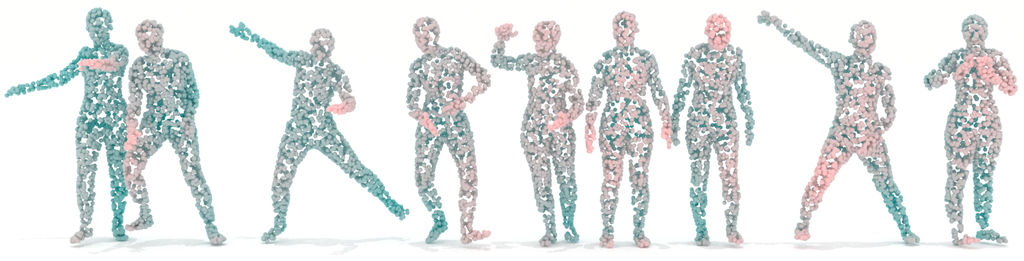} \\
		\includegraphics[width=0.055\textwidth,draft=false]{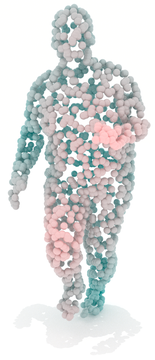}
		&
		\includegraphics[width=0.39\textwidth,draft=false]{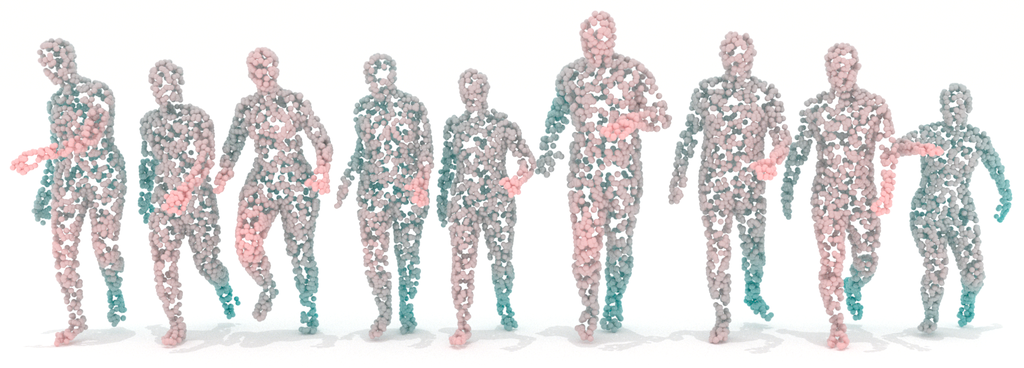}
		&
		\includegraphics[width=0.43\textwidth,draft=false]{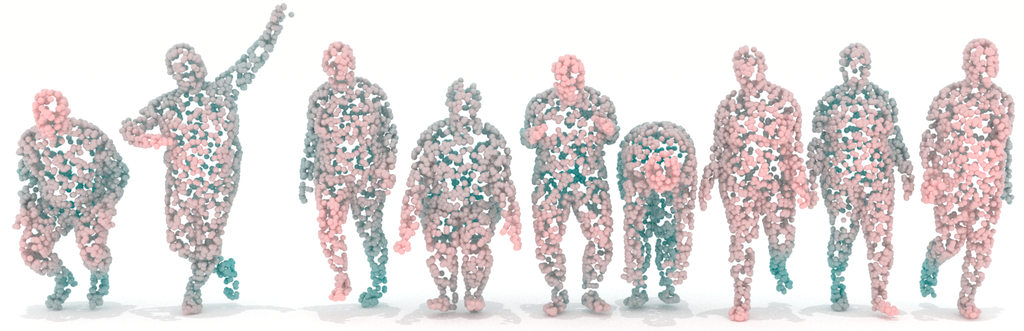} \\
		\includegraphics[width=0.045\textwidth,draft=false]{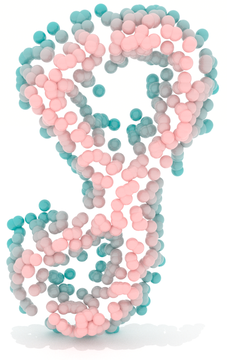}
		&
		\includegraphics[width=0.39\textwidth,draft=false]{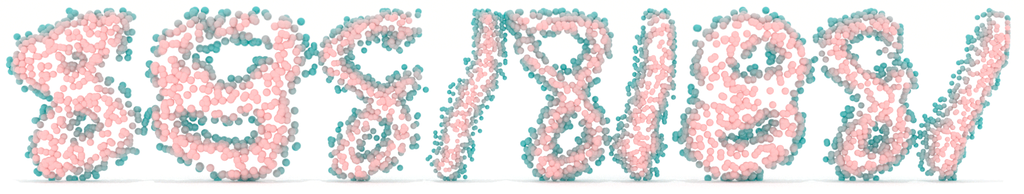}
		&
		\includegraphics[width=0.41\textwidth,draft=false]{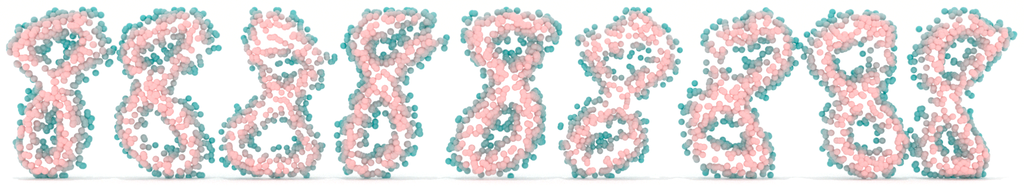} \\
		\includegraphics[width=0.055\textwidth,draft=false]{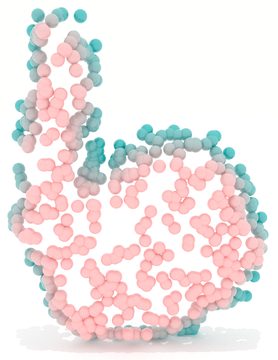}
		&
		\includegraphics[width=0.41\textwidth,draft=false]{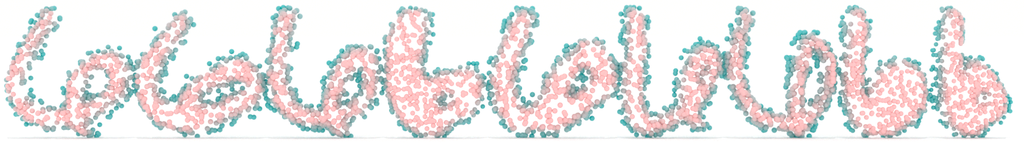}
		&
		\includegraphics[width=0.41\textwidth,draft=false]{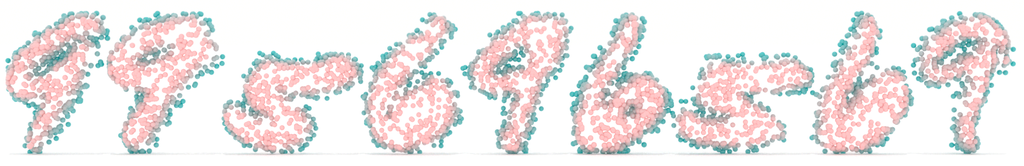} \\
	\end{tabular}
	\caption{ 
		Example retrievals using the disentangled factors of the GDVAE{\PPP}\  latent representation.
		Shapes are in order of latent similarity from left to right).
		SMAL, SMPL, and MNIST use FTL-S, FTL-S, and FTL-U AE models, respectively.
		For SMAL and SMPL, notice that $z_E$ retrieves a variety of animals/body types in the same articulated pose, while $z_I$ retrieves the same animal/body type in an array of different non-rigid poses.
		Though MNIST does not have a natural sense of articulation, 
			notice that the extrinsic retrievals tend to have the same digit identity, 
			but vary most noticeably in thickness (which is a non-isometric alteration).
			In contrast, digits retrieved via $z_I$ appear to be bent largely isometrically; 
				that is, ``wiggled'' around in a way that preserves the metric tensor 
				(and the distribution of geodesic distances among points) -- see the ``8'' digit.
			For the ``6'', notice that it retrieves several ``9'' digits, 
				showing its blindness to orientation, as well as two ``5'' shapes 
					that were ``thickened'' sufficiently in a manner similarly to the query.
	}
	\label{fig:vaeretrieval} %
\end{figure*}

\begin{figure}
    \centering
    \includegraphics[width=0.235\textwidth]{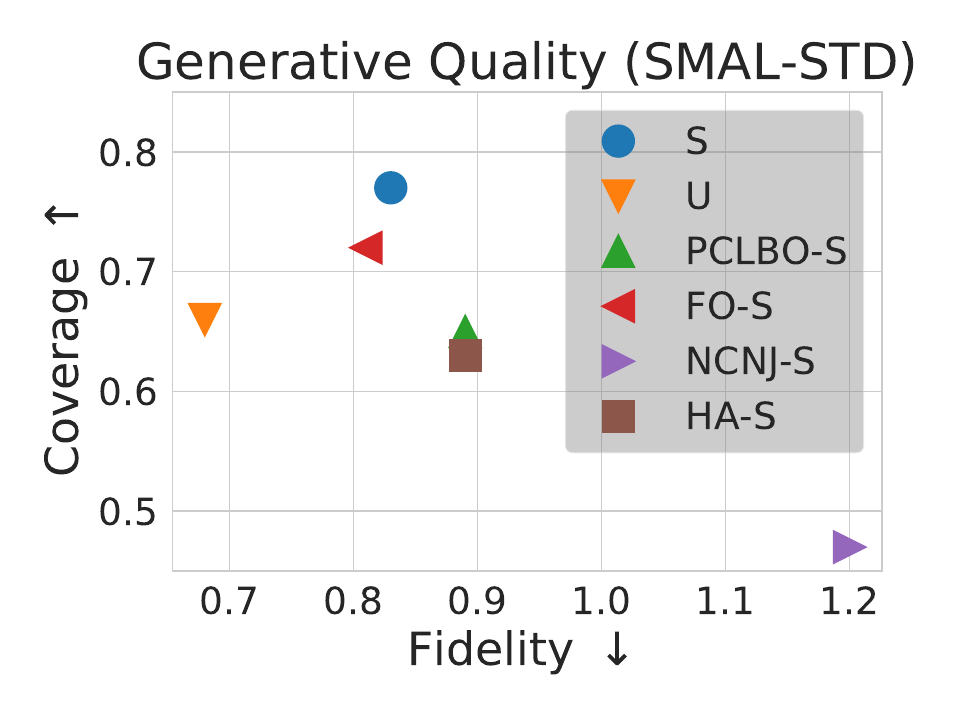}
    \includegraphics[width=0.235\textwidth]{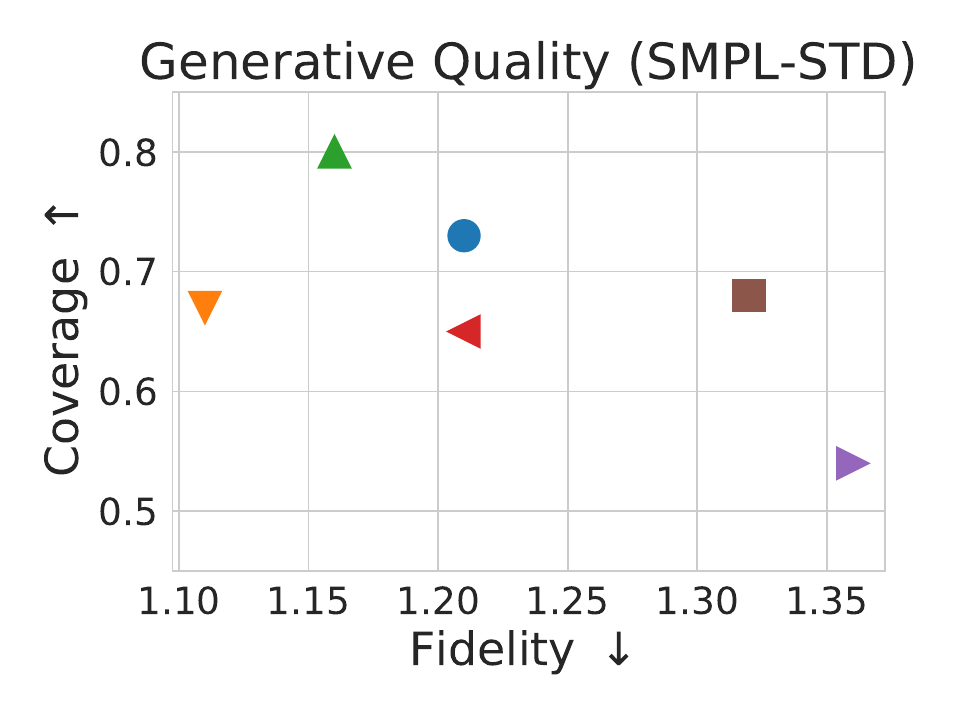} \\
    \includegraphics[width=0.235\textwidth]{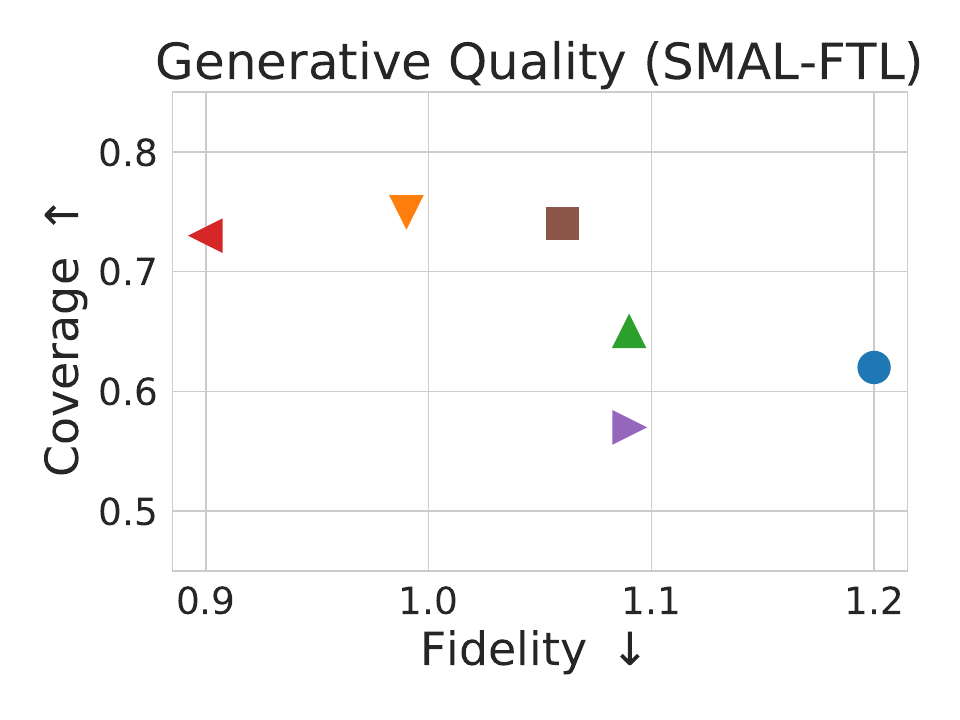}
    \includegraphics[width=0.235\textwidth]{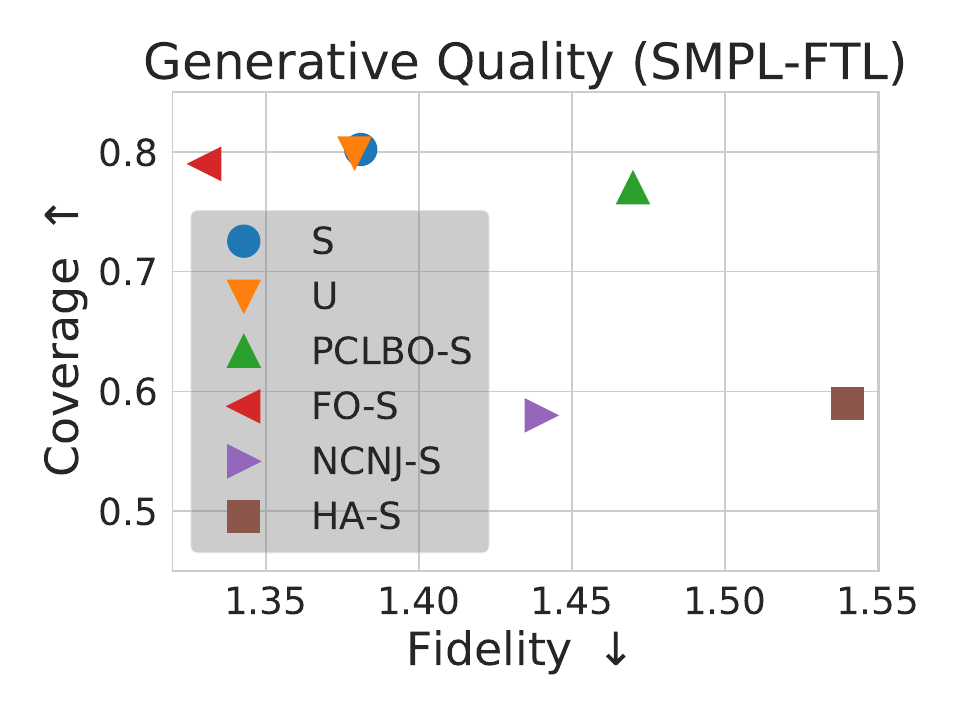}
    \caption{
        \new{
            VAE generative modelling evaluation scores on 
                SMAL (left) and SMPL (right) with the FTL AE architecture.
            Notationally,
            ``S''/``U'' refers to the supervised/unsupervised AE settings 
            (see \S\ref{sec:ae:loss:std}), 
            PCLBO refers to the LBOS being procured from a point cloud
            (see \S\ref{sec:srobust}), 
            NCNJ refers to our disentanglement ablation 
            (see \S\ref{sec:ablations}), 
            and HA refers to a model trained on both SMAL and SMPL at the same time
            (see \S\ref{sec:vaeresults}).
            All methods use the GDVAE\PP\ setting 
                (see \S\ref{sec:trainregimes}), 
                except for the FO case.
            The upper-left corner of the plots is preferred.
            See %
                Appendix Table \ref{tab:vaemass} for detailed values.
            Notice that NCNJ tends to have poor generative quality, 
                while the GDVAE\PP\ (S or U) and GDVAE-FO generally perform well.
            In most cases, HA underperforms the GDVAE\PP\  (except for the SMAL-FTL case),
                likely due to the additional complexity of the dataset straining model capacity. 
        }
    }
    \label{fig:t2f-genqual}
\end{figure}

\begin{figure}
    \centering
    \includegraphics[width=0.235\textwidth]{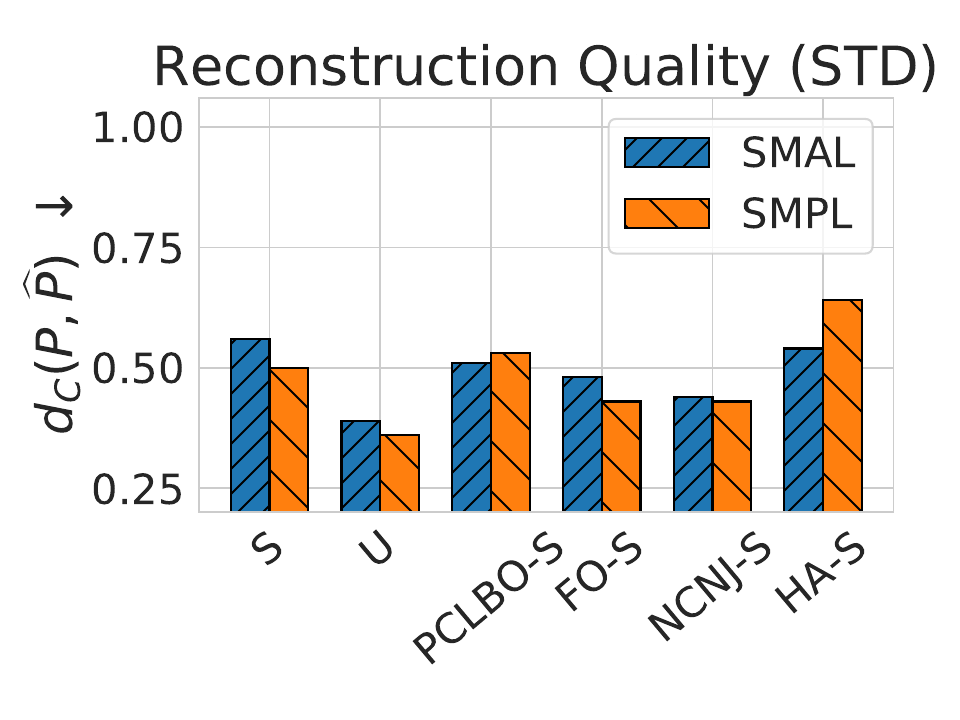}
    \includegraphics[width=0.235\textwidth]{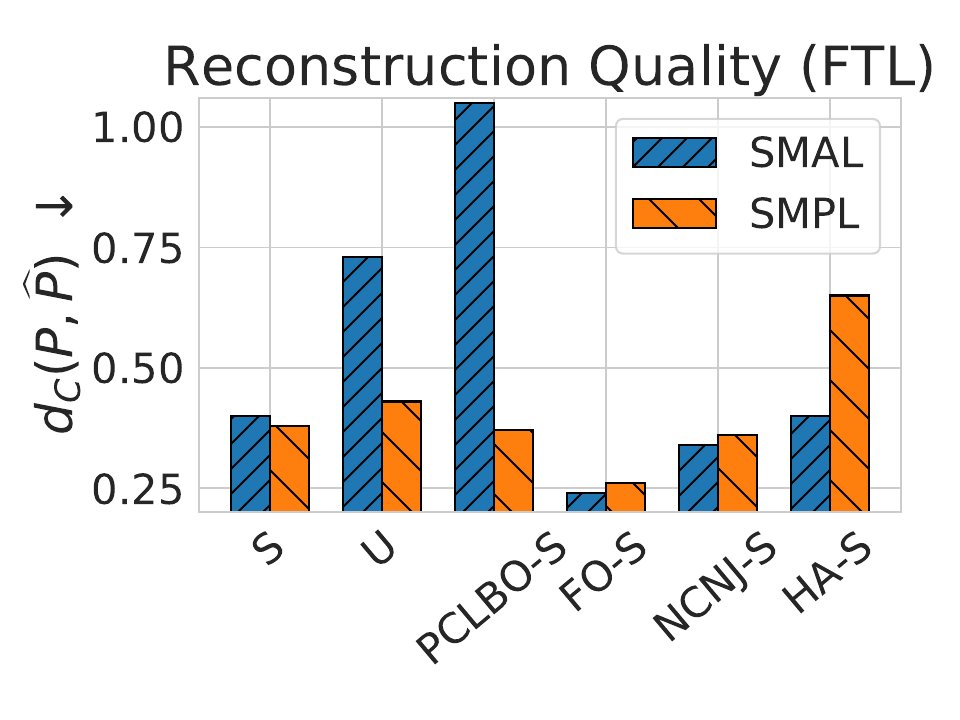}
    \caption{
        \new{
            VAE reconstruction quality evaluation, measured in Chamfer distance (lower is better), using the STD (left) and FTL (right) architectures.
            See Fig.\ \ref{fig:t2f-genqual} for explanation of model types and Appendix Table \ref{tab:vaemass} for detailed values.
            We find that the PCLBO and HA models are similar or worse, while NCNJ and FO are similar or slightly better, compared to GDVAE\PP\ (S). Note that NCNJ can take advantage of the weaker disentanglement requirements, while the FO case simply fails to disentangle (see Fig.\ \ref{fig:t2f-disent}).
        }
    }
    \label{fig:t2f-recon}
\end{figure}

\begin{figure}
    \centering
    \includegraphics[width=0.235\textwidth]{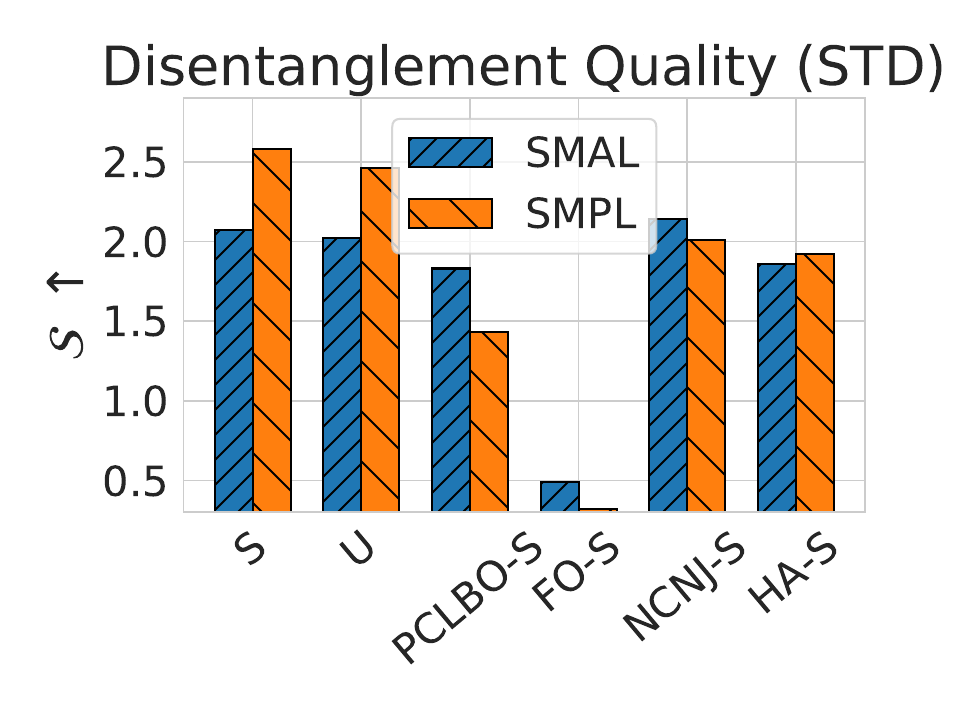}
    \includegraphics[width=0.235\textwidth]{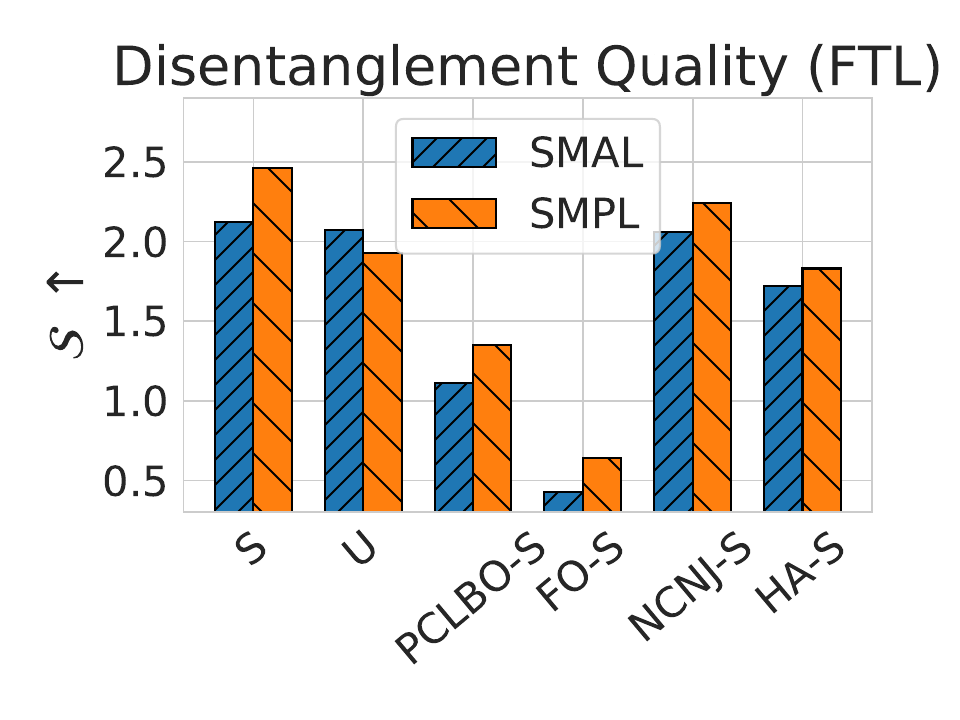}
    \caption{
        \new{
            VAE disentanglement quality evaluation 
                (see \S\ref{sec:ret} and Eq.\ \ref{eq:disentscore} for discussion of the disentanglement score $\mathcal{S}$).
            See Fig.\ \ref{fig:t2f-genqual} for explanation of model types and Appendix Table \ref{tab:vaemass} for detailed values.
            Compared to the regular GDVAE\PP, 
                the flow-only (FO) case is severely degraded,
                whereas PCLBO, HA, and NCNJ experience moderate deterioration (for the latter case, moreso on SMPL than SMAL).
        }
    }
    \label{fig:t2f-disent}
\end{figure}

\begin{figure}
    \centering
    \includegraphics[width=0.235\textwidth]{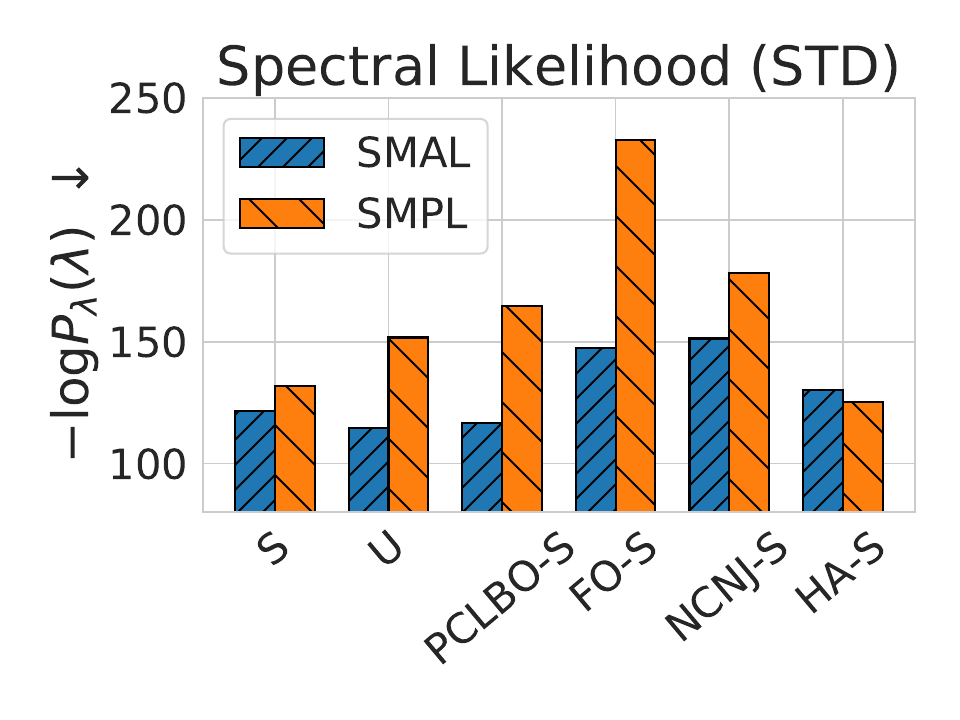}
    \includegraphics[width=0.235\textwidth]{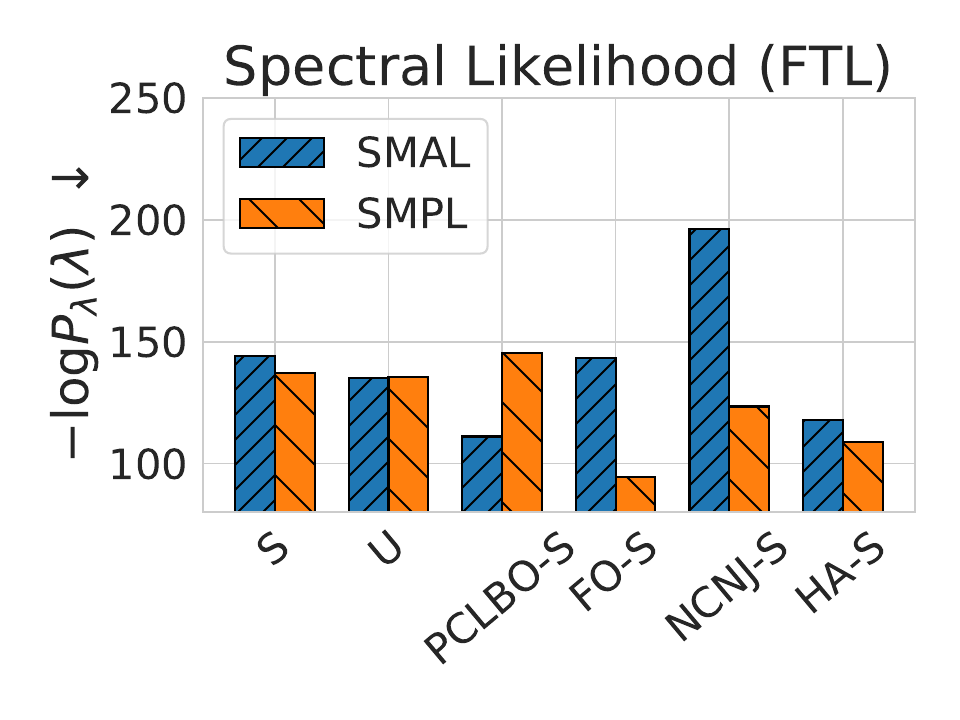}
    \caption{
        \new{
            VAE spectral negative log-likelihood (NLL) evaluation, measuring generative quality of the diffeomorphic flow network on the LBOS 
            (see \S\ref{sec:flowlikeloss}).
            See Fig.\ \ref{fig:t2f-genqual} for explanation of model types and Appendix Table \ref{tab:vaemass} for detailed values.
            In most scenarios, NCNJ experiences some degradation, while NLL in the FO case increases only for the STD case.
        }
    }
    \label{fig:t2f-spectrallikelihood}
\end{figure}

\begin{figure*}
    \centering
    \includegraphics[width=0.244\textwidth]{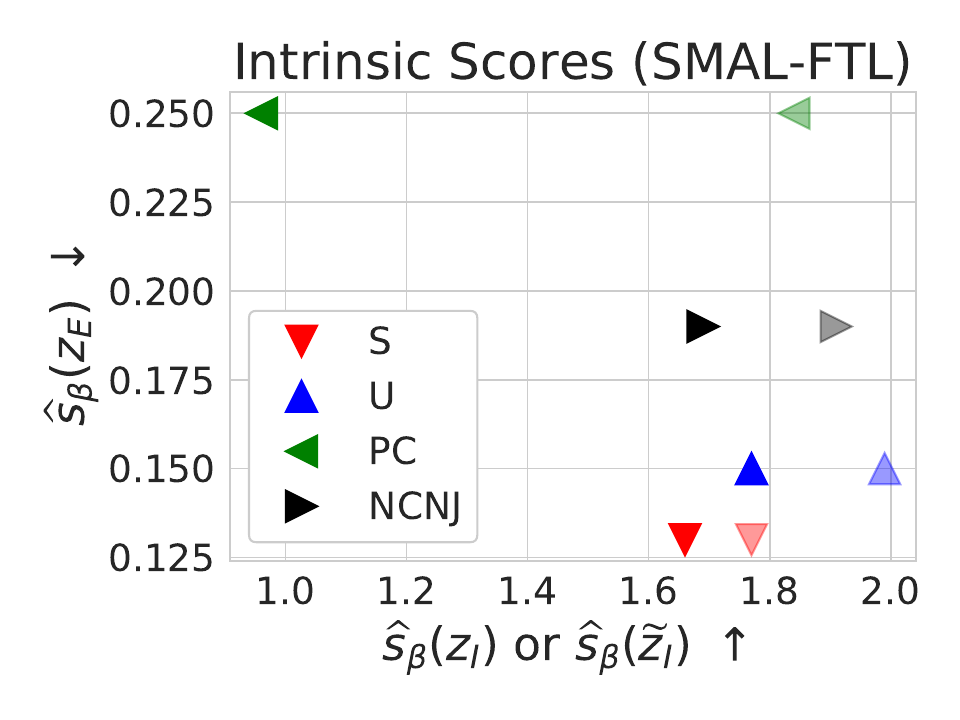}
    \includegraphics[width=0.244\textwidth]{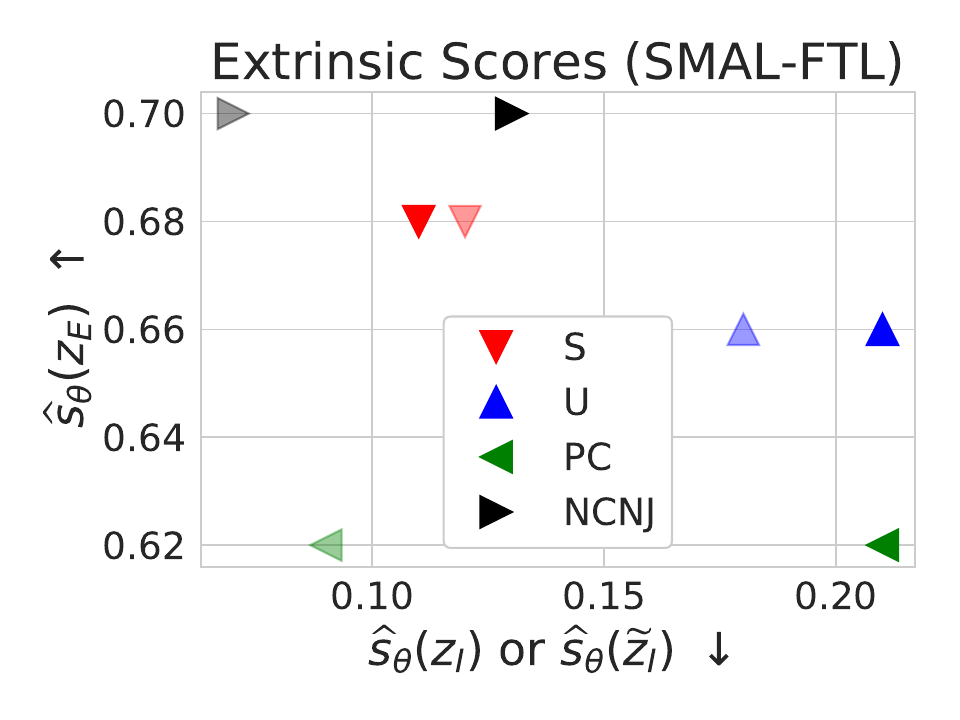}
    \includegraphics[width=0.244\textwidth]{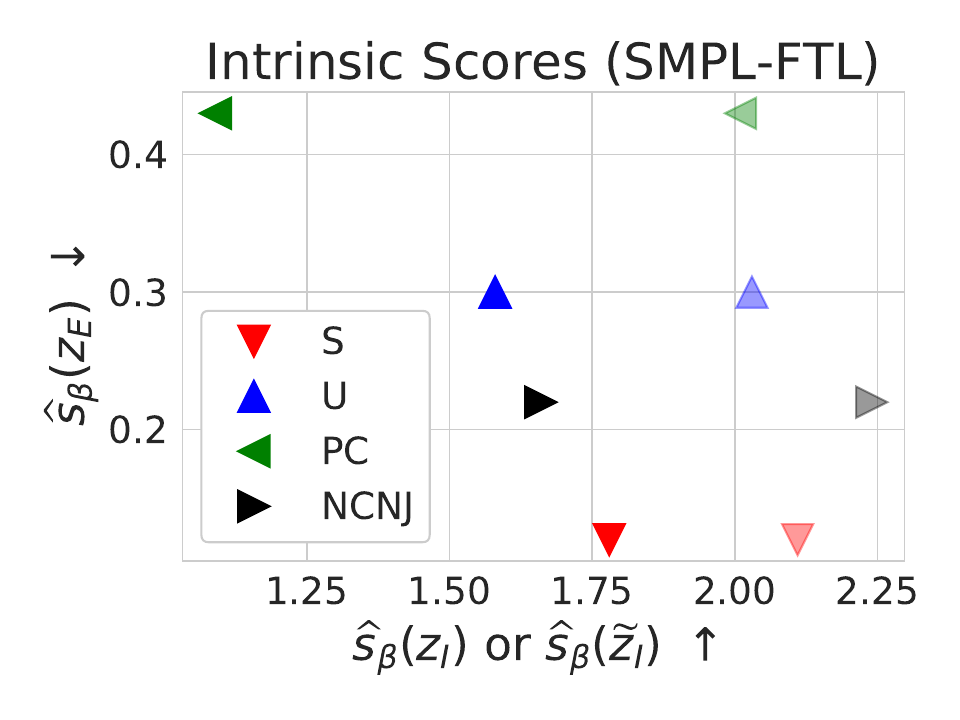}
    \includegraphics[width=0.244\textwidth]{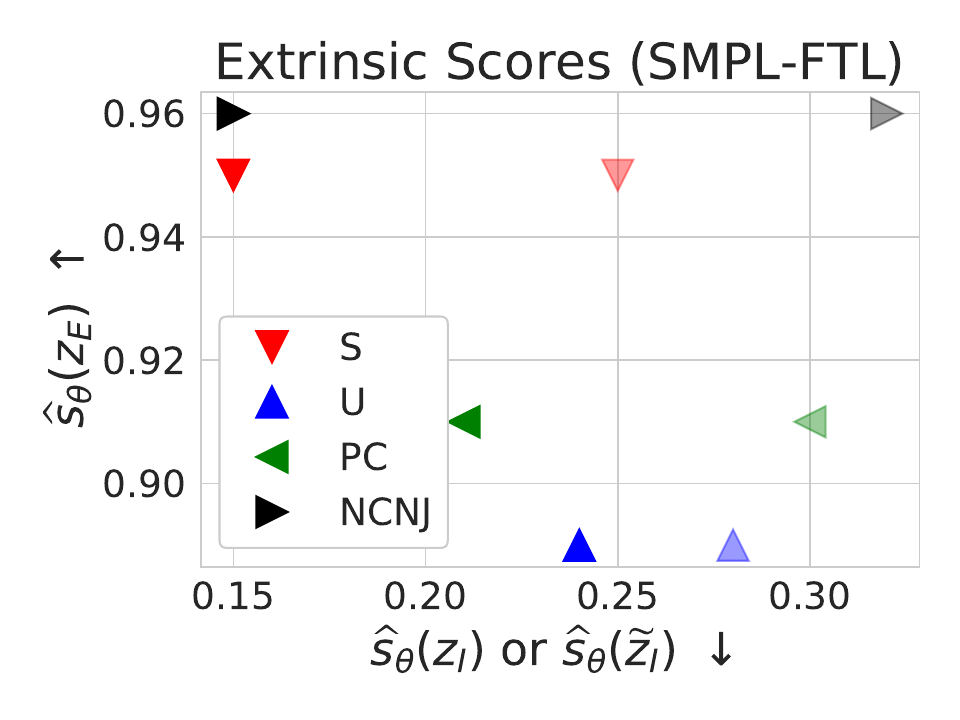}
    \caption{
        \new{
            Pose-aware retrieval scores with the FTL AE model.
            Model notation refers to the GDVAE\PP\ model with (S) or without (U) rotation supervision, use of the PC-derived LBOS (PC; see \S\ref{sec:srobust}), and the partial disentanglement loss ablation (NCNJ; see \S\ref{sec:ablations}).
            The lighter (partially transparent) counterparts of each point corresponds to using $\widetilde{z}_I = f_\lambda(\lambda)$ instead of $z_I$ for retrieval.
            The leftmost two insets show results with SMAL, while the rightmost two do so for SMPL; for each set of two, left and right correspond to scores in intrinsic ($\widehat{s}_\beta$) and extrinsic ($\widehat{s}_\theta$) retrieval, respectively.
            Preferred values lie in the bottom-right for intrinsic scores and top-left for extrinsic scores.
            See Appendix Fig.\ \ref{table:t2f-retstd} for plots with the STD AE and Appendix Table \ref{tab:ret} for detailed values.
            We see that 
            (i) using $\lambda$-derived (rather than mesh-derived) latents is consistently better for intrinsic retrieval, but more mixed for extrinsic retrieval (particularly for SMAL);
            (ii) the PCLBO struggles more on intrinsic than extrinsic retrieval scores (see also \S\ref{sec:srobust});
            and
            (iii) removing disentanglement losses (NCNJ) increases the intrinsic score \textit{on the extrinsics} $\widehat{s}_\beta(z_E)$ (meaning more intrinsic information is erroneously in $z_E$), but allows for a modest improvement in extrinsic score $s_\theta(z_E)$, potentially due to the weaker constraints on the latent representation.
        }
    }
    \label{table:t2f-retftl}
\end{figure*}

\subsection{Latent Variational Autoencoder Results}
\label{sec:vaeresults}

We evaluate our VAE model on three main criteria:
	(1) representational fidelity,
	(2) generative modeling, and
	(3) intrinsic-extrinsic disentanglement.
Representational fidelity is captured simply as the reconstruction error, 
	measured via the Chamfer distance between input and output
	(see Fig.\ \ref{fig:vaerecon} for qualitative examples). %
To assess generative modeling capability, 
	we utilize the coverage and fidelity metrics \citep{achlioptas2018learning},
	which examine how well samples from our VAE represent a held-out test set.
In addition, utilizing our flow network, 
	we can measure the quality of spectrum generation using the standard log-likelihood.	
Finally, using the known ground truth intrinsics and extrinsics of our synthetic SMAL and SMPL data, 
	we can measure disentanglement quality via a pose-aware retrieval task.
We discuss our results and the details of these metrics in the following sections.
Figs.\ \ref{fig:t2f-genqual}, \ref{fig:t2f-recon}, and \ref{fig:t2f-disent}
(as well as Appendix Table \ref{tab:vaemass}) 
show our quantitative results on metrics for all of these criteria.

\newer{
We explore two variants of our model, using the STD and FTL AEs, as well as several ablations. 
Two ablations involve the AE:
    removing rotational supervision (the ``S'' vs.\ ``U'' models) 
    and using only \textit{one} model for both SMAL and SMPL (via the HA dataset),
    as opposed to having specialized models for each.
    Note that the latter scenario not only
    increases data complexity without altering model capacity,
    but it also removes some regularities that are present in the independent datasets due to their restricted categories. 
The remaining ablations affect only the VAE:
    using a PC-derived LBO (rather than the mesh-derived one we use by default), 
    altering our algorithm to not use the spectrum-derived latent intrinsics in training (GDVAE-FO), 
    and removing the additional disentanglement loss $\mathcal{L}_D$ (see \S\ref{sec:additionaldisentloss}).
}

\subsubsection{Generative Modeling}

\label{sec:res:genmodel}

We measure generative modeling quality using the metrics 
	introduced by \citet{achlioptas2018learning}.
Consider two sets of PC shapes: 
	$\varsigma_G$, a random set of generated samples, 
	and 
	$\varsigma_R$, a set of real PCs.
Note that generations are computed via 
	$P = D(D_x(z_E,z_I))R(D_q(z_R)) \in\varsigma_G$, 
	where $z_R,z_E,z_I\sim\mathcal{N}(0,I)$ 
	(see \S\ref{sec:ae:arch} and \ref{sec:hfvae}).
Briefly, we consider two measures:
	\textit{coverage}, 
		which checks how well $\varsigma_G$ covers the modes of $\varsigma_R$
		(a proxy for set diversity),
	and
	\textit{fidelity},
		which considers how faithful each element in $\varsigma_G$ is to
		its closest counterpart in $\varsigma_R$
		(a proxy for per-element realism).
Coverage is computed by matching each $Q\in\varsigma_G$ to its closest PC in $\varsigma_R$, and counting the percent of PCs chosen (matched) in $\varsigma_R$ (high coverage meaning most of the PCs in $\varsigma_R$ are represented in $\varsigma_G$).
Fidelity (also called minimum matching distance) is computed by matching 
	each $P\in\varsigma_R$ to its closest pair in $\varsigma_G$,
	taking the Chamfer distance between them,
	and averaging these distances over the dataset.
Fidelity is needed because coverage does not measure the quality of the matchings 
	(e.g., low quality PCs could be used to cover a given real PC).
Matching is always computed as the minimum Chamfer distance.
Similar to \citep{achlioptas2018learning},
	we generate a synthetic set five times larger than the held-out test set,
	and report the average of running the same evaluation twice.
\new{See Fig.\ \ref{fig:t2f-genqual} for a plot of generative metrics and Appendix Table \ref{tab:vaemass} for quantitative scores.}
For qualitative visualizations, random sample generations are shown in Fig.\ \ref{fig:vaegenerative}. 

Separately, our flow model $f_\lambda$ provides a generative model on LBOSs.
Using its bijectivity, we can directly compute the log-likelihood 
    (shown in Fig.\ \ref{fig:t2f-spectrallikelihood} and 
    Appendix Table \ref{tab:vaemass}).
This measures how well our spectral encoder maps real spectra 
	into the Gaussian latent space of the intrinsics.

Looking at Figs.\ \ref{fig:t2f-genqual} and \ref{fig:t2f-spectrallikelihood} (as well as Appendix Table \ref{tab:vaemass}),
    we can see that the GDVAE\PP\ and GDVAE-FO 
    score similarly for generative fidelity and coverage,
    and
    obtain mixed results on $\log P_\lambda(\lambda)$
    (the FO method performs better or similar with the FTL AE, but worse with the STD AE),
    but GDVAE-FO always has better reconstruction results.
In terms of AE type, results are mixed, though
    the FTL approach does tend to have slightly better coverage and worse fidelity.
We discuss results related to disentanglement quality in the next subsection.
    
\subsubsection{Shape-Pose Disentanglement}

\label{sec:ret}

To measure disentanglement quantitatively, we rely on a \textit{pose-aware} retrieval task 
	in which ground truth continuous values for intrinsics and extrinsics are known.

We start with a set of shapes (SMAL or SMPL) for which parameters 
for intrinsic shape  $\beta$ and extrinsic pose $\theta$ are known.
These shapes \emph{were not} used in training. 
Let point cloud $P_i$ have parameters $(\beta_i, \theta_i)$.
Using our model, we encode $P_i$ into a latent representation 
$\rho_i \in \{x_c(P_i), z(P_i), z_E(P_i), z_I(P_i)\} $.
We then measure distances between representations as 
	$d_\rho(P_i,P_j) = || \rho_i - \rho_j ||_2^2$,
	and rank the retrieved shapes based on $\rho_i$.
We measure the disentangled retrieval quality for a retrieved PC, $P_j$, using query $P_i$,
    by separately checking how well the intrinsic shape and non-rigid pose match.
	This is done by comparing the query ground truth parameters, 
	$(\beta_i, \theta_i)$, 
	to $(\beta_j, \theta_j)$, from the retrieved shape.

We compute the distance between these parameters, 
as the mean squared error between $\beta$ values and the average rotational distance $d_q$ across corresponding joint rotations, 
denoted  $E_\beta$ and $E_\theta$, respectively.
Note that we normalize $E_\beta$ and $E_\theta$ 
by the mean pairwise error across the dataset for each measure, 
so that it is relative to the expected error 
of a \textit{uniformly random} retrieval algorithm 
(1 corresponds to random retrieval, 
while 0 implies obtaining the same parameter set).
More specifically, we use the encoding(s) $\rho$ of a PC $P$, to retrieve the
three closest shapes (in terms of $d_\rho$), 
and
compute the errors $E_\beta$ and $E_\theta$ averaged over these three retrievals,
to obtain two errors per shape.
For a fixed encoding type $\rho \in \{x_c(P), z(P), z_E(P), z_I(P)\} $,
we get a final error by averaging over an entire held-out test set.
Hence, we obtain two scalars $E_\beta$ and $E_\theta$ for each choice of $\rho$.		

We then convert these errors into scores, 		
$s_\psi(\rho) = 1 - E_\psi(\rho)$,
where $\psi\in\{ \theta,\beta \}$.
We expect using $z_E$ for retrieval (i.e., as $\rho$) to result in a 
\textit{high} intrinsic error $E_\beta(z_E)$ (\textit{low} score $s_\beta(z_E)$), 
but a 
\textit{low} extrinsic error $E_\theta(z_E)$ (\textit{high} score $s_\theta(z_E)$).
Using $z_I$ should result in the converse:
a \textit{high} intrinsic score $s_\beta(z_I)$
and 
a \textit{low} extrinsic score $s_\theta(z_I)$.
We expect retrieval with $x_c$ or $z$ to obtain high scores for both parameters.

Lastly, we wish to have a final scalar score that expresses the quality of disentanglement obtained by the model.
Notice that $s_\beta$ and $s_\theta$ are normalized with respect to a random retriever, 
but are still not comparable (as the errors are originally different units and at different scales).
Hence we compute
$\widehat{s}_\psi(z_g) = {s_\psi(z_g)}/{s_\psi(x_c)}$,
with $z_g\in\{z,z_E,z_I\}$,
normalizing beta and theta retrievals to be in approximately the ``same'' units 
(both are errors relative to the AE).

With these normalizations, we make the following interpretations:
$\widehat{s}_\psi(z_g) = 0$ means that using $z_g$ to retrieve shapes is no better 
(with respect to $\psi$) than random retrieval.
In contrast, 
$\widehat{s}_\psi(z_g) = 1$ 
implies that $z_g$ performs just as well as using $x_c$; this comparison is relevant, 
because the AE limits the amount of information available to the VAE.
Higher scores 
(e.g., $\widehat{s}_\psi(z_g) = 2$)
imply that $z_g$ performs $\widehat{s}_\psi \times$ \textit{better} than $x_c$ 
(specifically for retrieving pose alone, 
    when $\psi = \theta$, or 
    intrinsic shape alone, 
    when $\psi = \beta$).

Our normalized retrieval scores $\widehat{s}_\psi(z_g)$
are then used to compute a 
final disentanglement scalar
	\begin{equation}
		\mathcal{S} = 
		  \widehat{s}_\beta(z_I) 
		+ \widehat{s}_\theta(z_E) 
		- \widehat{s}_\beta(z_E) 
		- \widehat{s}_\theta(z_I).
		\label{eq:disentscore}
	\end{equation}
Higher $\mathcal{S}$ 
    requires accurate extrinsics-based retrieval 
    in terms of pose (high $\widehat{s}_\theta(z_E)$), 
    but poor retrieval 
    (when using $z_E$) 
    with respect to intrinsics $\beta$ 
    (low $\widehat{s}_\beta(z_E)$); 
at the same time, 
        it requires the opposite performance
        for the latent intrinsics $z_I$. 
Note that random retrieval performance results in all terms being zero 
(hence $\mathcal{S}=0$); 
however, one \textit{also} obtains 
$ \mathcal{S} = 0 $ if performance 
for each term is the same as the AE 
(since all four terms would be one).
In other words, 
good performance retrieving intrinsics (extrinsics) with $z_I$ ($z_E$) will be cancelled out 
by good performance retrieving extrinsics (intrinsics) with $z_I$ ($z_E$).
This shows that a high $ \mathcal{S} $ requires disentanglement between $z_E$ and $z_I$.

Disentanglement scores are shown in Fig.\ \ref{fig:t2f-disent}
    (as well as Appendix Table \ref{tab:vaemass}).
Note that retrieval scores $\mathcal{S}$ are 1.08 and 1.04, for SMAL and SMPL respectively, 
	in the original GDVAE work. %
As such, the GDVAE\PP\  model obtains significantly superior disentanglement scores 
	across both datasets (including from the HA model) -- around double the score of the original model.

From Fig.\ \ref{table:t2f-retftl}
(as well as Appendix Table \ref{tab:ret} and Fig.\ \ref{table:t2f-retstd}),
we also observe the superiority of 
    $\widetilde{z}_I = f_\lambda(\lambda)$ 
    over $z_I$ 
	in retrieving intrinsics, 
	suggesting one should use the spectrum directly 
	when it is available for such a task,
	though the raw spectrum $\lambda$ cannot be used for other tasks 
	(e.g., smooth interpolation, generation, or same-pose-different-shape retrieval).

Qualitatively, we can assess disentanglement by looking at interpolations within the factorized latent space (shown in Fig.\ \ref{fig:vaeinterp}).
The interpolation plots also show examples of \textit{pose transfers} 
	(upper-right and lower left corners per inset).
For SMAL and SMPL, one can see that the network correctly disentangles articulated pose and shape. 
For MNIST, where an obvious notion of articulation is not present, 
moving in $z_I$
tends to change digit thickness or allow large-scale shape alterations,
while changing $z_E$ 
approximately leaves geodesic distance distributions unchanged
(though it can change major factors, like topology).
	
We can also consider 
	the retrievals qualitatively 
	based on the disentangled latent vectors.
	Fig.\ \ref{fig:vaeretrieval}
	shows what shapes the networks think are most similar to each query,
	in terms of intrinsics versus extrinsics.
We observe that $z_E$ is able to retrieve very similar articulations across many animals and/or human body types, while $z_I$ correctly retrieves similar shapes without regard for non-rigid pose.
For MNIST, retrieval with 
$z_E$ tends to mostly return the same digit with differing thicknesses, 
while retrieval with $z_I$ also largely results in the same digit, but under isometric (non-geodesic-altering) deformations. 
There are some exceptions to these, 
such as the nines retrieved by the $z_I$ from the six 
(as the spectrum is unaffected by rotation)
or the fives there (potentially due to the closeness of the end of the last stroke in the five to the upper portion of the digit, as well as its thickness, leading to greater intrinsic similarity).
The ones retrieved for the eight by $z_E$ are less obvious to interpret; they may be due to the low dimensionality of $z_E$ or the similarity of ones to thin eights. 

\new{
We conclude by noting that the GDVAE\PP\ (S or U) generally has the best disentanglement scores (see Fig.\ \ref{fig:t2f-disent}),  
while NCNJ has the second-best, but suffers from worse generative quality (Figs.\ \ref{fig:t2f-genqual} and \ref{fig:t2f-spectrallikelihood}).
In comparison, the HA and PCLBO models are generally slightly worse across all metrics (generation, reconstruction, and disentanglement).
The FO scenario has by far the worst disentanglement score among all models, underscoring the importance of our altered training regime.
While there is some noise (e.g., higher reconstruction error for SMAL-FTL-U in Fig.\ \ref{fig:t2f-recon} or superior generative quality for HA on SMAL-FTL in Fig.~\ref{fig:t2f-genqual}), these trends broadly hold across datasets and AE types (STD and FTL), suggesting our new approach is generally better.
}

\subsubsection{Spectral Robustness}
\label{sec:srobust}
\newer{
Although we use PCs as our shape representation for these experiments, 
	our spectra are computed on the mesh forms of the shapes, via the cotan weight formulation \citep{meyer2003discrete}. 
This provides a useful measure of performance for our model (effectively bounding the performance we can expect with lower quality LBOs), as well as allowing comparison to the original GDVAE model, upon which we are trying to improve. 
Further, we expect methods for LBOS extraction from PCs to improve over time (e.g., via advances in machine learning \citep{marin2021spectral} and geometry processing \citep{sharp2021geometry}), making the use of higher quality operators more feasible. 
}

However, for completeness, 
	we also investigated the effect of computing the spectra directly 
	on our subsampled point clouds.
This mesh-to-point-cloud conversion process introduces several additional sources of noise:
	for instance, parts far in geodesic distance may be close in Euclidean space
	(altering the LBO), 
	and the subsampling of the surface %
	(our PCs being smaller than the number of vertices in SMPL and SMAL)
	also introduces noise. %
Hence, we expect results to be degraded, compared to the prior section.
For computing the point cloud LBO (PCLBO), 
	we use the robust ``tufted'' Laplacian operator \citep{sharp2020laplacian}.

The scalar disentanglement results are shown in Fig.\ \ref{fig:t2f-disent} and Appendix Table \ref{tab:vaemass}.
While the scores do decrease overall, 
	they are still superior to the scores from the original GDVAE 
	(which used mesh-derived LBOSs to obtain 1.08 and 1.04, for SMAL and SMPL respectively) 
	and the GDVAE-FO models.
From Fig.\ \ref{table:t2f-retftl}
(as well as Appendix Table \ref{tab:ret} and Fig.\ \ref{table:t2f-retstd}),
	we see that 
	two major terms are negatively affected in the PCLBO case,
	likely due to noise in the estimated LBOSs:
(1) the
	ability of $z_I$ to capture intrinsics degrades,
	indicated by the decline in 
	$\widehat{s}_\beta(z_I)$,
	scores;
and
(2) intrinsic information is not removed as effectively from $z_E$,
	indicated by high values of
	$\widehat{s}_\beta(z_E)$ 
	(especially for SMPL).

\subsubsection{Ablations}
\label{sec:ablations}

Lastly, we consider the effect of ablating two aspects of the model: 
the additional disentanglement loss 
$\mathcal{L}_D$ and the shape-from-spectrum reconstruction used in the GDVAE\PP\  training.

First, we investigate the utility of the additional disentanglement penalties.
By removing these losses, we have no covariance and no Jacobian terms; 
	we denote this scenario NCNJ.
For SMAL, the disentanglement scores seem unaffected by this ablation; 
	however, it seems to have introduced a trade-off between reconstruction and generative modelling errors,
	with $d_C$ improving (see Fig.\ \ref{fig:t2f-recon}), but coverage and $\log P_\lambda(\lambda)$ degrading (see Figs.\ \ref{fig:t2f-genqual} and \ref{fig:t2f-spectrallikelihood}).
For SMPL, 
	NCNJ results in degradations in the disentanglement and generative coverage scores (see Figs.\ \ref{fig:t2f-genqual} and \ref{fig:t2f-disent}).
Note that since the VAE prior is Gaussian, it presupposes latent independence \citep{higgins2017beta}; 
hence, disentanglement is likely to affect the prior fitting (and hence generative quality and $\log P_\lambda(\lambda)$ as well).	
	
Second, we look at the effectiveness of the ``flow-only'' training approach,
	where we do not perform latent shape-from-spectrum during training to perform reconstruction,
	and instead only use the direct encoding of the AE output.
	We find that this incurs the most significant degradation in terms of disentanglement score across both datasets (see Fig.\ \ref{fig:t2f-disent}),
	showing the importance of using the uncontaminated spectrum for training,
		rather than relying on the $\mathcal{L}_F$ to force $z_I$ to carry only intrinsic information.
One may notice that, even though GDVAE-FO is similar to the GDVAE model \citep{aumentado2019geometric}\footnote{Except for the flow network and altered AEs.}, 
it has a much lower disentanglement score. 
This can be partly explained by the increase in dimensionality of the latent intrinsics, 
as the newer model has a 4-5 times larger $\text{dim}(z_I)$ than the original GDVAE, making disentanglement more difficult. 

\subsubsection{Results Summary}

\newer{
The GDVAE\PP\  shows substantial improvements over the original GDVAE model in terms of disentanglement. 
Using the PCLBO or the combined model (HA dataset) ablations decrease performance, but still maintain this advantage.
This improvement also holds regardless of AE type or whether rotational supervision is ablated, 
    showcasing the robustness of our model to AE settings. 
Much of this gain stems from our shape-from-spectrum training regime: when ablated (the GDVAE-FO model), disentanglement capabilities are crippled.
}

\subsection{Mesh Experiments}

The previous results demonstrated the improvements of our approach over the prior GDVAE model. To illustrate applicability to a different 3D shape modality, as well as facilitate comparison to other works, we also tested our method on mesh data. 

\label{sec:meshexps}

\subsubsection{Human Bodies (AMASS)}

\label{sec:meshexps:amass}

\newcommand{\sww}{0.11}

\begin{figure*}
    \centering
    \renewcommand{\tabcolsep}{0mm}
    \renewcommand{\arraystretch}{0.11}
    \begin{tabular}[b]{cccc}
        \includegraphics[scale=\sww]{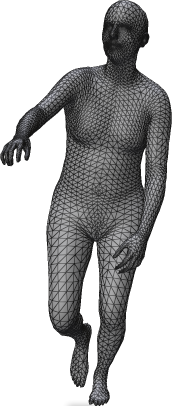} & 
        \includegraphics[scale=\sww]{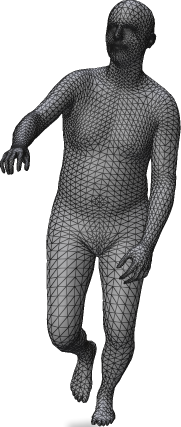} & 
        \includegraphics[scale=\sww]{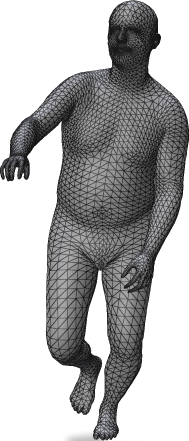} & 
        \includegraphics[scale=\sww]{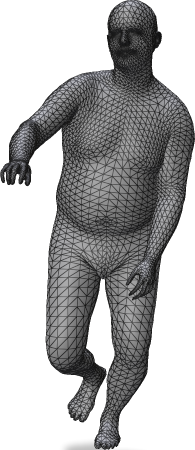} \\
        \includegraphics[scale=\sww]{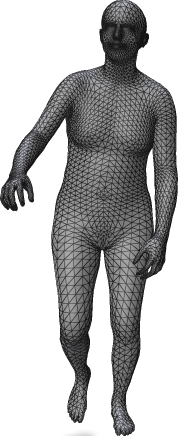} & 
        \includegraphics[scale=\sww]{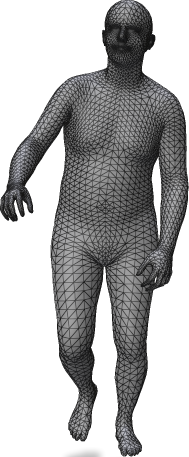} & 
        \includegraphics[scale=\sww]{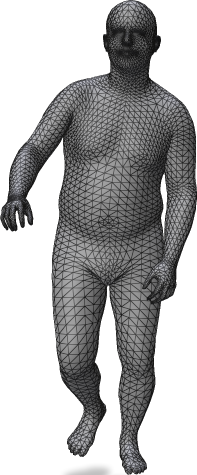} & 
        \includegraphics[scale=\sww]{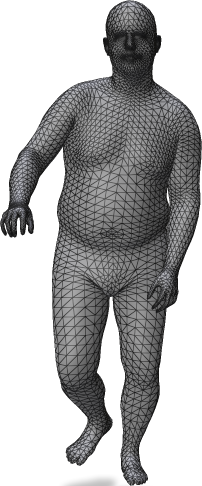} \\
        \includegraphics[scale=\sww]{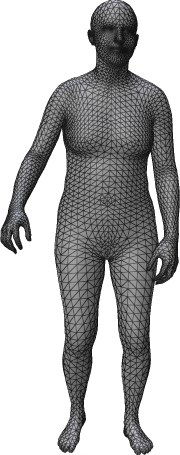} & 
        \includegraphics[scale=\sww]{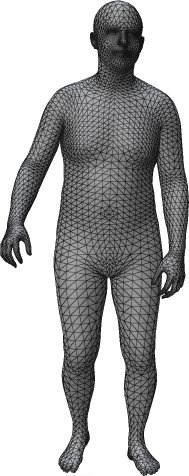} & 
        \includegraphics[scale=\sww]{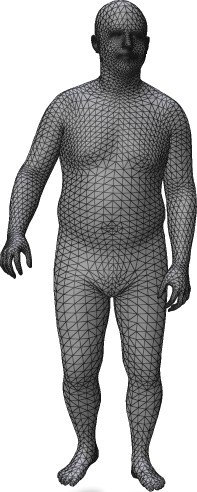} & 
        \includegraphics[scale=\sww]{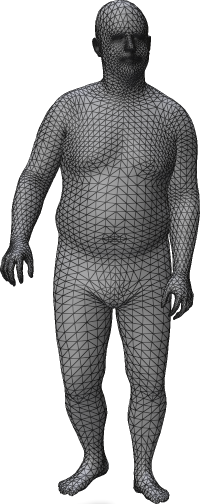} \\
        \includegraphics[scale=\sww]{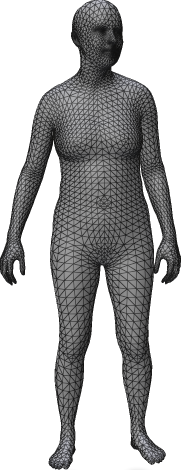} & 
        \includegraphics[scale=\sww]{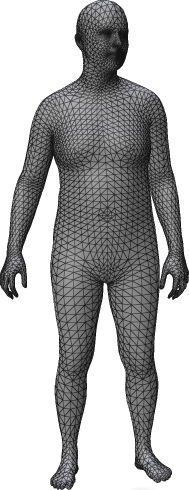} & 
        \includegraphics[scale=\sww]{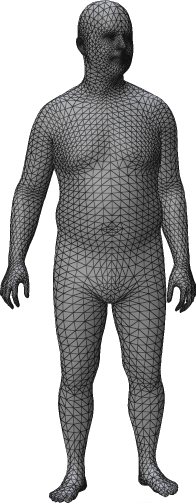} & 
        \includegraphics[scale=\sww]{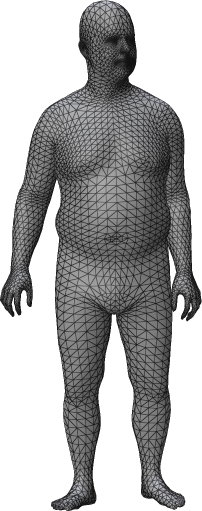} \\
    \end{tabular}\hfill
    \renewcommand{\sww}{0.105}
    \begin{tabular}[b]{cccc}
        \includegraphics[scale=\sww]{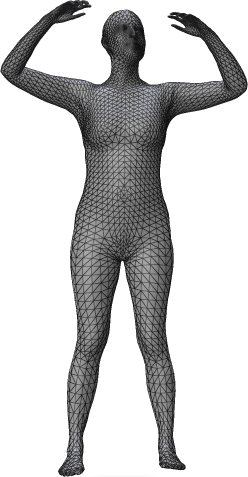} & 
        \includegraphics[scale=\sww]{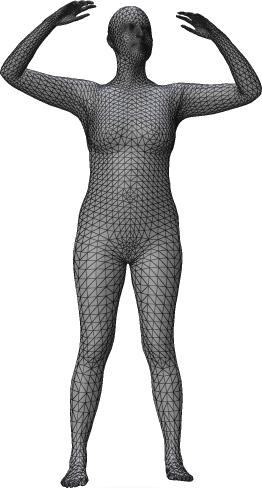} & 
        \includegraphics[scale=\sww]{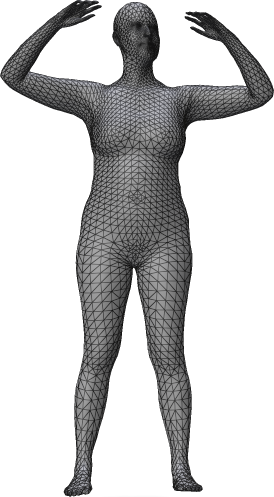} & 
        \includegraphics[scale=\sww]{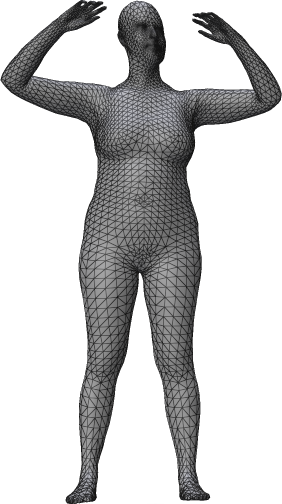} \\
        \includegraphics[scale=\sww]{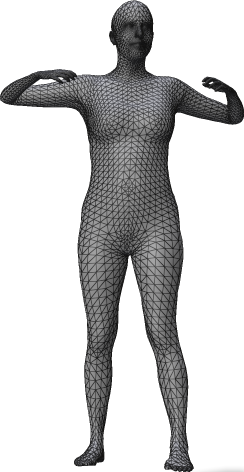} & 
        \includegraphics[scale=\sww]{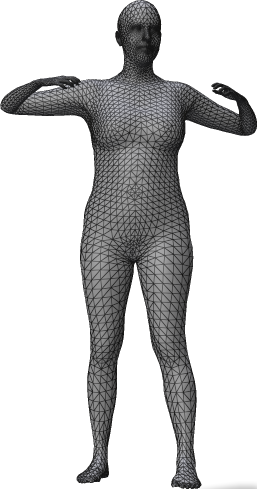} & 
        \includegraphics[scale=\sww]{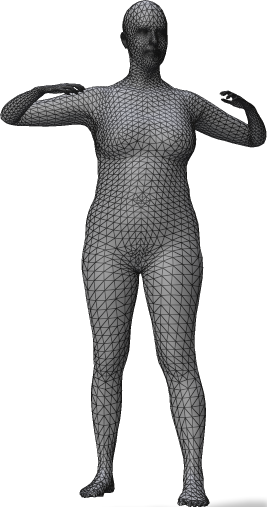} & 
        \includegraphics[scale=\sww]{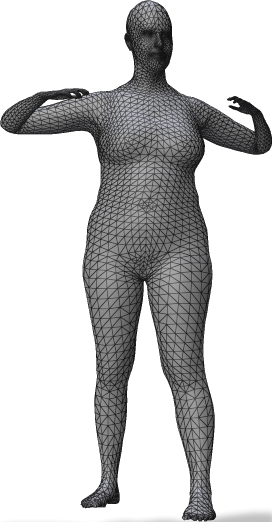} \\
        \includegraphics[scale=\sww]{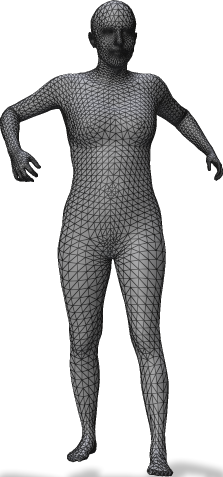} & 
        \includegraphics[scale=\sww]{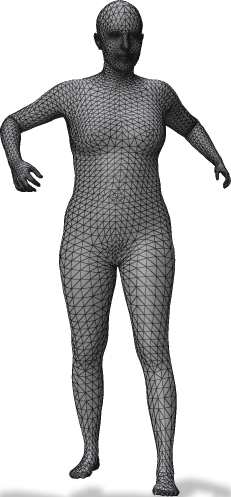} & 
        \includegraphics[scale=\sww]{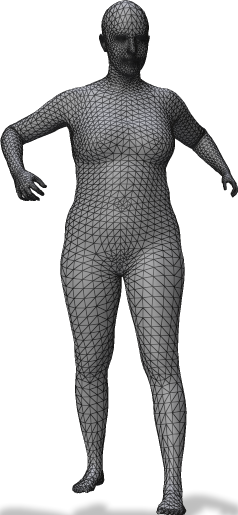} & 
        \includegraphics[scale=\sww]{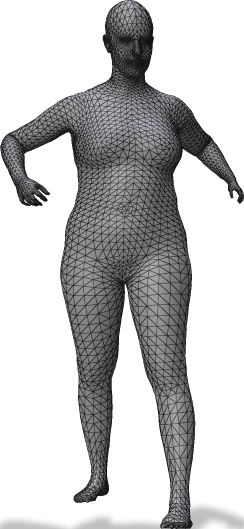} \\
        \includegraphics[scale=\sww]{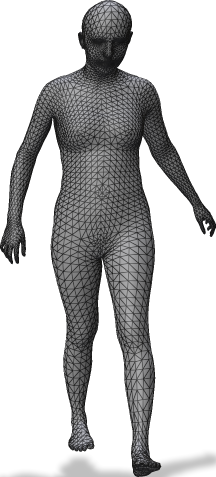} & 
        \includegraphics[scale=\sww]{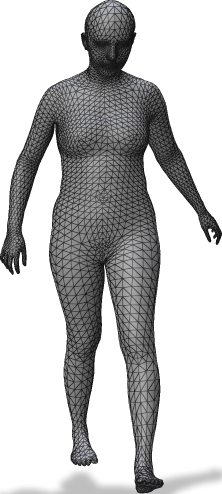} & 
        \includegraphics[scale=\sww]{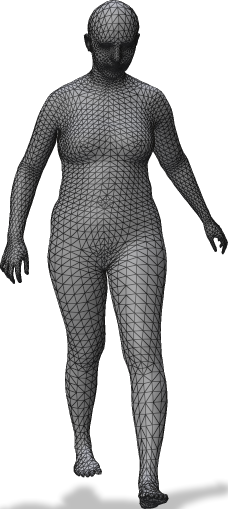} & 
        \includegraphics[scale=\sww]{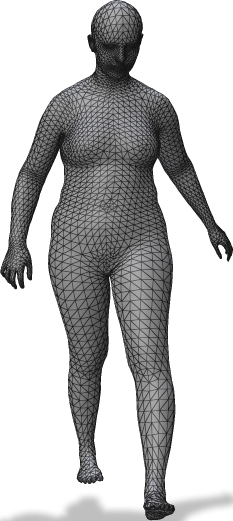} \\
    \end{tabular}\hfill
    \renewcommand{\sww}{0.105}
    \begin{tabular}[b]{cccc}
        \includegraphics[scale=\sww]{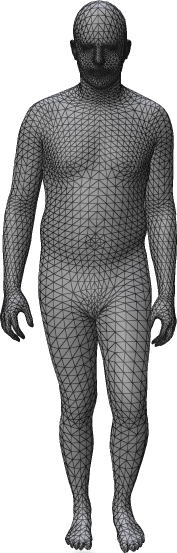} & 
        \includegraphics[scale=\sww]{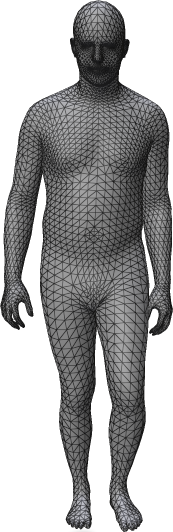} & 
        \includegraphics[scale=\sww]{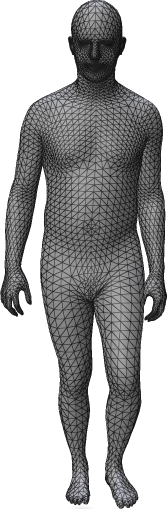} & 
        \includegraphics[scale=\sww]{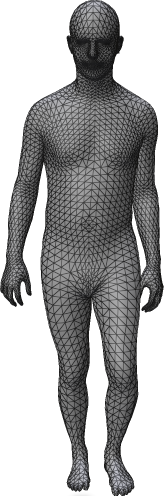} \\
        \includegraphics[scale=\sww]{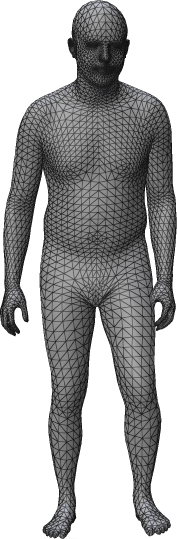} & 
        \includegraphics[scale=\sww]{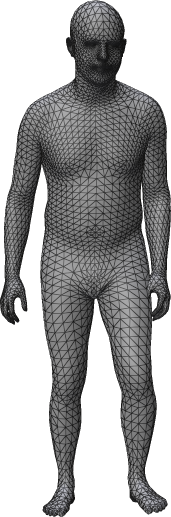} & 
        \includegraphics[scale=\sww]{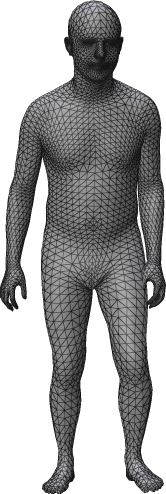} & 
        \includegraphics[scale=\sww]{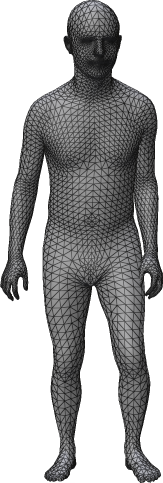} \\
        \includegraphics[scale=\sww]{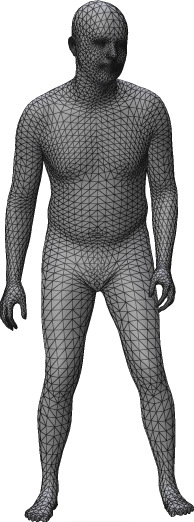} & 
        \includegraphics[scale=\sww]{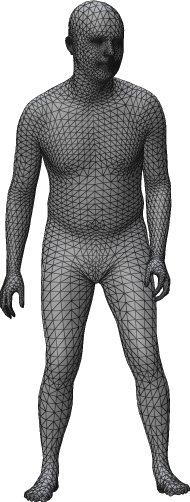} & 
        \includegraphics[scale=\sww]{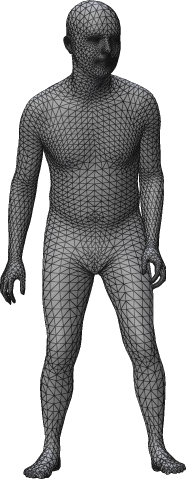} & 
        \includegraphics[scale=\sww]{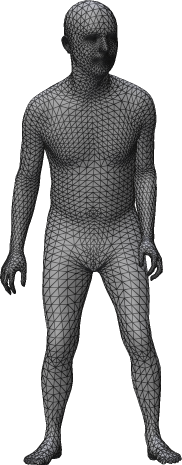} \\
        \includegraphics[scale=\sww]{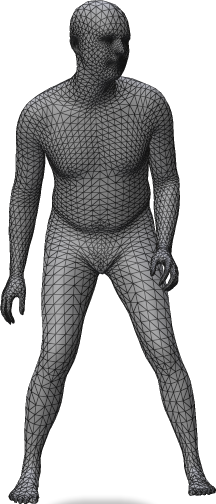} & 
        \includegraphics[scale=\sww]{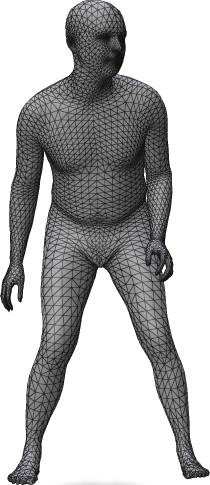} & 
        \includegraphics[scale=\sww]{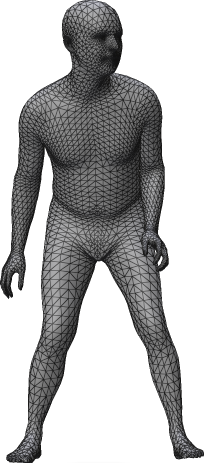} & 
        \includegraphics[scale=\sww]{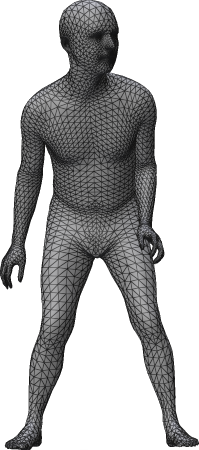} \\
    \end{tabular}\hfill
    \renewcommand{\sww}{0.11}
    \begin{tabular}[b]{cccc}
        \includegraphics[scale=\sww]{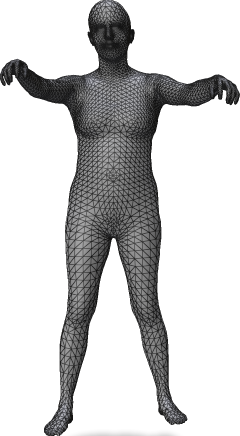} & 
        \includegraphics[scale=\sww]{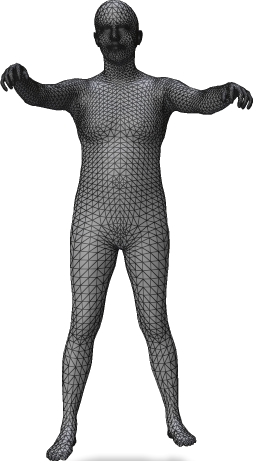} & 
        \includegraphics[scale=\sww]{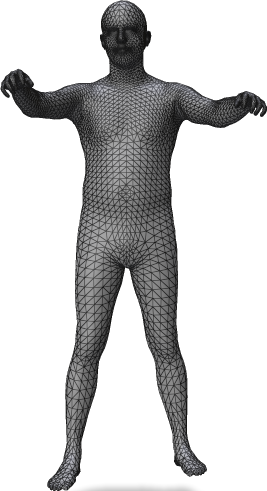} & 
        \includegraphics[scale=\sww]{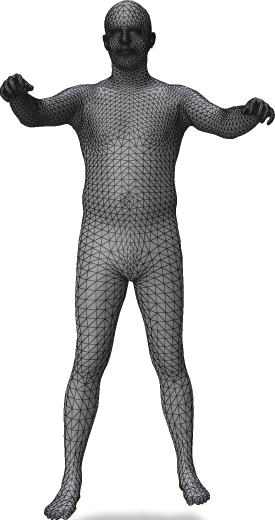} \\
        \includegraphics[scale=\sww]{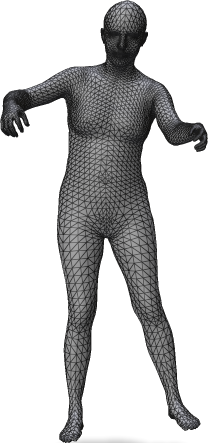} & 
        \includegraphics[scale=\sww]{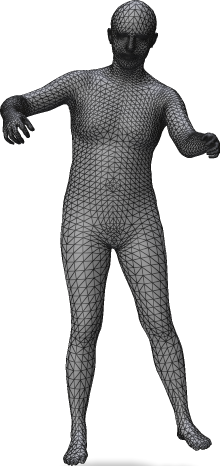} & 
        \includegraphics[scale=\sww]{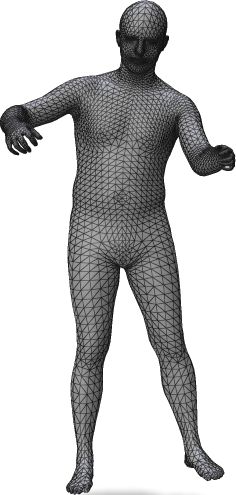} & 
        \includegraphics[scale=\sww]{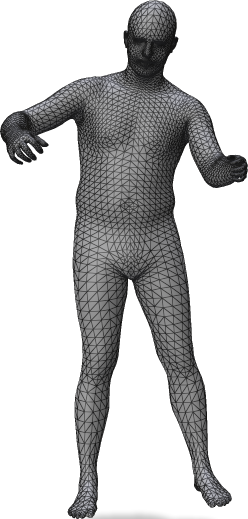} \\
        \includegraphics[scale=\sww]{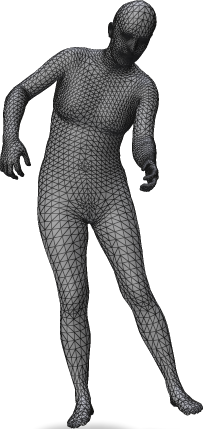} & 
        \includegraphics[scale=\sww]{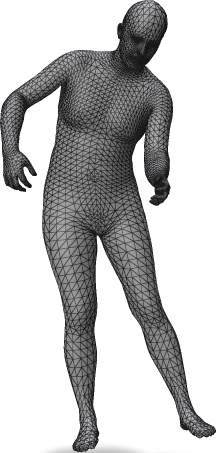} & 
        \includegraphics[scale=\sww]{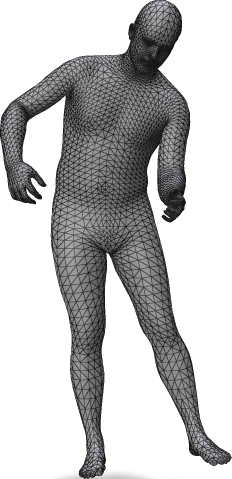} & 
        \includegraphics[scale=\sww]{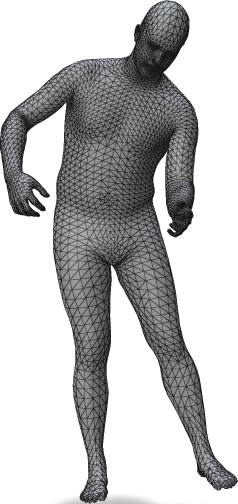} \\
        \includegraphics[scale=\sww]{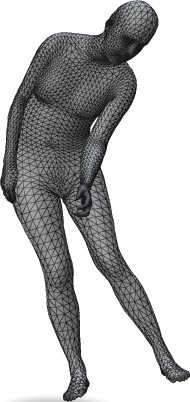} & 
        \includegraphics[scale=\sww]{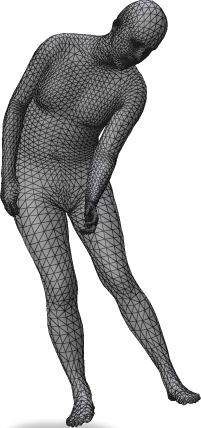} & 
        \includegraphics[scale=\sww]{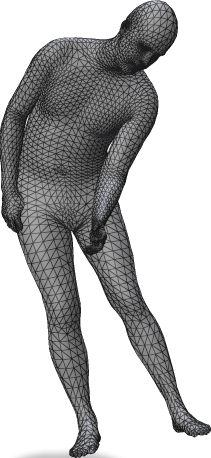} & 
        \includegraphics[scale=\sww]{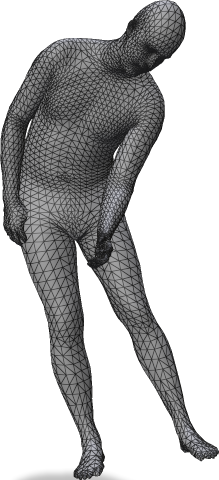} \\
    \end{tabular}\hfill
    \renewcommand{\sww}{0.11}
    \begin{tabular}[b]{cccc}
        \includegraphics[scale=\sww]{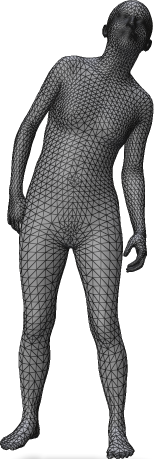} & 
        \includegraphics[scale=\sww]{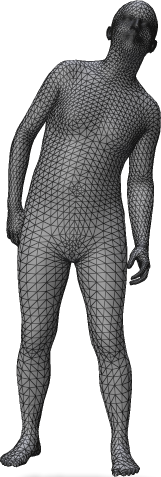} & 
        \includegraphics[scale=\sww]{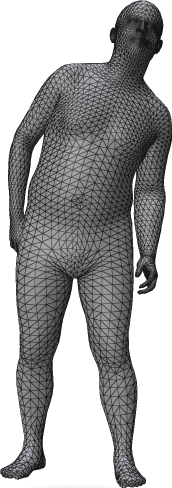} & 
        \includegraphics[scale=\sww]{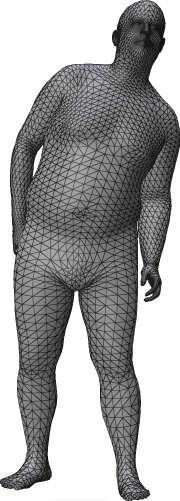} \\
        \includegraphics[scale=\sww]{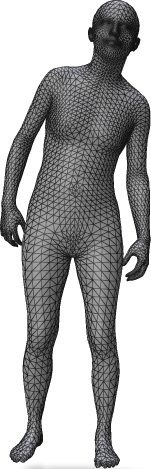} & 
        \includegraphics[scale=\sww]{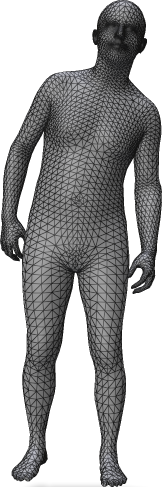} & 
        \includegraphics[scale=\sww]{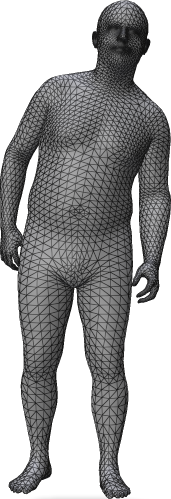} & 
        \includegraphics[scale=\sww]{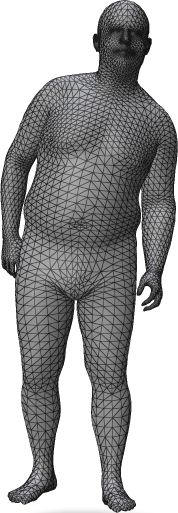} \\
        \includegraphics[scale=\sww]{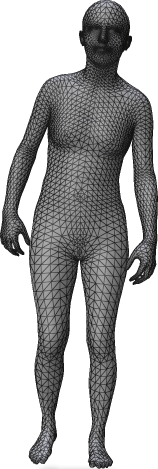} & 
        \includegraphics[scale=\sww]{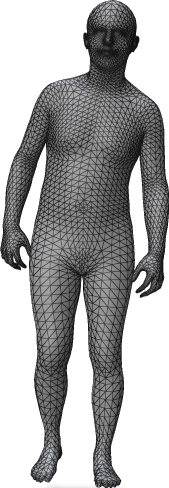} & 
        \includegraphics[scale=\sww]{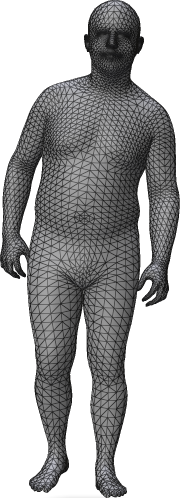} & 
        \includegraphics[scale=\sww]{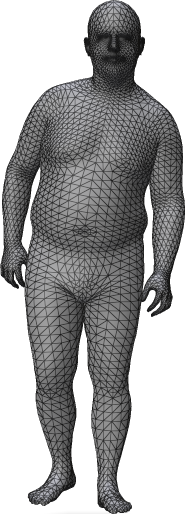} \\
        \includegraphics[scale=\sww]{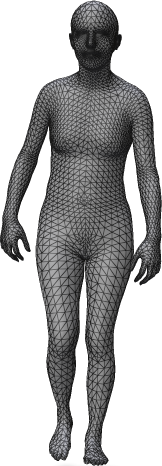} & 
        \includegraphics[scale=\sww]{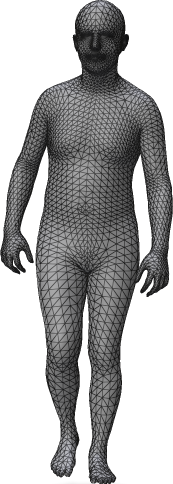} & 
        \includegraphics[scale=\sww]{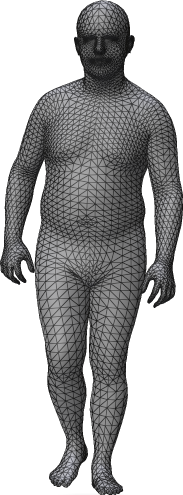} & 
        \includegraphics[scale=\sww]{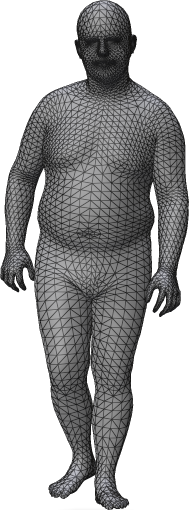} \\
    \end{tabular}
    \caption{
        Latent interpolations of AMASS validation mesh shapes.
        As in Fig.\ \ref{fig:vaeinterp}, per inset, horizontal or vertical movement traverses latent intrinsics $z_I$ or extrinsics $z_E$, respectively, via linear interpolation. See \S\ref{sec:meshexps:amass} for model details.
        Upper-left and lower-right shapes correspond to real input meshes; upper-right and lower-left shapes therefore correspond to \textit{pose transfers}.
    }
    \label{fig:amassinterps}
\end{figure*}

\newer{
First, we utilize the AMASS dataset \citep{AMASS:ICCV:2019}, which combines a number of human motion datasets and provides parametric fitting via SMPL, in order to compare with ``Unsupervised Shape and Pose Disentanglement for 3D Meshes'' (USPD) \citep{zhou2020unsupervised} on disentangled retrieval and pose transfer tasks.
}

\newer{ %
We alter the AE to (1) process a mesh input, instead of a PC, and (2) output ordered vertex coordinates instead of arbitrary PC sample points.
Following other work \citep{marin2021spectral,tan2018variational}, we use a fully connected encoder.
Each output position of the decoder is now semantically associated to a fixed vertex. 
We alter the loss function to use vertex-to-vertex mean squared error for reconstruction, rather than Eq.\ \ref{eq:aereconloss} 
(with other terms remaining the same).
Notice that we use the vertex correspondence to compute reconstruction loss during the AE training, but this information is \textit{not} utilized for disentanglement by the VAE, which only has access to latent encodings $x$ in our two-stage training regime.
See Appendix \S\ref{app:meshes:amass} for details, including hyper-parameter settings.
}

\begin{table}[t] %
	\setlength{\tabcolsep}{5.97pt}
	\centering
	\caption{
	    \newer{
	    Pose transfer scores (lower is better) on mesh data from AMASS, 
		measured in vertex-to-vertex distance in millimeters
		(with comparative numbers from \citet{zhou2020unsupervised}).
		Note that the GDVAE model (which outputs a PC) measures error via the more forgiving one-sided Chamfer distance instead (as in \citet{zhou2020unsupervised}). 
		See \S\ref{sec:meshexps:amass} for additional details.
		We find that our GDVAE\PP\  outperforms the GDVAE, 
		    but underperforms the more strongly supervised USPD,
		    which uses subject labels and mesh correspondence for disentanglement.
		We mark the GDVAE score with a *, as it is computed with a different (more lenient) metric.
		}
	}
	\label{tab:amassposetransfer}       %
	\begin{tabular}{cccc}
		\hline\noalign{\smallskip}
		& GDVAE & GDVAE\PP\  & USPD \\
		\noalign{\smallskip}\hline\noalign{\smallskip}
		Error $\downarrow$ & 54.44* & 31.54 & 19.43 \\
		\noalign{\smallskip}\hline
	\end{tabular}
\end{table}

\begin{table}[t] %
	\setlength{\tabcolsep}{5.1pt}
	\centering
	\caption{
	\newer{
	    Pose-aware disentangled retrieval scores on mesh data from AMASS. 
        Note that our latent intrinsics and extrinsics nomenclature refers to the latent ``shape'' and ``pose'' (or articulation) vectors in other works. 
        Comparative numbers from \citet{zhou2020unsupervised}.
		See \S\ref{sec:meshexps:amass} for additional details
		and
		Appendix \S\ref{app:meshvaria} for empirical standard deviations.
		We show the difference $\Delta$ between retrieval scores as well
		    ($\widetilde{E}_\beta(z_E) - \widetilde{E}_\beta(z_I)$
		     and
 		     $\widetilde{E}_\theta(z_I) - \widetilde{E}_\theta(z_E)$), 
		    such that higher is better. 
		The GDVAE appears to perform well on $\Delta(\widetilde{E}_\beta)$; however, this is due to the high overall error magnitude in intrinsics retrieval, $\widetilde{E}_\beta$.
        The authors of USPD previously observed a reduction in entanglement when using PCA (measured by $\Delta$); 
            we therefore compare against this dimensionally reduced version as well (using $\dim(z_I)=5$ and $\dim(z_E) = 15$).
        We show the best score across categories between the PCA and non-PCA models in \textbf{bold}.
		In particular, notice that our method underperforms the non-PCA USPD in terms of $\widetilde{E}_\beta(z_I)$ and $\widetilde{E}_\theta(z_E)$, but outperforms it in terms of the $\Delta$ differences; in other words, while USPD retrieves shapes with close intrinsics/extrinsics (when querying with latent intrinsics/extrinsics), those shapes also have similar extrinsics/intrinsics, suggesting a level of shape-pose entanglement remains.
		In contrast, the PCA-reduced version of USPD has better $\Delta$ values;
		    however, in this case, our PCA-based method has better $\widetilde{E}_\theta(z_E)$, as well as better $\Delta(\widetilde{E}_\theta)$ overall.
		}
	}
	\label{tab:amassretrievals}       %
	\begin{tabular}{cc|cc|c}
		\hline\noalign{\smallskip}
        \multicolumn{2}{c}{Retrieval with latent:} & Intrinsics & Extrinsics & $\Delta$ \\
		\noalign{\smallskip}\hline\noalign{\smallskip}
		\multirow{2}{*}{GDVAE} 
		    & $\widetilde{E}_\beta$  & 2.80 $\downarrow$ & 4.71  $\uparrow$ & 1.91 $\uparrow$ \\
		    & $\widetilde{E}_\theta$ & 1.47 $\uparrow$  & 1.44 $\downarrow$ & 0.03 $\uparrow$ \\\noalign{\smallskip}\hline\hline\noalign{\smallskip}
        \multirow{2}{*}{GDVAE\PP\  } %
            & $\widetilde{E}_\beta$  & {0.41}  $\downarrow$ & \textbf{1.36} $\uparrow$ & \textbf{0.94} $\uparrow$ \\
            & $\widetilde{E}_\theta$ & \textbf{1.15} $\uparrow$  & {0.80} $\downarrow$ & \textbf{0.35} $\uparrow$ \\\noalign{\smallskip}\hline\noalign{\smallskip}
        \multirow{2}{*}{USPD} 
	        & $\widetilde{E}_\beta$  & \textbf{0.14} $\downarrow$ & 0.92 $\uparrow$ & 0.78 $\uparrow$  \\
		    & $\widetilde{E}_\theta$ & 0.94 $\uparrow$  & \textbf{0.76} $\downarrow$ & 0.18 $\uparrow$ 
		    \\\noalign{\smallskip}\hline\hline\noalign{\smallskip}
        \multirow{2}{*}{GDVAE\PP\ (PCA) } %
            & $\widetilde{E}_\beta$  & {0.50}  $\downarrow$ & {1.49} $\uparrow$ & 0.98 $\uparrow$ \\
            & $\widetilde{E}_\theta$ & {1.21} $\uparrow$  & \textbf{0.82} $\downarrow$ & \textbf{0.40} $\uparrow$ \\\noalign{\smallskip}\hline\noalign{\smallskip}
		\multirow{2}{*}{USPD (PCA)} 
		    & $\widetilde{E}_\beta$  & \textbf{0.34} $\downarrow$ & \textbf{2.14} $\uparrow$ & \textbf{1.80} $\uparrow$ \\
		    & $\widetilde{E}_\theta$ & \textbf{1.23} $\uparrow$  & 0.87 $\downarrow$ & {0.36} $\uparrow$ \\
		\noalign{\smallskip}\hline
	\end{tabular}
\end{table}

\newer{ %
We test on two tasks, pose transfer and pose-aware retrieval, on held-out subsets of AMASS.
We use the same evaluation methodology and splits as USPD for consistency, which induces minor differences with the evaluations on PCs from previous sections.
We first measure pose transfer quality: given two meshes, we can obtain a ground truth transfer by exchanging the SMPL parameters for articulation $\theta$, while fixing those for body shape $\beta$, and obtain our prediction by doing so for $z_I$ and $z_E$. After decoding, we can measure the average \textit{vertex-to-vertex} Euclidean distance between the predicted and true transfers.
These values are shown in Table \ref{tab:amassposetransfer}.
While we greatly outperform the original GDVAE, we still underperform USPD for this task.
Nevertheless, beyond the additional requirements of USPD (subject labels and vertex correspondence), we note that our VAE is trained to reconstruct AE latent vectors (i.e., it is not trained end-to-end to reduce real-space vertex-to-vertex error), which also potentially contributes to worse performance on this task.
In Fig.\ \ref{fig:amassinterps}, 
    we show example latent interpolations in the disentangled space, including pose transfers.
}

\newer{
We then examine pose-aware retrieval quality.
For ease of comparison, we use the error measures on SMPL parameters from USPD:
$ \widetilde{E}_\beta(z_\psi)  = 
    \mathbb{E}_{M_Q} || \beta(M_Q) - \beta(M(z_\psi(M_Q))) ||_2 $
and
$ \widetilde{E}_\theta(z_\psi) = 
    \mathbb{E}_{M_Q} || q(\theta(M_Q)) - q(\theta(M(z_\psi(M_Q)))) ||_2 $,
where 
$ \psi\in \{E,I\} $,
$\beta, \theta$ refer to shape and pose SMPL parameters,
$M_Q$ is a query mesh from a held-out test set,
$M(z_\psi(M_Q))$ is the nearest neighbour mesh to $M_Q$ as measured by MSE in $z_\psi$ space,
and
$q$ converts pose angles to unit quaternions. 
We also examine the differences
$ \Delta(\widetilde{E}_\beta)  = \widetilde{E}_\beta(z_E) - \widetilde{E}_\beta(z_I) $ and
$ \Delta(\widetilde{E}_\theta) = \widetilde{E}_\theta(z_I) - \widetilde{E}_\theta(z_E) $, which should ideally be high. 
Quantitative results are compiled in Table~\ref{tab:amassretrievals}.
Compared to USPD, our method has higher $\widetilde{E}_\beta(z_I)$ and $\widetilde{E}_\theta(z_E)$, 
but outperforms in terms of both differences $\Delta(\widetilde{E}_\beta)$ and $\Delta(\widetilde{E}_\theta)$. 
Intuitively, 
    when querying with latent intrinsics/extrinsics, USPD obtains shapes with very close intrinsics/extrinsics, but those shapes also have similar extrinsics/intrinsics;
    in other words, some shape-pose entanglement remains.
By comparison, 
    the GDVAE\PP\  has less entanglement (higher error when retrieving intrinsics/extrinsics with latent extrinsics/intrinsics), 
    but also higher error in terms of retrieving intrinsics/extrinsics via latent intrinsics/extrinsics.
}

\newer{
The authors of USPD also considered a version of their model with reduced dimensionality via PCA, which controlled for the difference in dimensionality between USPD and the GDVAE.
They found it had better disentanglement properties, as evidenced by the higher differences $\Delta$, but worse $\widetilde{E}_\theta(z_E)$ and $\widetilde{E}_\beta(z_I)$ values.
We observe a similar effect occurs with our model when using PCA to transform $z_E$ and $z_I$ to that same dimensionality as well (from 9 to 5 for $\dim(z_I)$ and 18 to 15 for $\dim(z_E)$).
Comparing the PCA-reduced case, USPD has superior retrieval results in terms of intrinsics, but ours has better values in terms of extrinsics ($\widetilde{E}_\theta(z_E)$ and $\Delta(\widetilde{E}_\theta)$).  
}

\newer{
We note that these $\Delta$ measures effectively weight the two terms equally, which may not be ideal. 
However, we find that a uniformly random retrieval algorithm incurs average errors of 6.5 for $\widetilde{E}_\beta$ and 1.76 for $\widetilde{E}_\theta$ (as well as $\Delta$ values close to zero), suggesting none of these models are actually selecting random intrinsics/extrinsics for given query extrinsics/intrinsics, as one would expect from perfectly disentangled retrieval.
}

\newer{
Overall, our model underperforms USPD on pose transfer, but is more competitive on retrieval.
However, we remark that USPD relies on known subject identities to obtain sets of people with identical intrinsics, but different extrinsic pose, providing the network with explicit information about the articulated pose space for a given shape.
It also utilizes vertex correspondence, which our method does not use for disentanglement.
Together, these provide powerful learning signals to the network. %
This is different than our use of the LBOS, which is specific to a geometric entity, extractable from raw geometry, and not based on semantic knowledge about identity.
In other words, USPD performs better for these tasks, but is more specialized, whereas our approach defines a generic structural prior on the deformation space of objects, which happens to disentangle articulation and intrinsic shape as a natural geometric consequence.
Other factors, such as our need for low latent dimensionality and inability to do end-to-end training (necessitated by our information-theoretic disentanglement) also contribute to reduced performance.
}

\subsubsection{Human Faces (CoMA)}

\label{sec:meshexps:coma}

\newer{
We also investigated our approach on human face meshes, derived from the CoMA dataset \citep{COMA:ECCV18}. In particular, we consider the utility of our approach on a  shape-from-spectrum task, under identical experimental conditions to recent work by \citet{marin2021spectral}. %
Given an LBOS $\lambda$, our goal is to reconstruct the original shape $S$. Due to our use of a flow network, we can easily encode $\lambda$, to obtain the latent intrinsics $\widetilde{z}_I(\lambda)$. However, we also require latent extrinsics, which we must obtain without access to $S$. Fortunately, our VAE-based formulation permits a straightforward, principled solution: simply use the mode of the  Gaussian prior over the latent extrinsics, meaning we set $z_E \equiv \Vec{0}$. We can then decode $z = (\widetilde{z}_I(S), \Vec{0})$ to obtain the reconstructed shape $\widehat{S}$ with ``mean'' extrinsic pose, according to the prior. In practice, if we use more eigenvalues, more of the shape will be represented in $\widetilde{z}_I$; for fair comparison, we use the same number as \citet{marin2021spectral} (i.e., $\dim(\lambda) = 30$).
Error is simply the vertex-to-vertex Euclidean distance between the meshes $S$ and $\widehat{S}$.
Appendix \S\ref{app:meshes:coma} contains additional details.
}

\newer{
Our results are displayed in Table \ref{tab:coma.shapefromspectrum}. 
We consider two nearest neighbour baselines ($\lambda$-NN-$L_2$ and $\lambda$-NN-$d_\lambda$), which simply retrieve the closest shape in the training set to the given spectrum, using the Euclidean distance or our weighted $d_\lambda$ (Eq.\ \ref{eq:specdist}), respectively.
We remark that using $d_\lambda$ provides superior retrievals than the $L_2$ metric, as it corrects for the growth of the monotonic LBOS, which overweights high frequency geometric details.
The method by \citet{marin2021spectral} outperforms these baselines, but our method (using the mode of the VAE prior for $z_E$) performs the best overall.
We observe that there is still a performance gap compared to using $z_E(x)$ (bottom row of the table); however, this is to be expected, since using the truncated spectrum alone will lose some information. %
}

\newer{
We also provide example latent interpolations on the CoMA dataset in Fig.\ \ref{fig:coma.interps}.
Notice that our latent intrinsics capture overall head shape, while the latent extrinsics contain deformations of the mouth and other facial expressions, despite only using raw meshes as input to the algorithm.
Compared to \citet{marin2021spectral}, which must perform a regularized optimization to obtain such disentanglement, our method simply linearly interpolates $z_I$ and $z_E$.
}

\begin{table}[t]
    \centering
    \begin{tabular}{ccc}
    \hline\noalign{\smallskip}
        Method & Error $\downarrow$ & Spectrum Only \\
        \noalign{\smallskip}\hline\noalign{\smallskip}
        $\lambda$-NN-$L_2$ 
            & 4.47 & Yes \\ 
        $\lambda$-NN-$d_\lambda$ 
            & 2.63 & Yes \\ 
        \citet{marin2021spectral}
            & 1.61 & Yes \\
        $\widetilde{z}_I(\lambda)$ \& $ z_E \equiv 0 $ (Ours) 
            & 1.52 & Yes \\  \noalign{\smallskip}\hline\noalign{\smallskip} %
        Full $z$ (Ours) 
            & 1.24 & No \\
    \noalign{\smallskip}\hline
    \end{tabular}
    \caption{
    \newer{
        Empirical shape-from-spectrum results on CoMA, following the experimental settings of \citet{marin2021spectral}.
        Columns: reconstruction methods, test set error (in terms of vertex-to-vertex $L_2$ distance), and whether or not some form of information about the shape extrinsics is used.
        Rows refer to different approaches:
        $\lambda$-NN-$L_2$ simply retrieves the closest shape in the training set, based on the $L_2$ distance between LBOSs; 
        $\lambda$-NN-$d_\lambda$ is the same nearest neighbour approach, but using our $d_\lambda$ metric (Eq.\ \ref{eq:specdist}) instead (which avoids over-emphasizing high frequency geometric details);
        ``$\widetilde{z}_I(\lambda)$ \& $ z_E \equiv 0 $'' denotes simply setting $z_E$ to be zero;
        and
        ``Full $z$'' means $z=(z_I(x), z_E(x))$ is used, 
        which forms a lower bound on the error we can expect, as it uses both intrinsic and extrinsic information from the full shape $S$.
        Our approach with $z_E \equiv 0$ uses only the spectrum $\lambda$ of the shape (and no other information from $S$ or $x$); it is equivalent to simply choosing $z_E$ as the mode of the VAE prior, over the space of latent extrinsics.
        Overall, our method, which separates latent intrinsics and extrinsics, as well as guarantees invertibility, performs best.
        For our two VAE-based approaches, we observe a standard error of the mean of 0.02 (using only $\lambda$) and 0.009 (using full $z$). 
        All error values are $\times 10^{-5}$.
        }
    }
    \label{tab:coma.shapefromspectrum}
\end{table}

\newcommand{\sw}{0.085}
\begin{figure*}[t]
    \centering
    {
    \renewcommand{\tabcolsep}{0mm}
    \renewcommand{\arraystretch}{0.1}
    \begin{tabular}[b]{cccc}
        \includegraphics[scale=\sw]{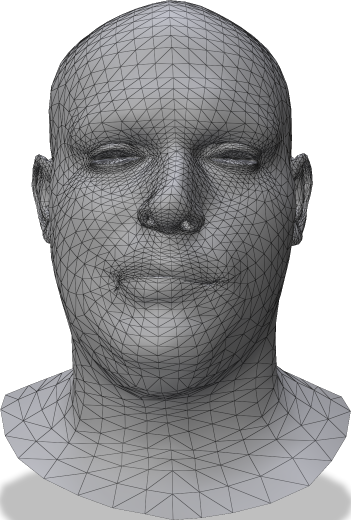} & 
        \includegraphics[scale=\sw]{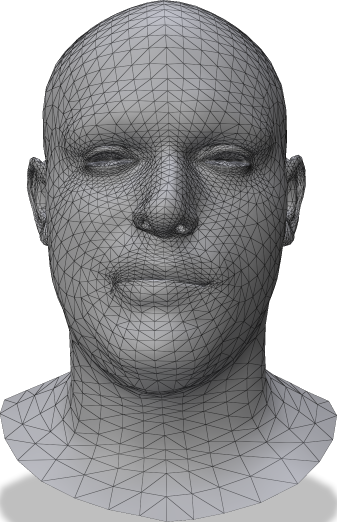} & 
        \includegraphics[scale=\sw]{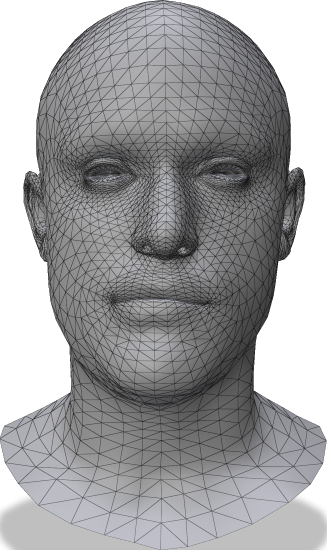} & 
        \includegraphics[scale=\sw]{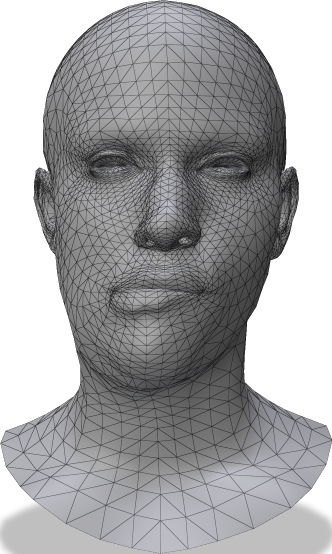} \\
        \includegraphics[scale=\sw]{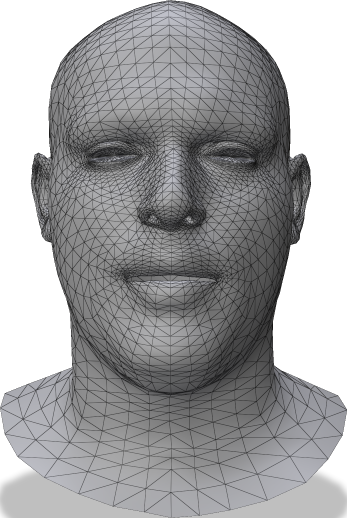} & 
        \includegraphics[scale=\sw]{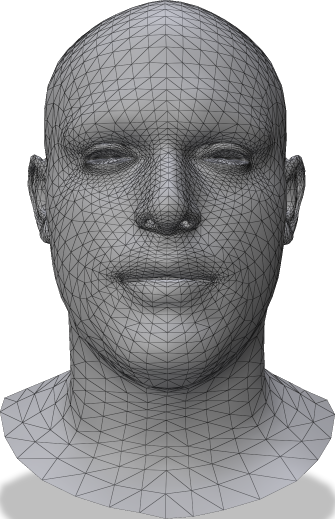} & 
        \includegraphics[scale=\sw]{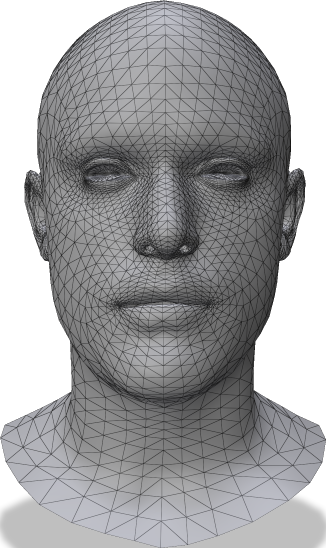} & 
        \includegraphics[scale=\sw]{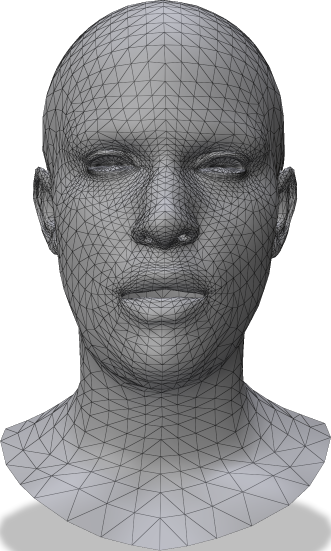} \\
        \includegraphics[scale=\sw]{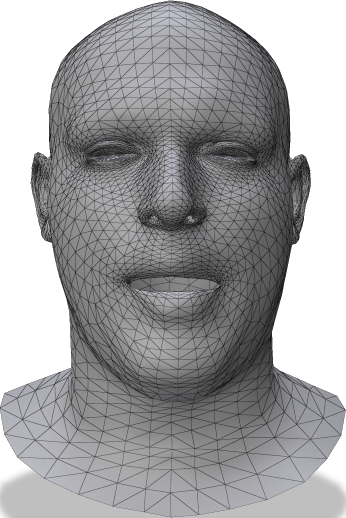} & 
        \includegraphics[scale=\sw]{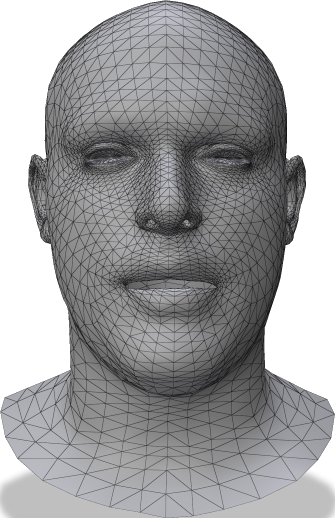} & 
        \includegraphics[scale=\sw]{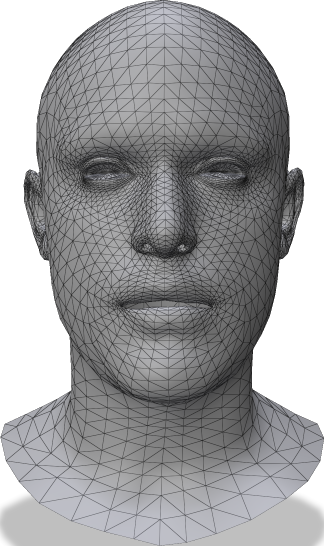} & 
        \includegraphics[scale=\sw]{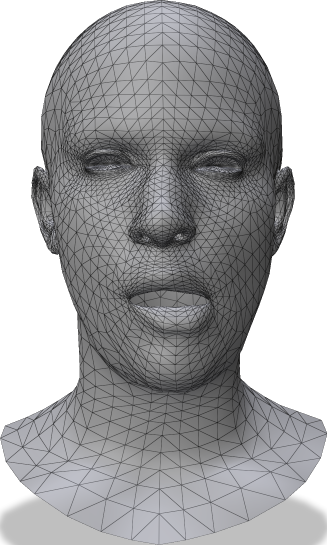} \\
        \includegraphics[scale=\sw]{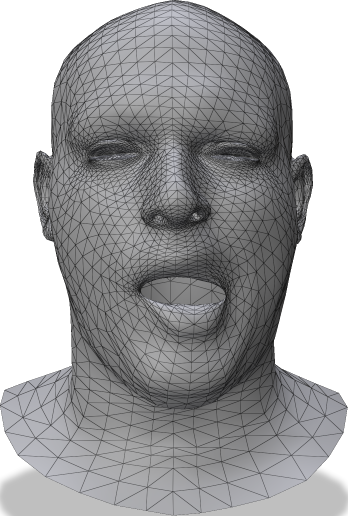} & 
        \includegraphics[scale=\sw]{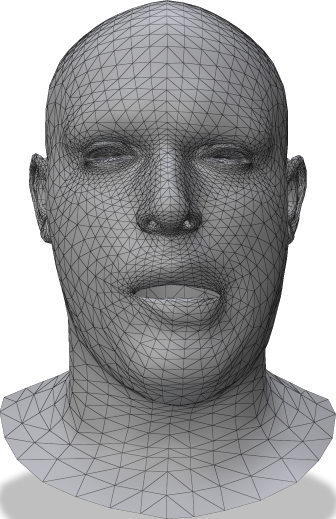} & 
        \includegraphics[scale=\sw]{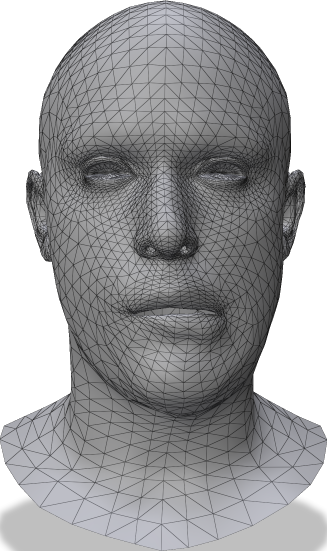} & 
        \includegraphics[scale=\sw]{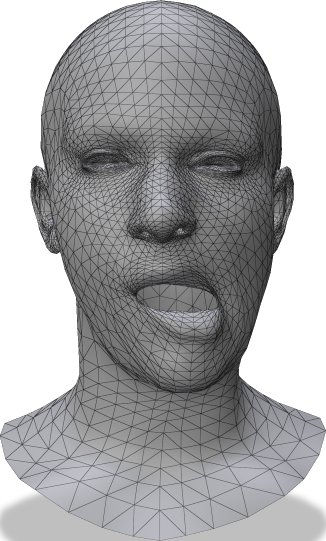} \\
    \end{tabular}\hfill
    \begin{tabular}[b]{cccc}
        \includegraphics[scale=\sw]{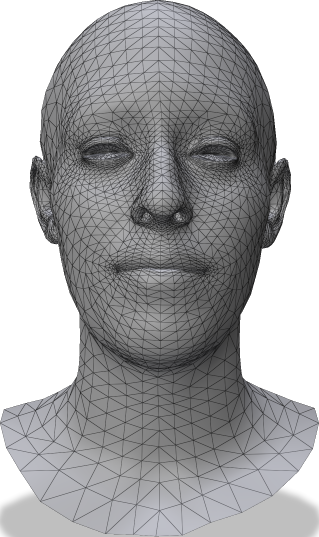} & 
        \includegraphics[scale=\sw]{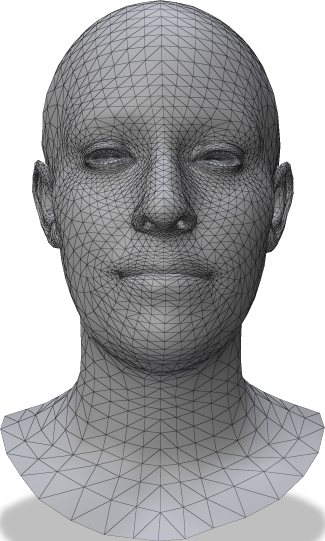} & 
        \includegraphics[scale=\sw]{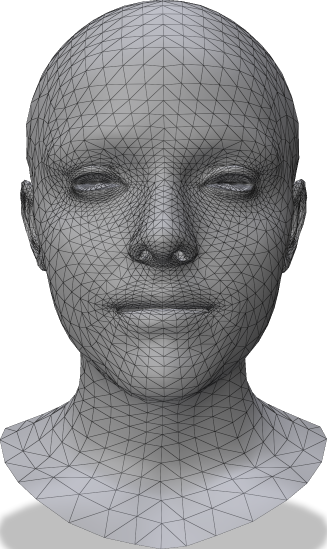} & 
        \includegraphics[scale=\sw]{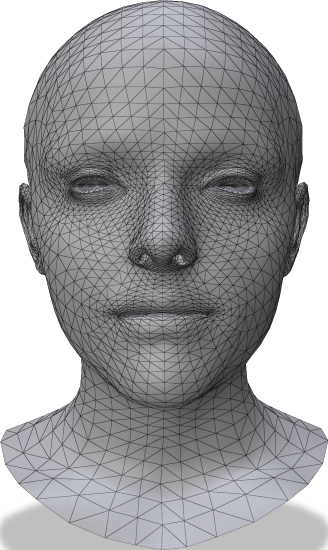} \\
        \includegraphics[scale=\sw]{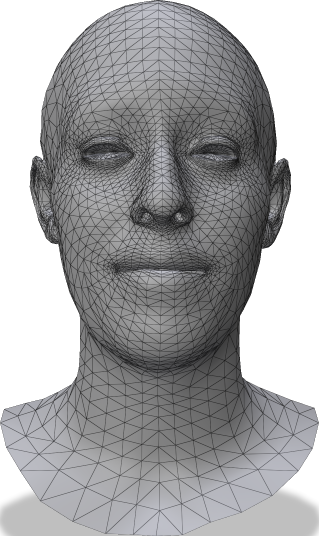} & 
        \includegraphics[scale=\sw]{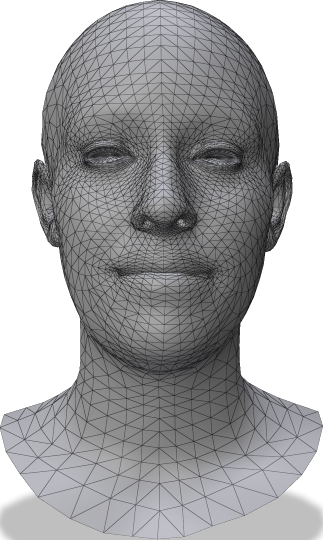} & 
        \includegraphics[scale=\sw]{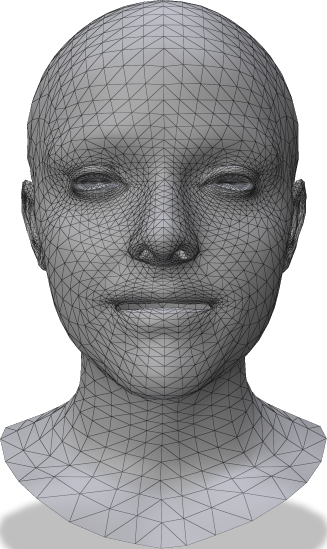} & 
        \includegraphics[scale=\sw]{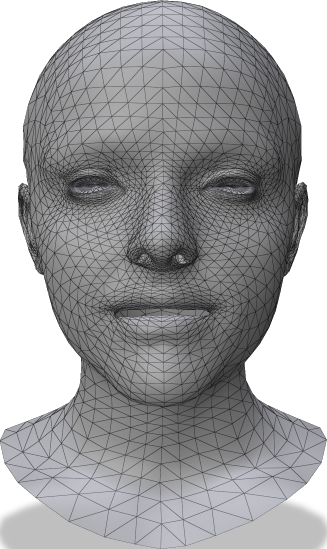} \\
        \includegraphics[scale=\sw]{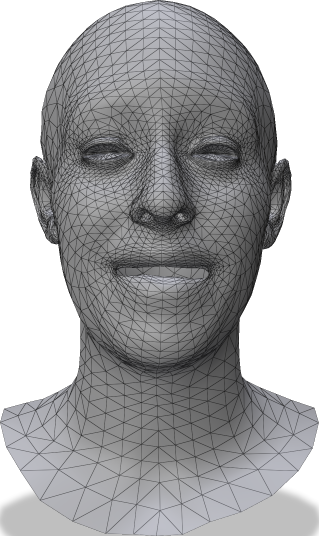} & 
        \includegraphics[scale=\sw]{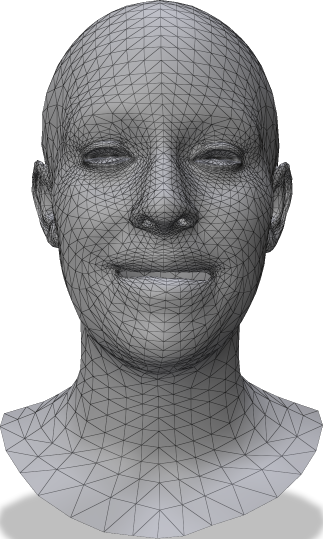} & 
        \includegraphics[scale=\sw]{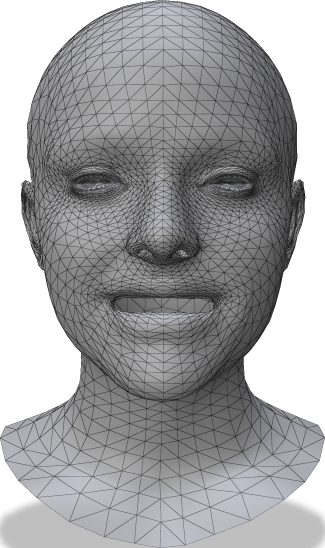} & 
        \includegraphics[scale=\sw]{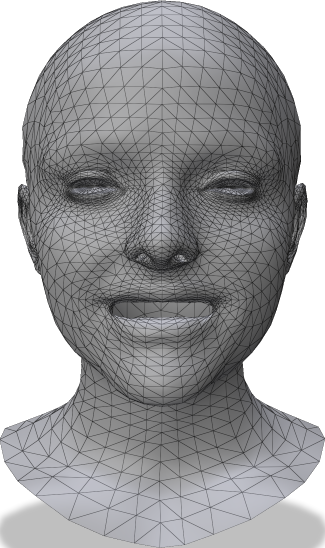} \\
        \includegraphics[scale=\sw]{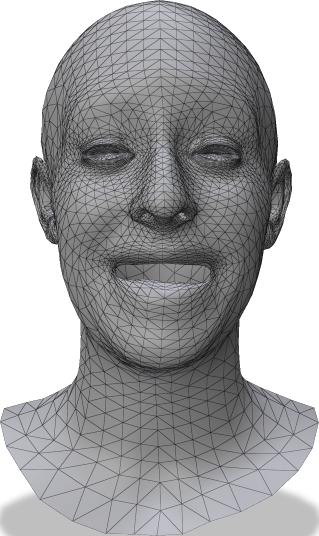} & 
        \includegraphics[scale=\sw]{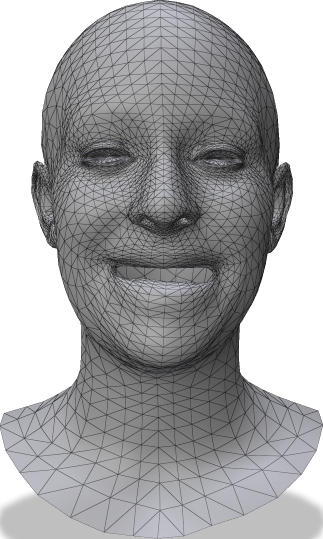} & 
        \includegraphics[scale=\sw]{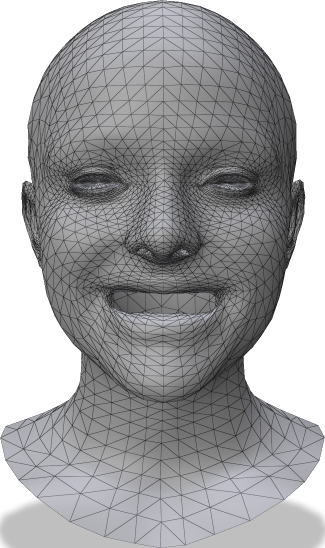} & 
        \includegraphics[scale=\sw]{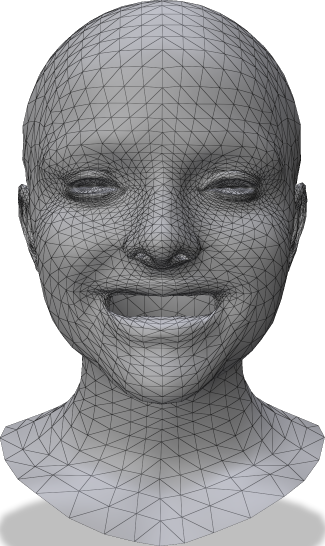} \\
    \end{tabular}\hfill
    \begin{tabular}[b]{cccc}
        \includegraphics[scale=\sw]{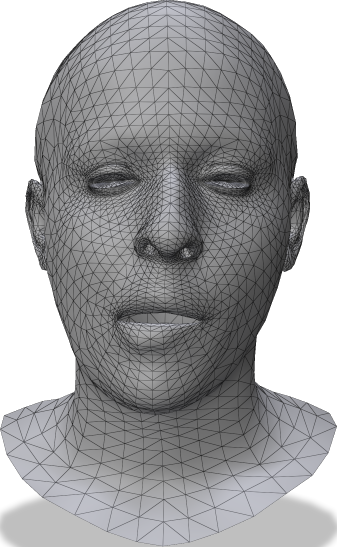} & 
        \includegraphics[scale=\sw]{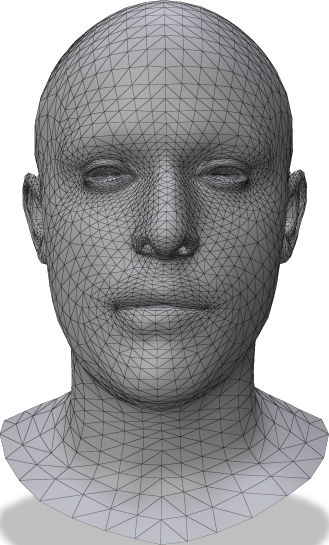} & 
        \includegraphics[scale=\sw]{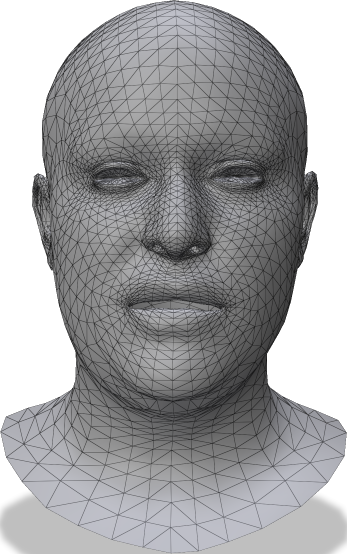} & 
        \includegraphics[scale=\sw]{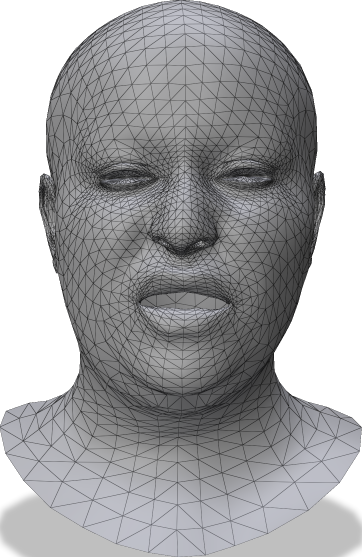} \\
        \includegraphics[scale=\sw]{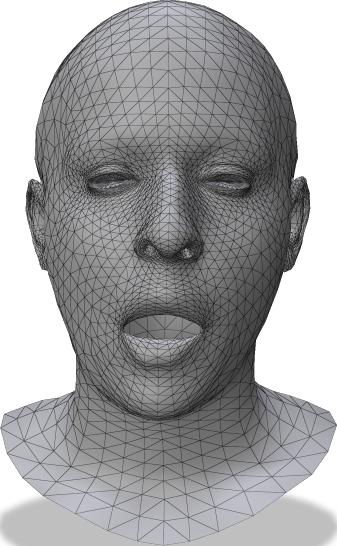} & 
        \includegraphics[scale=\sw]{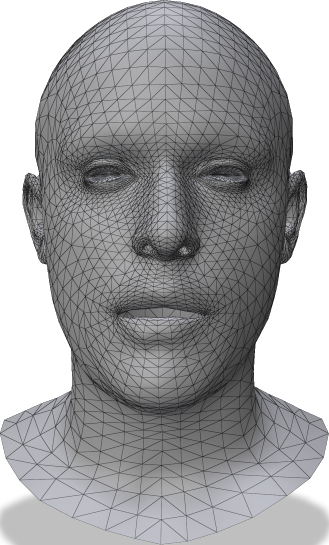} & 
        \includegraphics[scale=\sw]{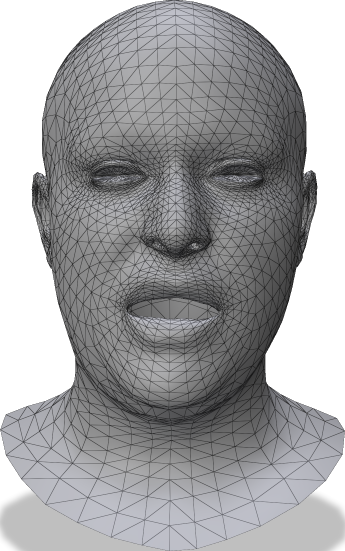} & 
        \includegraphics[scale=\sw]{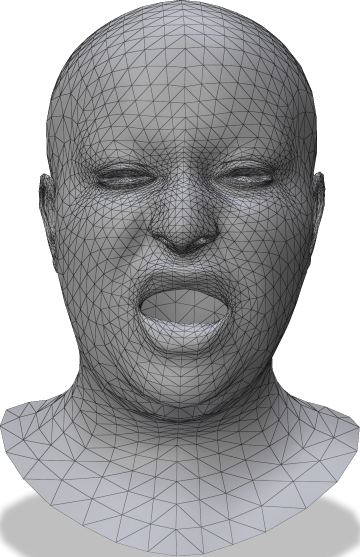} \\
        \includegraphics[scale=\sw]{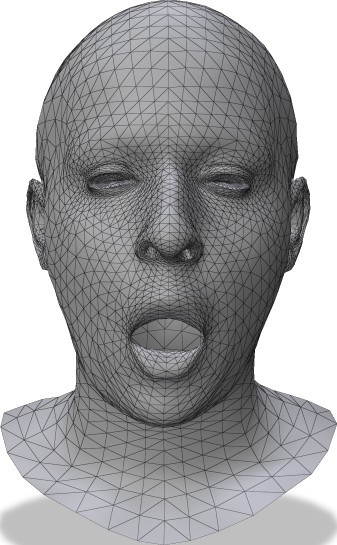} & 
        \includegraphics[scale=\sw]{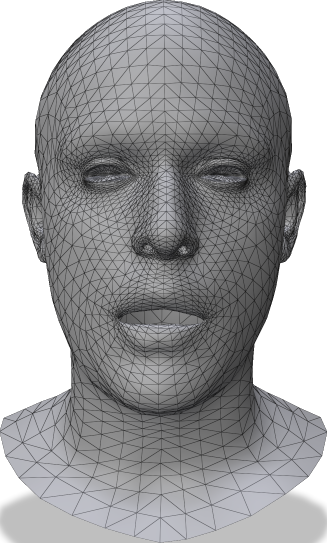} & 
        \includegraphics[scale=\sw]{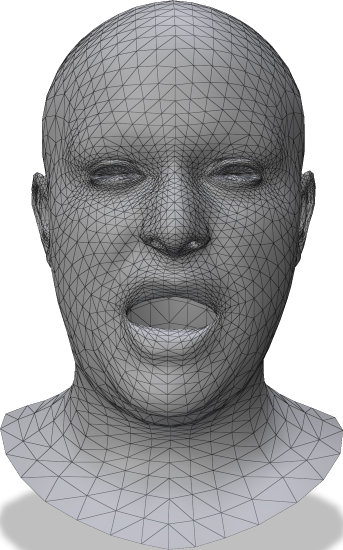} & 
        \includegraphics[scale=\sw]{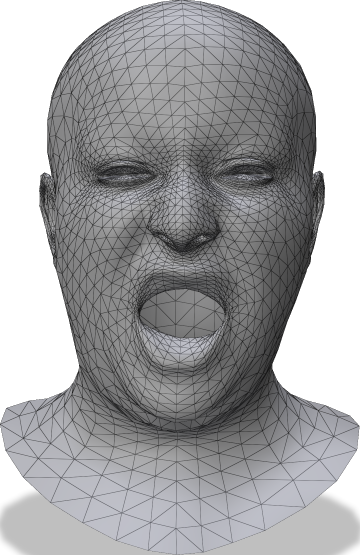} \\
        \includegraphics[scale=\sw]{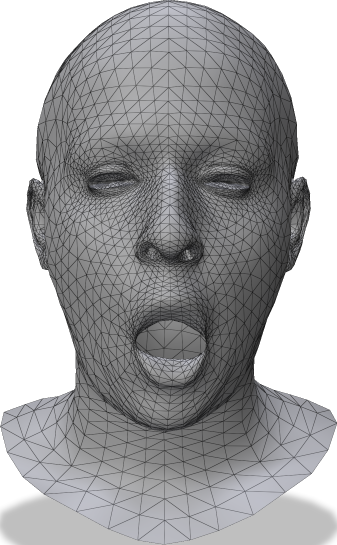} & 
        \includegraphics[scale=\sw]{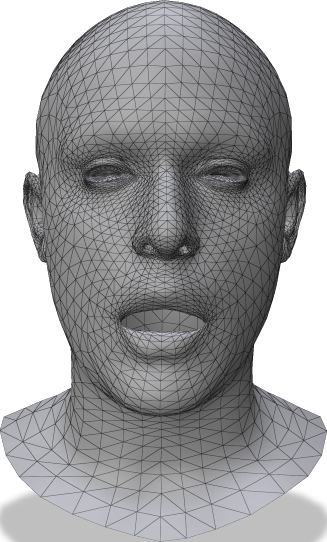} & 
        \includegraphics[scale=\sw]{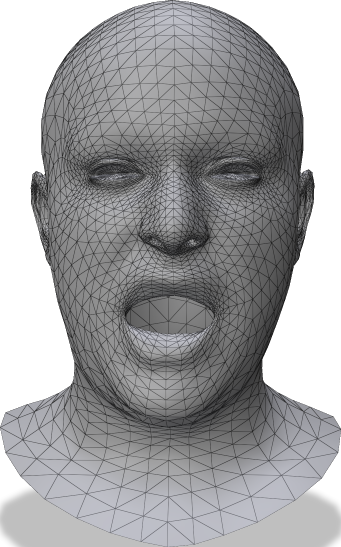} & 
        \includegraphics[scale=\sw]{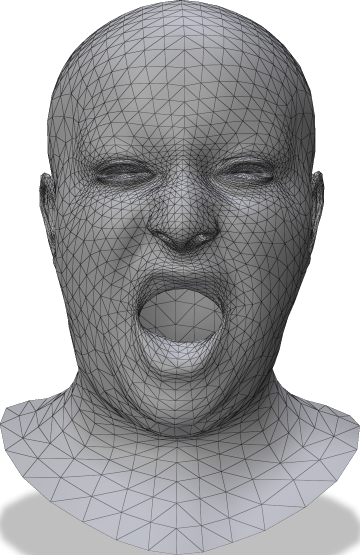} \\
    \end{tabular}\hfill
    \begin{tabular}[b]{cccc}
        \includegraphics[scale=\sw]{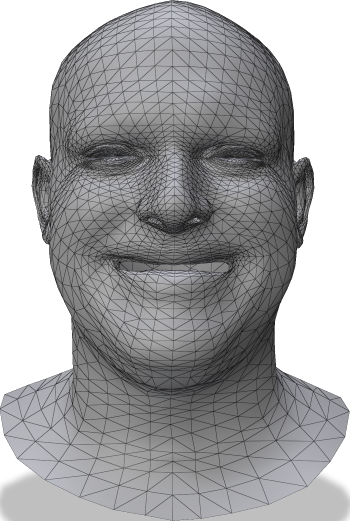} & 
        \includegraphics[scale=\sw]{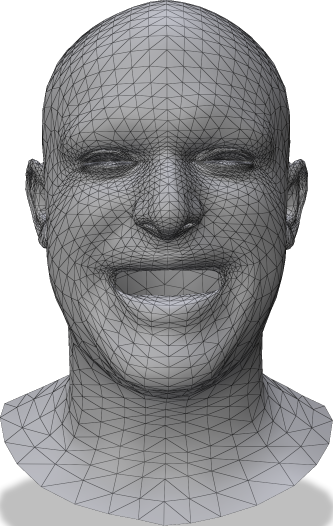} & 
        \includegraphics[scale=\sw]{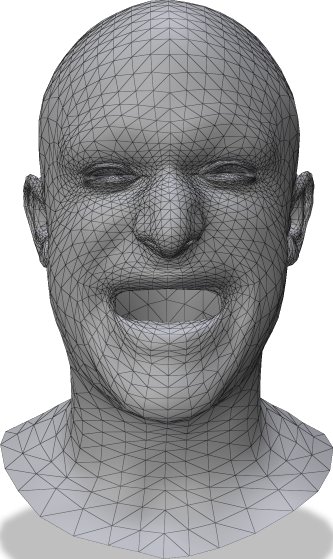} & 
        \includegraphics[scale=\sw]{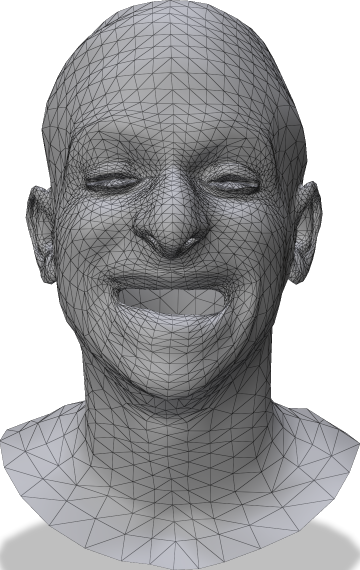} \\
        \includegraphics[scale=\sw]{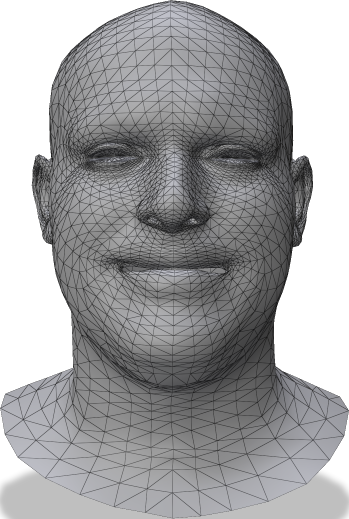} & 
        \includegraphics[scale=\sw]{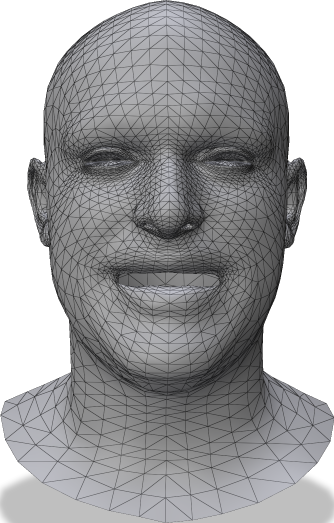} & 
        \includegraphics[scale=\sw]{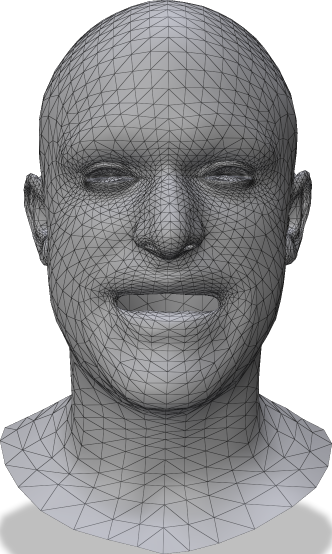} & 
        \includegraphics[scale=\sw]{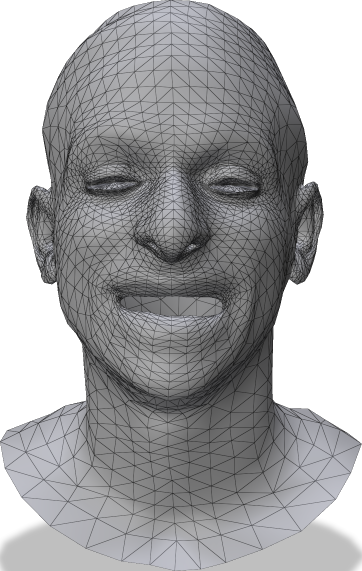} \\
        \includegraphics[scale=\sw]{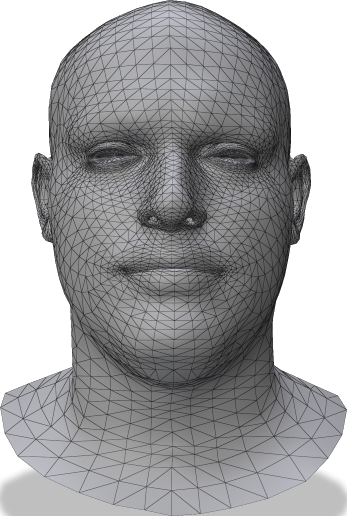} & 
        \includegraphics[scale=\sw]{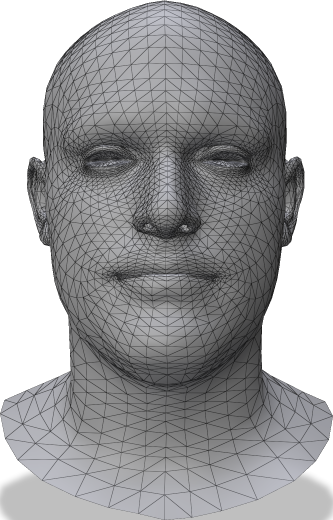} & 
        \includegraphics[scale=\sw]{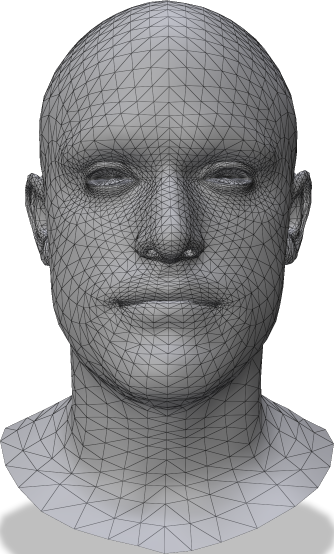} & 
        \includegraphics[scale=\sw]{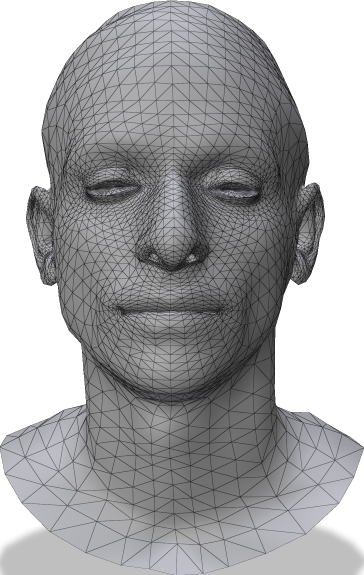} \\
        \includegraphics[scale=\sw]{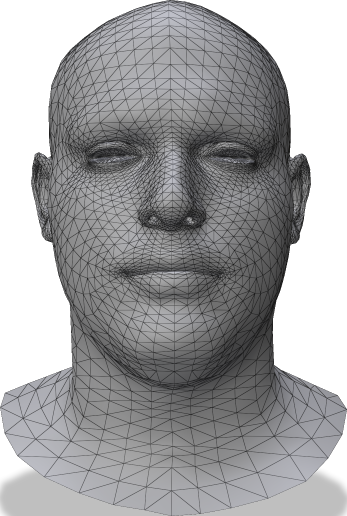} & 
        \includegraphics[scale=\sw]{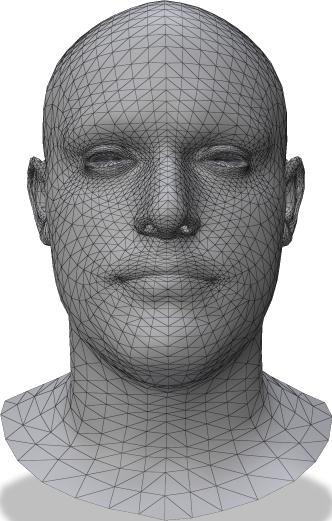} & 
        \includegraphics[scale=\sw]{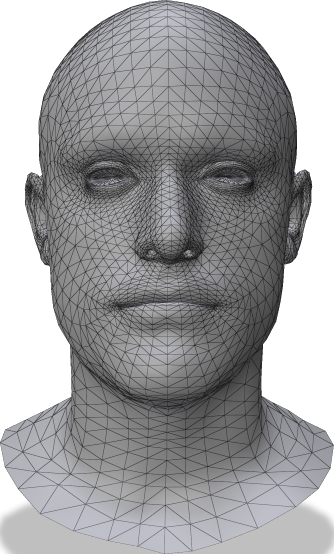} & 
        \includegraphics[scale=\sw]{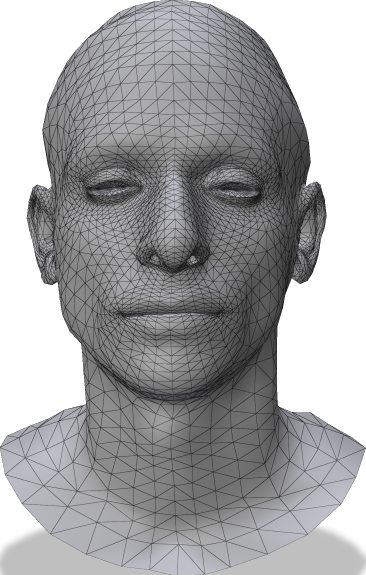} \\
    \end{tabular}
    }
    \caption{
        \newer{
        Example latent interpolations on the CoMA dataset.
        Per inset, horizontal movement corresponds to linear changes in $z_I$, while vertical movement corresponds to linear changes in $z_E$.
        Upper left and lower right images are starting and ending points; upper right and lower left are thus pose (or intrinsics) transfers.
        The model is capable of correctly preserving identity when the intrinsics, $z_I$, are fixed (i.e., within columns), as well as representing subtle expressions (e.g., the orientation of the mouth in the leftmost inset).
        We remark that there are a few failure cases (e.g., bottom row of the third inset, top row of the fourth inset) where the non-rigid pose (facial expression) is not exactly preserved as the latent intrinsics change.
        However note that our formulation demands only that latent shape and extrinsics are disentangled; it does not directly enforce which non-rigid pose {should} semantically correspond across different intrinsics.
        }
    }
    \label{fig:coma.interps}
\end{figure*}

\section{Discussion}
\label{sec:discussion}

In this work, we have devised a method for separating the deformation space of an object into rigid orientation, non-rigid extrinsic pose, and intrinsic shape.
We require no information other than the geometry of the shapes themselves
(i.e., no labels or correspondences).
Our method relies on the isometry invariance of the LBOS, 
	which can be estimated from the geometry directly, 
	and uses %
	disentanglement techniques 
	to partition the latent space of a generative model 
	into these independent components.

In particular, we have built upon the GDVAE model \citep{aumentado2019geometric} with two primary technical improvements.
First, we investigated two approaches to improving rotation factorization: STD, which utilizes randomly rotated inputs to enforce rotation invariance \citep{li2019discrete,sanghitowards},
and
FTL, which provides an interpretable latent space in which 3D rotations in a ``folded'' latent space mirror the effects of those rotations in real-space \citep{worrall2017interpretable,remelli2020lightweight}.
Compared to the GDVAE, which was only able to maintain robustness to small rotations, both new AEs can handle arbitrary rotations about a single axis; the FTL method has the additional benefit of latent interpretability.
Second, we utilized a diffeomorphic normalizing flow network to map between LBOSs and latent intrinsic space.
Unlike the GDVAE, 
which did \textit{not} have a mapping from LBOS space to latent intrinsic space
(and thus could not \textit{architecturally} stop latent intrinsics $z_I$ computed from an encoded shape $x$ from being affecting by extrinsic pose information), 
utilizing this mapping in our GDVAE\PP\ training procedure
(see \S\ref{sec:trainregimes}) allows us to compute reconstructions through $\widetilde{z}_I$ instead, guaranteeing this separation.
Further, the bijectivity of the flow ensures that 
(i) spectral information is not lost and 
(ii) generative likelihood is tractably computable.
Altogether, these changes result in greatly improved unsupervised disentanglement, without sacrificing other representational aspects. 

Our results show that we have significantly improved 
	on the GDVAE.
Firstly, we are able to handle larger orientation changes 
	with far better robustness
	in both the 3D data space and latent space 
	(see \S\ref{sec:aeresults}),
	utilizing rotational invariance techniques 
	that do not rely on a specific feature extraction or neural architecture. 
Secondly, we obtain nearly double the quantitative disentanglement score,
	for data from both SMAL and SMPL, using our GDVAE\PP\ training scheme 
	(see \S\ref{sec:ret}).
We also examined %
	the ability of the model to generate novel shape samples
    (see Fig.~\ref{fig:vaegenerative}),
	its capacity to smoothly and independently control latent shape and non-rigid pose 
		(see Figs.\ \ref{fig:vaeinterp}, \ref{fig:amassinterps}, and \ref{fig:coma.interps}),
	and the effect of several ablations and modifications of the model 
        (see \S\ref{sec:srobust} and \S\ref{sec:ablations}).
\newer{Finally, we compare the GDVAE\PP\  to existing techniques for disentanglement and shape-from-spectrum recovery (see \S\ref{sec:meshexps}).}

For future work,
	we expect research on
	\textit{localized} spectral geometry \citep{neumann2014compressed,melzi2018localized}, 
	LBO modifications \citep{choukroun2018hamiltonian,andreux2014anisotropic},
	and \textit{ex}trinsic spectral shape 
	\citep{liu2017dirac,ye2018unified,wang2017steklov}
	to be potentially useful.
Furthermore, 
	our formulation is readily applicable to other 3D shape modalities 
	(e.g., tetrahderal meshes or implicit fields),
	as the only elements of our architecture 
	that would require alteration are the AE encoders ($E_r$ and $E_x$) and decoder ($D$),
	provided one has a way to estimate the LBOS.
Our VAE model is also agnostic to the neural architecture of the AE.
Hence, our approach could be used in conjunction with other methods 
	for factorizing deformations.
Lastly, 
	our method can also be utilized for applications in computer vision.
For instance, 
	it can be used for controllable shape generation or manipulation, 
	for regularizing visual inference 
	(e.g., by acting as a prior on expected deformation types),
	or for pose-aware shape retrieval.
In general, we hope that our model can serve as an interpretable unsupervised prior for understanding shape deformations. \\

\noindent
\textbf{Acknowledgments}
We are grateful for support from NSERC (CGSD3-534955-2019) and Samsung Research.

{
\small
\bibliographystyle{spbasic}      %
\bibliography{ttaa}   %

\begin{thebibliography}{119}
\providecommand{\natexlab}[1]{#1}
\providecommand{\url}[1]{{#1}}
\providecommand{\urlprefix}{URL }
\expandafter\ifx\csname urlstyle\endcsname\relax
  \providecommand{\doi}[1]{DOI~\discretionary{}{}{}#1}\else
  \providecommand{\doi}{DOI~\discretionary{}{}{}\begingroup
  \urlstyle{rm}\Url}\fi
\providecommand{\eprint}[2][]{\url{#2}}

\bibitem[{Achlioptas et~al.(2017)Achlioptas, Diamanti, Mitliagkas, and
  Guibas}]{achlioptas2018learning}
Achlioptas P, Diamanti O, Mitliagkas I, Guibas L (2017) Learning
  representations and generative models for {3D} point clouds. arXiv preprint
  arXiv:170702392

\bibitem[{Andreux et~al.(2014)Andreux, Rodola, Aubry, and
  Cremers}]{andreux2014anisotropic}
Andreux M, Rodola E, Aubry M, Cremers D (2014) Anisotropic {Laplace-Beltrami}
  operators for shape analysis. In: ECCV

\bibitem[{Aubry et~al.(2011)Aubry, Schlickewei, and Cremers}]{aubry2011wave}
Aubry M, Schlickewei U, Cremers D (2011) The wave kernel signature: A quantum
  mechanical approach to shape analysis. In: ICCV Workshops

\bibitem[{Aumentado-Armstrong et~al.(2019)Aumentado-Armstrong, Tsogkas, Jepson,
  and Dickinson}]{aumentado2019geometric}
Aumentado-Armstrong T, Tsogkas S, Jepson A, Dickinson S (2019) Geometric
  disentanglement for generative latent shape models. In: ICCV

\bibitem[{Ba et~al.(2016)Ba, Kiros, and Hinton}]{ba2016layer}
Ba JL, Kiros JR, Hinton GE (2016) Layer normalization. arXiv preprint
  arXiv:160706450

\bibitem[{Baek et~al.(2015)Baek, Lim, and Lee}]{baek2015isometric}
Baek SY, Lim J, Lee K (2015) Isometric shape interpolation. Computers \&
  Graphics 46:257--263

\bibitem[{Basset et~al.(2020)Basset, Wuhrer, Boyer, and
  Multon}]{basset2020contact}
Basset J, Wuhrer S, Boyer E, Multon F (2020) Contact preserving shape transfer:
  Retargeting motion from one shape to another. Computers \& Graphics

\bibitem[{Bengio et~al.(2013)Bengio, Courville, and
  Vincent}]{bengio2013representation}
Bengio Y, Courville A, Vincent P (2013) Representation learning: A review and
  new perspectives. IEEE transactions on pattern analysis and machine
  intelligence 35(8):1798--1828

\bibitem[{Berkiten et~al.(2017)Berkiten, Halber, Solomon, Ma, Li, and
  Rusinkiewicz}]{berkiten2017learning}
Berkiten S, Halber M, Solomon J, Ma C, Li H, Rusinkiewicz S (2017) Learning
  detail transfer based on geometric features. In: Computer Graphics Forum

\bibitem[{Boscaini et~al.(2015{\natexlab{a}})Boscaini, Eynard, Kourounis, and
  Bronstein}]{boscaini2015shape}
Boscaini D, Eynard D, Kourounis D, Bronstein MM (2015{\natexlab{a}})
  Shape-from-operator: Recovering shapes from intrinsic operators. In: Computer
  Graphics Forum

\bibitem[{Boscaini et~al.(2015{\natexlab{b}})Boscaini, Masci, Melzi, Bronstein,
  Castellani, and Vandergheynst}]{boscaini2015learning}
Boscaini D, Masci J, Melzi S, Bronstein MM, Castellani U, Vandergheynst P
  (2015{\natexlab{b}}) Learning class-specific descriptors for deformable
  shapes using localized spectral convolutional networks. In: Computer Graphics
  Forum

\bibitem[{Bronstein et~al.(2011)Bronstein, Bronstein, Guibas, and
  Ovsjanikov}]{bronstein2011shape}
Bronstein AM, Bronstein MM, Guibas LJ, Ovsjanikov M (2011) Shape google:
  Geometric words and expressions for invariant shape retrieval. ACM
  Transactions on Graphics (TOG) 30(1):1

\bibitem[{Chen et~al.(2019{\natexlab{a}})Chen, Li, Xu, Chen, Wang, and
  Lin}]{chen2019clusternet}
Chen C, Li G, Xu R, Chen T, Wang M, Lin L (2019{\natexlab{a}}) Clusternet: Deep
  hierarchical cluster network with rigorously rotation-invariant
  representation for point cloud analysis. In: CVPR

\bibitem[{Chen et~al.(2019{\natexlab{b}})Chen, Chen, and
  Mitra}]{chen2019unpaired}
Chen X, Chen B, Mitra NJ (2019{\natexlab{b}}) Unpaired point cloud completion
  on real scans using adversarial training. arXiv preprint arXiv:190400069

\bibitem[{Chen et~al.(2019{\natexlab{c}})Chen, Lin, Liu, Qian, and
  Lin}]{chen2019weakly}
Chen X, Lin KY, Liu W, Qian C, Lin L (2019{\natexlab{c}}) Weakly-supervised
  discovery of geometry-aware representation for {3D} human pose estimation.
  In: CVPR

\bibitem[{Chen et~al.(2019{\natexlab{d}})Chen, Song, and
  Hilliges}]{chen2019monocular}
Chen X, Song J, Hilliges O (2019{\natexlab{d}}) Monocular neural image based
  rendering with continuous view control. In: ICCV

\bibitem[{Chern et~al.(2018)Chern, Kn{\"o}ppel, Pinkall, and
  Schr{\"o}der}]{chern2018shape}
Chern A, Kn{\"o}ppel F, Pinkall U, Schr{\"o}der P (2018) Shape from metric. ACM
  Transactions on Graphics (TOG) 37(4):1--17

\bibitem[{Choukroun et~al.(2018)Choukroun, Shtern, Bronstein, and
  Kimmel}]{choukroun2018hamiltonian}
Choukroun Y, Shtern A, Bronstein AM, Kimmel R (2018) Hamiltonian operator for
  spectral shape analysis. IEEE transactions on visualization and computer
  graphics

\bibitem[{Chu and Golub(2005)}]{chu2005inverse}
Chu M, Golub G (2005) Inverse eigenvalue problems: theory, algorithms, and
  applications. OUP Oxford

\bibitem[{Chua and Jarvis(1997)}]{chua1997point}
Chua CS, Jarvis R (1997) Point signatures: A new representation for {3D} object
  recognition. International Journal of Computer Vision 25(1):63--85

\bibitem[{Cohen et~al.(2018)Cohen, Geiger, K{\"o}hler, and
  Welling}]{cohen2018spherical}
Cohen TS, Geiger M, K{\"o}hler J, Welling M (2018) Spherical cnns. arXiv
  preprint arXiv:180110130

\bibitem[{Corman et~al.(2017)Corman, Solomon, Ben-Chen, Guibas, and
  Ovsjanikov}]{corman2017functional}
Corman E, Solomon J, Ben-Chen M, Guibas L, Ovsjanikov M (2017) Functional
  characterization of intrinsic and extrinsic geometry. ACM Transactions on
  Graphics (TOG) 36(2):14

\bibitem[{Cosmo et~al.(2019)Cosmo, Panine, Rampini, Ovsjanikov, Bronstein, and
  Rodol{\`a}}]{cosmo2019isospectralization}
Cosmo L, Panine M, Rampini A, Ovsjanikov M, Bronstein MM, Rodol{\`a} E (2019)
  Isospectralization, or how to hear shape, style, and correspondence. In: CVPR

\bibitem[{Cosmo et~al.(2020)Cosmo, Norelli, Halimi, Kimmel, and
  Rodol{\`a}}]{cosmo2020limp}
Cosmo L, Norelli A, Halimi O, Kimmel R, Rodol{\`a} E (2020) Limp: Learning
  latent shape representations with metric preservation priors. arXiv preprint
  arXiv:200312283

\bibitem[{Dinh et~al.(2014)Dinh, Krueger, and Bengio}]{dinh2014nice}
Dinh L, Krueger D, Bengio Y (2014) {NICE}: Non-linear independent components
  estimation. arXiv preprint arXiv:14108516

\bibitem[{Dinh et~al.(2016)Dinh, Sohl-Dickstein, and Bengio}]{dinh2016density}
Dinh L, Sohl-Dickstein J, Bengio S (2016) Density estimation using real {NVP}.
  arXiv preprint arXiv:160508803

\bibitem[{Durkan et~al.(2020)Durkan, Bekasov, Murray, and
  Papamakarios}]{nflows}
Durkan C, Bekasov A, Murray I, Papamakarios G (2020) nflows: normalizing flows
  in {PyTorch}. Zenodo \doi{10.5281/zenodo.4296287},
  \urlprefix\url{https://doi.org/10.5281/zenodo.4296287}

\bibitem[{Dym and Maron(2020)}]{dym2020universality}
Dym N, Maron H (2020) On the universality of rotation equivariant point cloud
  networks. arXiv preprint arXiv:201002449

\bibitem[{Esmaeili et~al.(2018)Esmaeili, Wu, Jain, Bozkurt, Siddharth, Paige,
  Brooks, Dy, and van~de Meent}]{esmaeili2018structured}
Esmaeili B, Wu H, Jain S, Bozkurt A, Siddharth N, Paige B, Brooks DH, Dy J,
  van~de Meent JW (2018) Structured disentangled representations. arXiv
  preprint arXiv:180402086

\bibitem[{Fuchs et~al.(2020)Fuchs, Worrall, Fischer, and Welling}]{fuchs2020se}
Fuchs FB, Worrall DE, Fischer V, Welling M (2020) {SE(3)}-transformers: {3D}
  roto-translation equivariant attention networks. arXiv preprint
  arXiv:200610503

\bibitem[{Fumero et~al.(2021)Fumero, Cosmo, Melzi, and
  Rodol{\`a}}]{fumero2021learning}
Fumero M, Cosmo L, Melzi S, Rodol{\`a} E (2021) Learning disentangled
  representations via product manifold projection. In: ICML

\bibitem[{Gao et~al.(2018)Gao, Yang, Qiao, Lai, Rosin, Xu, and
  Xia}]{gao2018automatic}
Gao L, Yang J, Qiao YL, Lai YK, Rosin PL, Xu W, Xia S (2018) Automatic unpaired
  shape deformation transfer. ACM Transactions on Graphics (TOG) 37(6):1--15

\bibitem[{Gebal et~al.(2009)Gebal, B{\ae}rentzen, Aan{\ae}s, and
  Larsen}]{gebal2009shape}
Gebal K, B{\ae}rentzen JA, Aan{\ae}s H, Larsen R (2009) Shape analysis using
  the auto diffusion function. In: Computer Graphics Forum

\bibitem[{Ghosh et~al.(2019)Ghosh, Sajjadi, Vergari, Black, and
  Sch{\"o}lkopf}]{ghosh2019variational}
Ghosh P, Sajjadi MS, Vergari A, Black M, Sch{\"o}lkopf B (2019) From
  variational to deterministic autoencoders. arXiv preprint arXiv:190312436

\bibitem[{Gordon et~al.(1992)Gordon, Webb, and Wolpert}]{gordon1992one}
Gordon C, Webb DL, Wolpert S (1992) One cannot hear the shape of a drum.
  Bulletin of the American Mathematical Society 27(1):134--138

\bibitem[{Groueix et~al.(2018)Groueix, Fisher, Kim, Russell, and
  Aubry}]{groueix20183d}
Groueix T, Fisher M, Kim VG, Russell BC, Aubry M (2018) {3D-CODED}: {3D}
  correspondences by deep deformation. In: ECCV

\bibitem[{Guo et~al.(2014)Guo, Bennamoun, Sohel, Lu, and Wan}]{guo20143d}
Guo Y, Bennamoun M, Sohel F, Lu M, Wan J (2014) {3D} object recognition in
  cluttered scenes with local surface features: a survey. IEEE Transactions on
  Pattern Analysis and Machine Intelligence 36(11):2270--2287

\bibitem[{Higgins et~al.(2017)Higgins, Matthey, Pal, Burgess, Glorot,
  Botvinick, Mohamed, and Lerchner}]{higgins2017beta}
Higgins I, Matthey L, Pal A, Burgess C, Glorot X, Botvinick M, Mohamed S,
  Lerchner A (2017) $\beta$-{VAE}: Learning basic visual concepts with a
  constrained variational framework. In: ICLR

\bibitem[{Huang et~al.(2019)Huang, Rakotosaona, Achlioptas, Guibas, and
  Ovsjanikov}]{huang2019operatornet}
Huang R, Rakotosaona MJ, Achlioptas P, Guibas LJ, Ovsjanikov M (2019)
  {Operatornet}: Recovering {3D} shapes from difference operators. In: ICCV

\bibitem[{Huynh(2009)}]{huynh2009metrics}
Huynh DQ (2009) Metrics for {3D} rotations: Comparison and analysis. Journal of
  Mathematical Imaging and Vision 35(2):155--164

\bibitem[{Johnson and Hebert(1999)}]{johnson1999using}
Johnson AE, Hebert M (1999) Using spin images for efficient object recognition
  in cluttered {3D} scenes. IEEE Transactions on pattern analysis and machine
  intelligence 21(5):433--449

\bibitem[{Kac(1966)}]{kac1966can}
Kac M (1966) Can one hear the shape of a drum? The american mathematical
  monthly 73(4P2):1--23

\bibitem[{Kingma and Ba(2014)}]{kingma2014adam}
Kingma DP, Ba J (2014) Adam: A method for stochastic optimization. arXiv
  preprint arXiv:14126980

\bibitem[{Kingma and Dhariwal(2018)}]{kingma2018glow}
Kingma DP, Dhariwal P (2018) Glow: Generative flow with invertible 1x1
  convolutions. In: NeurIPS

\bibitem[{Kingma et~al.(2016)Kingma, Salimans, Jozefowicz, Chen, Sutskever, and
  Welling}]{kingma2016improved}
Kingma DP, Salimans T, Jozefowicz R, Chen X, Sutskever I, Welling M (2016)
  Improved variational inference with inverse autoregressive flow. NeurIPS

\bibitem[{Klambauer et~al.(2017)Klambauer, Unterthiner, Mayr, and
  Hochreiter}]{klambauer2017self}
Klambauer G, Unterthiner T, Mayr A, Hochreiter S (2017) Self-normalizing neural
  networks. NeurIPS

\bibitem[{Kobyzev et~al.(2020)Kobyzev, Prince, and
  Brubaker}]{kobyzev2020normalizing}
Kobyzev I, Prince S, Brubaker M (2020) Normalizing flows: An introduction and
  review of current methods. IEEE Transactions on Pattern Analysis and Machine
  Intelligence

\bibitem[{Kondor et~al.(2018)Kondor, Son, Pan, Anderson, and
  Trivedi}]{kondor2018covariant}
Kondor R, Son HT, Pan H, Anderson B, Trivedi S (2018) Covariant compositional
  networks for learning graphs. arXiv preprint arXiv:180102144

\bibitem[{Kovnatsky et~al.(2013)Kovnatsky, Bronstein, Bronstein, Glashoff, and
  Kimmel}]{kovnatsky2013coupled}
Kovnatsky A, Bronstein MM, Bronstein AM, Glashoff K, Kimmel R (2013) Coupled
  quasi-harmonic bases. In: Computer Graphics Forum

\bibitem[{Kumar et~al.(2017)Kumar, Sattigeri, and
  Balakrishnan}]{kumar2017variational}
Kumar A, Sattigeri P, Balakrishnan A (2017) Variational inference of
  disentangled latent concepts from unlabeled observations. arXiv preprint
  arXiv:171100848

\bibitem[{LeCun et~al.(1998)LeCun, Bottou, Bengio, and
  Haffner}]{lecun1998gradient}
LeCun Y, Bottou L, Bengio Y, Haffner P (1998) Gradient-based learning applied
  to document recognition. Proceedings of the IEEE 86(11):2278--2324

\bibitem[{Levinson et~al.(2019)Levinson, Sud, and Makadia}]{levinson2019latent}
Levinson J, Sud A, Makadia A (2019) Latent feature disentanglement for {3D}
  meshes. arXiv preprint arXiv:190603281

\bibitem[{L{\'e}vy(2006)}]{levy2006laplace}
L{\'e}vy B (2006) {Laplace-Beltrami} eigenfunctions: towards an algorithm that
  understands geometry. In: Shape Modeling and Applications, 2006. SMI 2006.
  IEEE International Conference on, IEEE, pp 13--13

\bibitem[{Li et~al.(2019)Li, Bi, and Lee}]{li2019discrete}
Li J, Bi Y, Lee GH (2019) Discrete rotation equivariance for point cloud
  recognition. In: ICRA

\bibitem[{Liu et~al.(2017)Liu, Jacobson, and Crane}]{liu2017dirac}
Liu HTD, Jacobson A, Crane K (2017) A {Dirac} operator for extrinsic shape
  analysis. In: Computer Graphics Forum

\bibitem[{Loper et~al.(2015)Loper, Mahmood, Romero, Pons-Moll, and
  Black}]{SMPL:2015}
Loper M, Mahmood N, Romero J, Pons-Moll G, Black MJ (2015) {SMPL}: A skinned
  multi-person linear model. ACM Trans Graphics (Proc SIGGRAPH Asia)
  34(6):248:1--248:16

\bibitem[{Loshchilov and Hutter(2017)}]{loshchilov2017decoupled}
Loshchilov I, Hutter F (2017) Decoupled weight decay regularization. arXiv
  preprint arXiv:171105101

\bibitem[{van~der Maaten and Hinton(2008)}]{maaten2008visualizing}
van~der Maaten L, Hinton G (2008) Visualizing data using t-sne. Journal of
  machine learning research 9(Nov):2579--2605

\bibitem[{MacQueen et~al.(1967)}]{macqueen1967some}
MacQueen J, et~al. (1967) Some methods for classification and analysis of
  multivariate observations. In: Proceedings of the fifth Berkeley symposium on
  mathematical statistics and probability, Oakland, CA, USA

\bibitem[{Mahmood et~al.(2019)Mahmood, Ghorbani, Troje, Pons-Moll, and
  Black}]{AMASS:ICCV:2019}
Mahmood N, Ghorbani N, Troje NF, Pons-Moll G, Black MJ (2019) {AMASS}: Archive
  of motion capture as surface shapes. In: ICCV

\bibitem[{Marin et~al.(2020)Marin, Rampini, Castellani, Rodola, Ovsjanikov, and
  Melzi}]{marin2020instant}
Marin R, Rampini A, Castellani U, Rodola E, Ovsjanikov M, Melzi S (2020)
  Instant recovery of shape from spectrum via latent space connections. In:
  2020 International Conference on 3D Vision (3DV), IEEE, pp 120--129

\bibitem[{Marin et~al.(2021)Marin, Rampini, Castellani, Rodol{\`a}, Ovsjanikov,
  and Melzi}]{marin2021spectral}
Marin R, Rampini A, Castellani U, Rodol{\`a} E, Ovsjanikov M, Melzi S (2021)
  Spectral shape recovery and analysis via data-driven connections.
  International Journal of Computer Vision pp 1--16

\bibitem[{Masoumi and Hamza(2017)}]{masoumi2017spectral}
Masoumi M, Hamza AB (2017) Spectral shape classification: A deep learning
  approach. Journal of Visual Communication and Image Representation
  43:198--211

\bibitem[{Melzi et~al.(2018)Melzi, Rodol{\`a}, Castellani, and
  Bronstein}]{melzi2018localized}
Melzi S, Rodol{\`a} E, Castellani U, Bronstein MM (2018) Localized manifold
  harmonics for spectral shape analysis. In: Computer Graphics Forum

\bibitem[{Meyer et~al.(2003)Meyer, Desbrun, Schr{\"o}der, and
  Barr}]{meyer2003discrete}
Meyer M, Desbrun M, Schr{\"o}der P, Barr AH (2003) Discrete
  differential-geometry operators for triangulated 2-manifolds. In:
  Visualization and mathematics III, Springer, pp 35--57

\bibitem[{Moschella et~al.(2022)Moschella, Melzi, Cosmo, Maggioli, Litany,
  Ovsjanikov, Guibas, and Rodol{\`a}}]{moschella2022learning}
Moschella L, Melzi S, Cosmo L, Maggioli F, Litany O, Ovsjanikov M, Guibas L,
  Rodol{\`a} E (2022) Learning spectral unions of partial deformable {3D}
  shapes. In: Computer Graphics Forum

\bibitem[{Narayanaswamy et~al.(2017)Narayanaswamy, Paige, Van~de Meent,
  Desmaison, Goodman, Kohli, Wood, and Torr}]{siddharth2017learning}
Narayanaswamy S, Paige B, Van~de Meent JW, Desmaison A, Goodman N, Kohli P,
  Wood F, Torr P (2017) Learning disentangled representations with
  semi-supervised deep generative models. In: NeurIPS

\bibitem[{Neumann et~al.(2014)Neumann, Varanasi, Theobalt, Magnor, and
  Wacker}]{neumann2014compressed}
Neumann T, Varanasi K, Theobalt C, Magnor M, Wacker M (2014) Compressed
  manifold modes for mesh processing. In: Computer Graphics Forum

\bibitem[{Ovsjanikov et~al.(2012)Ovsjanikov, Ben-Chen, Solomon, Butscher, and
  Guibas}]{ovsjanikov2012functional}
Ovsjanikov M, Ben-Chen M, Solomon J, Butscher A, Guibas L (2012) Functional
  maps: a flexible representation of maps between shapes. ACM Transactions on
  Graphics (TOG) 31(4):1--11

\bibitem[{Panine and Kempf(2016)}]{panine2016towards}
Panine M, Kempf A (2016) Towards spectral geometric methods for euclidean
  quantum gravity. Physical Review D 93(8):084033

\bibitem[{Papamakarios et~al.(2019)Papamakarios, Nalisnick, Rezende, Mohamed,
  and Lakshminarayanan}]{papamakarios2019normalizing}
Papamakarios G, Nalisnick E, Rezende DJ, Mohamed S, Lakshminarayanan B (2019)
  Normalizing flows for probabilistic modeling and inference. arXiv preprint
  arXiv:191202762

\bibitem[{Paszke et~al.(2019)Paszke, Gross, Massa, Lerer, Bradbury, Chanan,
  Killeen, Lin, Gimelshein, Antiga, Desmaison, Kopf, Yang, DeVito, Raison,
  Tejani, Chilamkurthy, Steiner, Fang, Bai, and Chintala}]{pytorch}
Paszke A, Gross S, Massa F, Lerer A, Bradbury J, Chanan G, Killeen T, Lin Z,
  Gimelshein N, Antiga L, Desmaison A, Kopf A, Yang E, DeVito Z, Raison M,
  Tejani A, Chilamkurthy S, Steiner B, Fang L, Bai J, Chintala S (2019)
  Pytorch: An imperative style, high-performance deep learning library. NeurIPS

\bibitem[{Patan{\'e}(2016)}]{patane2016star}
Patan{\'e} G (2016) Star-laplacian spectral kernels and distances for geometry
  processing and shape analysis. In: Computer Graphics Forum

\bibitem[{Pedregosa et~al.(2012)Pedregosa, Varoquaux, Gramfort, Michel,
  Thirion, Grisel, Blondel, Prettenhofer, Weiss, Dubourg, VanderPlas, Passos,
  Cournapeau, Brucher, Perrot, and Duchesnay}]{scikit-learn}
Pedregosa F, Varoquaux G, Gramfort A, Michel V, Thirion B, Grisel O, Blondel M,
  Prettenhofer P, Weiss R, Dubourg V, VanderPlas J, Passos A, Cournapeau D,
  Brucher M, Perrot M, Duchesnay E (2012) Scikit-learn: Machine learning in
  python. CoRR abs/1201.0490, \urlprefix\url{http://arxiv.org/abs/1201.0490},
  \eprint{1201.0490}

\bibitem[{Pons-Moll et~al.(2015)Pons-Moll, Romero, Mahmood, and Black}]{dyna}
Pons-Moll G, Romero J, Mahmood N, Black MJ (2015) Dyna: A model of dynamic
  human shape in motion. ACM Transactions on Graphics, (Proc SIGGRAPH)
  34(4):120:1--120:14

\bibitem[{Poulenard et~al.(2019)Poulenard, Rakotosaona, Ponty, and
  Ovsjanikov}]{poulenard2019effective}
Poulenard A, Rakotosaona MJ, Ponty Y, Ovsjanikov M (2019) Effective
  rotation-invariant point cnn with spherical harmonics kernels. arXiv preprint
  arXiv:190611555

\bibitem[{Qi et~al.(2017)Qi, Su, Mo, and Guibas}]{qi2017pointnet}
Qi CR, Su H, Mo K, Guibas LJ (2017) {Pointnet}: Deep learning on point sets for
  {3D} classification and segmentation. CVPR

\bibitem[{Rampini et~al.(2021)Rampini, Pestarini, Cosmo, Melzi, and
  Rodola}]{rampini2021universal}
Rampini A, Pestarini F, Cosmo L, Melzi S, Rodola E (2021) Universal spectral
  adversarial attacks for deformable shapes. In: CVPR

\bibitem[{Ranjan et~al.(2018)Ranjan, Bolkart, Sanyal, and Black}]{COMA:ECCV18}
Ranjan A, Bolkart T, Sanyal S, Black MJ (2018) Generating {3D} faces using
  convolutional mesh autoencoders. In: ECCV

\bibitem[{Remelli et~al.(2020)Remelli, Han, Honari, Fua, and
  Wang}]{remelli2020lightweight}
Remelli E, Han S, Honari S, Fua P, Wang R (2020) Lightweight multi-view {3D}
  pose estimation through camera-disentangled representation. In: CVPR

\bibitem[{Reuter(2010)}]{reuter2010hierarchical}
Reuter M (2010) Hierarchical shape segmentation and registration via
  topological features of {Laplace-Beltrami} eigenfunctions. International
  Journal of Computer Vision 89(2-3):287--308

\bibitem[{Reuter et~al.(2006)Reuter, Wolter, and Peinecke}]{reuter2006laplace}
Reuter M, Wolter FE, Peinecke N (2006) {Laplace--Beltrami} spectra as
  {Shape-DNA} of surfaces and solids. Computer-Aided Design 38(4):342--366

\bibitem[{Rhodin et~al.(2018)Rhodin, Salzmann, and
  Fua}]{rhodin2018unsupervised}
Rhodin H, Salzmann M, Fua P (2018) Unsupervised geometry-aware representation
  for {3D} human pose estimation. In: ECCV

\bibitem[{Rhodin et~al.(2019)Rhodin, Constantin, Katircioglu, Salzmann, and
  Fua}]{rhodin2019neural}
Rhodin H, Constantin V, Katircioglu I, Salzmann M, Fua P (2019) Neural scene
  decomposition for multi-person motion capture. In: CVPR

\bibitem[{Roberts et~al.(2020)Roberts, dos Anjos, Maejima, and
  Anjyo}]{roberts2020deformation}
Roberts RA, dos Anjos RK, Maejima A, Anjyo K (2020) Deformation transfer
  survey. Computers \& Graphics

\bibitem[{Rodol{\`a} et~al.(2017)Rodol{\`a}, Cosmo, Bronstein, Torsello, and
  Cremers}]{rodola2017partial}
Rodol{\`a} E, Cosmo L, Bronstein MM, Torsello A, Cremers D (2017) Partial
  functional correspondence. In: Computer Graphics Forum

\bibitem[{Rustamov(2007)}]{rustamov2007laplace}
Rustamov RM (2007) {Laplace-Beltrami} eigenfunctions for deformation invariant
  shape representation. In: Proceedings of the fifth Eurographics symposium on
  Geometry processing, Eurographics Association, pp 225--233

\bibitem[{Sanghi(2020)}]{sanghi2020info3d}
Sanghi A (2020) {Info3D}: Representation learning on {3D} objects using mutual
  information maximization and contrastive learning. arXiv preprint
  arXiv:200602598

\bibitem[{Sanghi and Danielyan(2019)}]{sanghitowards}
Sanghi A, Danielyan A (2019) Towards {3D} rotation invariant embeddings. CVPR
  2019 Workshop on 3D Scene Understanding for Vision, Graphics, and Robotics

\bibitem[{Sharp and Crane(2020)}]{sharp2020laplacian}
Sharp N, Crane K (2020) A {Laplacian} for nonmanifold triangle meshes. In:
  Computer Graphics Forum

\bibitem[{Sharp et~al.(2021)Sharp, Gillespie, and Crane}]{sharp2021geometry}
Sharp N, Gillespie M, Crane K (2021) Geometry processing with intrinsic
  triangulations. SIGGRAPH'21: ACM SIGGRAPH 2021 Courses

\bibitem[{Shoemake(1985)}]{shoemake1985animating}
Shoemake K (1985) Animating rotation with quaternion curves. In: Proceedings of
  the 12th annual conference on Computer graphics and interactive techniques,
  pp 245--254

\bibitem[{Stein et~al.(1992)Stein, Medioni et~al.}]{stein1992structural}
Stein F, Medioni G, et~al. (1992) Structural indexing: Efficient {3-D} object
  recognition. IEEE Transactions on Pattern Analysis and Machine Intelligence
  14(2):125--145

\bibitem[{Su et~al.(2021)Su, Lin, and Wang}]{su2021learning}
Su FG, Lin CS, Wang YCF (2021) Learning interpretable representation for {3D}
  point clouds. In: ICPR, IEEE

\bibitem[{Sumner and Popovi{\'c}(2004)}]{sumner2004deformation}
Sumner RW, Popovi{\'c} J (2004) Deformation transfer for triangle meshes. ACM
  Transactions on graphics (TOG) 23(3):399--405

\bibitem[{Sun et~al.(2009)Sun, Ovsjanikov, and Guibas}]{sun2009concise}
Sun J, Ovsjanikov M, Guibas L (2009) A concise and provably informative
  multi-scale signature based on heat diffusion. In: Computer Graphics Forum

\bibitem[{Sun et~al.(2019)Sun, Lian, and Xiao}]{sun2019srinet}
Sun X, Lian Z, Xiao J (2019) {SRINet}: Learning strictly rotation-invariant
  representations for point cloud classification and segmentation. In:
  Proceedings of the 27th ACM International Conference on Multimedia, pp
  980--988

\bibitem[{Tan et~al.(2018)Tan, Gao, Lai, and Xia}]{tan2018variational}
Tan Q, Gao L, Lai YK, Xia S (2018) Variational autoencoders for deforming {3D}
  mesh models. In: CVPR

\bibitem[{Taubin(1995)}]{taubin1995signal}
Taubin G (1995) A signal processing approach to fair surface design. In:
  Proceedings of the 22nd annual conference on Computer graphics and
  interactive techniques, pp 351--358

\bibitem[{Thomas et~al.(2018)Thomas, Smidt, Kearnes, Yang, Li, Kohlhoff, and
  Riley}]{thomas2018tensor}
Thomas N, Smidt T, Kearnes S, Yang L, Li L, Kohlhoff K, Riley P (2018) Tensor
  field networks: Rotation-and translation-equivariant neural networks for {3D}
  point clouds. arXiv preprint arXiv:180208219

\bibitem[{Tombari et~al.(2010)Tombari, Salti, and
  Di~Stefano}]{tombari2010unique}
Tombari F, Salti S, Di~Stefano L (2010) Unique signatures of histograms for
  local surface description. In: ECCV

\bibitem[{Vallet and L{\'e}vy(2008)}]{vallet2008spectral}
Vallet B, L{\'e}vy B (2008) Spectral geometry processing with manifold
  harmonics. In: Computer Graphics Forum

\bibitem[{Varol et~al.(2017)Varol, Romero, Martin, Mahmood, Black, Laptev, and
  Schmid}]{varol17_surreal}
Varol G, Romero J, Martin X, Mahmood N, Black MJ, Laptev I, Schmid C (2017)
  Learning from synthetic humans. In: CVPR

\bibitem[{Vinh et~al.(2010)Vinh, Epps, and Bailey}]{vinh2010information}
Vinh NX, Epps J, Bailey J (2010) Information theoretic measures for clusterings
  comparison: Variants, properties, normalization and correction for chance.
  The Journal of Machine Learning Research 11:2837--2854

\bibitem[{Wang et~al.(2017)Wang, Ben-Chen, Polterovich, and
  Solomon}]{wang2017steklov}
Wang Y, Ben-Chen M, Polterovich I, Solomon J (2017) Steklov spectral geometry
  for extrinsic shape analysis. arXiv preprint arXiv:170707070

\bibitem[{Watanabe(1960)}]{watanabe1960information}
Watanabe S (1960) Information theoretical analysis of multivariate correlation.
  IBM Journal of research and development 4(1):66--82

\bibitem[{Weyl(1911)}]{weyl1911asymptotische}
Weyl H (1911) {\"U}ber die asymptotische verteilung der eigenwerte. Nachrichten
  von der Gesellschaft der Wissenschaften zu G{\"o}ttingen,
  Mathematisch-Physikalische Klasse 1911:110--117

\bibitem[{Worrall and Brostow(2018)}]{worrall2018cubenet}
Worrall D, Brostow G (2018) {CubeNet}: Equivariance to {3D} rotation and
  translation. In: ECCV

\bibitem[{Worrall et~al.(2017)Worrall, Garbin, Turmukhambetov, and
  Brostow}]{worrall2017interpretable}
Worrall DE, Garbin SJ, Turmukhambetov D, Brostow GJ (2017) Interpretable
  transformations with encoder-decoder networks. In: ICCV

\bibitem[{Xiao et~al.(2020)Xiao, Lin, Li, Geng, Chao, and
  Ding}]{xiao2020endowing}
Xiao Z, Lin H, Li R, Geng L, Chao H, Ding S (2020) Endowing deep {3D} models
  with rotation invariance based on principal component analysis. In: 2020 IEEE
  International Conference on Multimedia and Expo (ICME)

\bibitem[{Yang et~al.(2019)Yang, Huang, Hao, Liu, Belongie, and
  Hariharan}]{yang2019pointflow}
Yang G, Huang X, Hao Z, Liu MY, Belongie S, Hariharan B (2019) {PointFlow}:
  {3D} point cloud generation with continuous normalizing flows. arXiv preprint
  arXiv:190612320

\bibitem[{Ye et~al.(2018)Ye, Diamanti, Tang, Guibas, and
  Hoffmann}]{ye2018unified}
Ye Z, Diamanti O, Tang C, Guibas L, Hoffmann T (2018) A unified discrete
  framework for intrinsic and extrinsic {Dirac} operators for geometry
  processing. In: Computer Graphics Forum

\bibitem[{Yin et~al.(2015)Yin, Li, Lu, Ouyang, Zhang, and
  Xian}]{yin2015spectral}
Yin M, Li G, Lu H, Ouyang Y, Zhang Z, Xian C (2015) Spectral pose transfer.
  Computer Aided Geometric Design 35:82--94

\bibitem[{You et~al.(2018)You, Lou, Liu, Tai, Ma, Lu, and
  Wang}]{you2018pointwise}
You Y, Lou Y, Liu Q, Tai YW, Ma L, Lu C, Wang W (2018) Pointwise
  rotation-invariant network with adaptive sampling and {3D} spherical voxel
  convolution. arXiv preprint arXiv:181109361

\bibitem[{Zhang et~al.(2020)Zhang, Qin, Xu, and Xu}]{zhang2020quaternion}
Zhang X, Qin S, Xu Y, Xu H (2020) Quaternion product units for deep learning on
  {3D} rotation groups. In: CVPR

\bibitem[{Zhang et~al.(2019)Zhang, Hua, Rosen, and Yeung}]{zhang-riconv-3dv19}
Zhang Z, Hua BS, Rosen DW, Yeung SK (2019) Rotation invariant convolutions for
  {3D} point clouds deep learning. In: International Conference on 3D Vision
  (3DV)

\bibitem[{Zhao et~al.(2020)Zhao, Birdal, Lenssen, Menegatti, Guibas, and
  Tombari}]{zhao2020quaternion}
Zhao Y, Birdal T, Lenssen JE, Menegatti E, Guibas L, Tombari F (2020)
  Quaternion equivariant capsule networks for {3D} point clouds. In: ECCV

\bibitem[{Zhou et~al.(2020)Zhou, Bhatnagar, and
  Pons-Moll}]{zhou2020unsupervised}
Zhou K, Bhatnagar BL, Pons-Moll G (2020) Unsupervised shape and pose
  disentanglement for {3D} meshes. In: ECCV

\bibitem[{Zuffi et~al.(2017)Zuffi, Kanazawa, Jacobs, and
  Black}]{Zuffi:CVPR:2017}
Zuffi S, Kanazawa A, Jacobs D, Black MJ (2017) {3D} menagerie: Modeling the
  {3D} shape and pose of animals. In: CVPR

\end{thebibliography}
}

\appendix

\section{Glossary of Notation}
\label{app:glossary}
\begin{center}
\begin{tabular}{l|l|l} %
    Symbol & Sec/Eq & Definition \\\hline
    $P$   & \S\ref{sec:ae} & Shape \\
    $x_c$ & \S\ref{sec:ae} & Canonical AE encoding \\
    $\widetilde{x}$ & \S\ref{sec:ae:ftl} & Non-canon FTL-AE  encoding \\
    ${q}$ & \S\ref{sec:ae} & Quaternion \\
    $D_P$ & Eq.\ \ref{eq:aereconloss} & Distance between PCs \\
    $z_I$ & \S\ref{sec:hfvae} & Latent intrinsics \\
    $z_E$ & \S\ref{sec:hfvae} & Latent extrinsics \\
    $z_R$ & \S\ref{sec:hfvae} & Latent rigid pose \\
    $\lambda$ & \S\ref{sec:relwork:spec} & LBO Spectrum \\
    $\widetilde{z}_I$ & \S\ref{sec:normflowforspectrum} & Latent intrinsics (from $\lambda$) \\ %
    $f_\lambda$ & \S\ref{sec:normflowforspectrum} & Spectral flow network \\
    $g_\lambda$ & \S\ref{sec:normflowforspectrum} & Inverse of $f_\lambda$ \\
    $\mathcal{S}$ & Eq.\ \ref{eq:disentscore} & Disentanglement score \\ 
    $E_\beta$ & \S\ref{sec:ret} & Retrieval error wrt SMPL shape \\
    $E_\theta$ & \S\ref{sec:ret} & Retrieval error wrt SMPL pose \\
    $P_\lambda(\lambda)$ & Eq.\ \ref{eq:plambda} & Spectral likelihood \\
    $d_\lambda$ & Eq.\ \ref{eq:specdist} & Distance between spectra \\
    $d_R$ & Eq.\ \ref{eq:rotdist} & Distance between rotations \\
    $\mathfrak{L}_\text{AE}$ & \S\ref{sec:ae:loss} & AE total loss \\
    $\mathcal{L}_c$ & \S\ref{sec:ae:loss:std} & AE $x$-consistency \\
    $\mathcal{L}_R$ & \S\ref{sec:ae:loss:std} & AE rotation prediction \\
    $\mathcal{L}_P$ & \S\ref{sec:ae:loss:std}/\ref{sec:ae:loss:ftl} & AE shape prediction \\
    $\mathfrak{L}_\text{VAE}$ & \S\ref{sec:vae:loss} & GDVAE total loss \\
    $\mathcal{L}_\mathrm{HF}$ & \S\ref{sec:vae:loss:hfvae} & HFVAE loss \\
    ${L}_R$ & \S\ref{sec:vae:loss:hfvae} & VAE reconstruction loss \\
    $\mathcal{L}_\lambda$ & \S\ref{sec:flowlikeloss} & Spectral log-likelihood loss \\
    $\mathcal{L}_D$ & \S\ref{sec:additionaldisentloss} & Additional disentanglement loss \\
    $\mathcal{L}_F$ & \S\ref{sec:spectralconsis} & Intrinsics-Spectrum consistency \\
\end{tabular}
\end{center}

\section{Invariant FTL-based Mapping}

\label{note:invarify}

As an aside, in an FTL-based model, 
we remark that it is possible to transform ${x} \in \mathcal{X}$, 
in a way that is invariant to latent-space rotation operators. 
Let $\mathcal{I}[x] = (U(x)_i^T U(x)_j)_{i,j\in[1,N_s]; i \leq j}$ 
be the collection of inner products of the subvectors of $x$. 
Then $\mathcal{I}[x]$ is rotation invariant; i.e., $\mathcal{I}[x] = \mathcal{I}[ F(R,x) ]$, 
for any $R\in SO(3)$. This idea is noted by \citet{worrall2017interpretable}.

However, we found that using $\mathcal{I}[x]$ only slightly improved rotation invariance, 
yet slightly decreased reconstruction performance, and further was computationally expensive, 
due to the quadratic dependence of $\dim(\mathcal{I}[x])$ on $N_s$. 
Nevertheless, this may be specific to our particular architectural setup, 
and could still be an interesting direction for future work.

\section{Evaluation Metric Details}

\subsection{Latent Rotation Invariance Measure}

\label{app:CX}

We provide a more detailed description of the latent clustering metric $\mathcal{C}_X$ here.
Recall that our goal is to take a set of random shapes (potentially differing in intrinsics, non-rigid extrinsics, and orientation) and duplicate each shape, before randomly rotating the copies.
We then encode each shape into $x_c$-space and cluster them.
We use K-means clustering \citep{macqueen1967some} to obtain the labeling.
We expect our canonicalization method to bring the 
latent representations close in latent space, such that rotated copies should cluster together.
We can therefore measure rotational invariance 
by supervised clustering quality metrics, in which the instance identity (i.e., which shape a vector originated from) is a ground truth cluster label.
We use Adjusted Mutual Information (AMI) for this \citep{vinh2010information}, 
which returns 1 for a perfect partitioning (as compared to the ground truth) and a 0 for a random clustering.

While this captures the representational invariance to rotation in the embedding space,
the number of disparate sample shapes to use is unclear.
We therefore average over a sequentially larger set of samples,
thus giving an ``area-under-the-curve''-like measure of quality across sample sizes.

More formally, let $\Gamma = \{P_1, \ldots, P_{N_I}\}$ be a set of $|\Gamma|=N_I$ PC shape instances.
Now consider a set that includes $N_{rc}$ rotated copies of each PC:
$ \tilde{\Gamma} = \bigcup_{k=1}^{N_I} \{ \widetilde{P}_{k,1}, \ldots, \widetilde{P}_{k,N_{rc}} \} $,
where $\widetilde{P}_{k,j} = P_k \widetilde{R}_j$ for a randomly sampled $\widetilde{R}_j$
and $|\tilde{\Gamma}| = N_I N_{rc}$.
We then encode the set into canonical representations
$\tilde{\Gamma}_E = \{ E_x(\tilde{P}_{k,j})\;\forall\;\tilde{P}_{k,j} \in \tilde{\Gamma} \}$
and run our clustering algorithm on $\tilde{\Gamma}_E$ to get $\text{AMI}(N_I)$ 
for a given instance set size $N_I$.
Let $\mathcal{N}_S$ be a set of sample sizes (we chose eight sizes, linearly spaced from 20 to $10^3$).
Finally, the $x_c$-space rotational consistency metric is given by
\begin{equation}
	\mathcal{C}_X = \frac{1}{|\mathcal{N}_S|} \sum_{N_I \in \mathcal{N}_S} \text{AMI}(N_I).
\end{equation}
Note that for each size we always run two clusterings with different randomly chosen sample shapes, 
and use their average AMI in the above equation.
For implementations, we use scikit-learn \citep{scikit-learn}.

\section{Dataset Details}
\label{app:datasets}

Except for HA, all our datasets are identical to those in \citep{aumentado2019geometric}.
We denote $N_p$ as the size of the input point cloud (PC) and $N_\lambda$ as the dimensionality of the spectrum used.
In all cases, we output the same number of points as we input.

We also perform a scalar rescaling of the dataset such that 
	the largest bounding box length is scaled down to unit length.
This scale is the same across PCs in a given dataset
	(otherwise the change in scale would affect the spectrum for each shape differently).
For augmentation and rigid orientation learning, 
    we apply random rotations about the gravity axis (SMAL and SMPL) or the out-of-image axis (MNIST).
For rotation supervision, 
    the orientation of the raw data is treated as canonical.

\subsection{MNIST}

Meshes are extracted from the greyscale MNIST images, followed by area-weighted point cloud (PC) sampling.
See \citep{aumentado2019geometric} for extraction details.
We set $N_p = 512$ and $N_\lambda = 20$.
The dataset has 59483 training examples and 9914 testing examples.

\subsection{SMAL}

Using the SMAL model \citep{Zuffi:CVPR:2017}, we generate set of 3D animal shapes with varying shape and pose. 
Using densities provided by authors, we generate 3200 shapes per animal category,
Following 3D-CODED \citep{groueix20183d}, we sample poses by taking a Gaussian about the joint angles with a standard deviation of 0.2.
We use 15000 shapes for training and 1000 for testing, and set $N_p = 1600$ and $N_\lambda = 24$.

\subsection{SMPL}

Based on the SMPL model \citep{SMPL:2015}, 
	we again follow the procedure in 3D-CODED \citep{groueix20183d} to assemble a dataset of human models.
This results in 20500 meshes per gender, using random samples from the SURREAL dataset \citep{varol17_surreal}, 
	plus an additional 3100 meshes of ``bent'' people per gender, following \citet{groueix20183d}.
Ultimately, we get 45992 training and 1199 testing meshes, equally divided by gender, after spectral calculations.
We used $N_p = 1600$ and $N_\lambda = 20$.

\subsection{Human-Animal (HA)}

Since our model uses only geometry, we are able to simply mix SMAL and SMPL data together.
The testing sets are left alone, and used separately during evaluation (for comparison to the unmixed models).
For training, we use the entire SMAL training set, 
	plus 9000 unbent and 1500 bent samples from the SMPL training set, per gender.
We set $N_p = 1600$ and $N_\lambda = 22$.

Note that the use of a single scalar scaling factor (setting the maximum bounding box length to 1) means that
SMPL models are smaller in the HA data than in the isolated SMPL dataset. 
We correct for this in the evaluation tables so that they are comparable (e.g., for Chamfer distances).

\begin{table*}
	\centering
	\caption{ 
	VAE Evaluation on held-out test data. 
 	The first three columns denote the dataset, the VAE type, and the AE type 
 		(S and U mean supervised and unsupervised, respectively).
 	GDVAE-FO means using the $x_c$-derived latent intrinsics 
 		(as opposed to the $\lambda$-derived one) in training; 
 		PCLBO and NCNJ denote using point cloud-based LBOs
 		 and ablating the $\mathcal{L}_D$ loss, respectively.
 	The five right-most columns are evaluation metrics, 
 		considering reconstruction ($d_C$),
 		generative modelling (Fidelity, Coverage, and $\log P_\lambda(\lambda)$),
 		and unsupervised disentanglement ($\mathcal{S}$).
	See \S\ref{sec:ablations} and \ref{sec:srobust} for additional details on the alterations (PCLBO and NCNJ), 
	\S\ref{sec:trainregimes} for the difference between GDVAE{\PPP} and GDVAE-FO,
	\S\ref{sec:res:genmodel} for the generative modelling metrics,
	and \S\ref{sec:ret} for the scalar disentanglement measure.
	For the HA dataset, we show $A/B$ as the scores on the SMAL and SMPL test sets, respectively,
	and
	we also scale Chamfer distances for the HA dataset to make them comparable across datasets, as in Table \ref{tab:ae}. 
	We see that the GDVAE{\PPP}\ model obtains much better disentanglement scores than the GDVAE model (across SMAL, SMPL, and HA), 
		while GDVAE-FO does significantly worse. 
	In terms of retrieval quality,
		using the PCLBO degrades performance,
		but it stays above the GDVAE as well,
		while
		ablating $\mathcal{L}_D$ (the NCNJ scenario) worsens performance on SMPL 
			but has little effect for SMAL.}
	\label{tab:vaemass}       %
	\begin{tabular}{ccc|ccccc}
		\hline\noalign{\smallskip}
		Dataset & VAE Model & AE Model & $d_C(P,\widehat{P}) \downarrow$ & Fidelity $\downarrow$ & Coverage $\uparrow$ & $\log P_\lambda(\lambda) \uparrow$ & $\mathcal{S} \uparrow$ \\
		\noalign{\smallskip}\hline\noalign{\smallskip}
		\multirow{12}{*}{SMAL} &
		\multirow{4}{*}{GDVAE\PP} &
					  STD-S & 0.56 & 0.83 & 0.77 & -121.60 & 2.07 \\
		$\,$ & $\,$ & FTL-S & 0.40 & 1.20 & 0.62 & -144.17 & 2.12 \\
		$\,$ & $\,$ & STD-U & 0.39 & 0.68 & 0.66 & -114.73 & 2.02 \\ 
		$\,$ & $\,$ & FTL-U & 0.73 & 0.99 & 0.75 & -135.21 & 2.07 \\ 
		\noalign{\smallskip}\cline{2-8} \noalign{\smallskip}
		$\,$ & \multirow{2}{*}{GDVAE-FO} &
					  STD-S &  0.48 & 0.81 & 0.72 & -147.60 & 0.49  \\
		$\,$ & $\,$ & FTL-S &  0.24 & 0.90 & 0.73 & -143.39 & 0.43  \\
		\noalign{\smallskip}\cline{2-8} \noalign{\smallskip}
		$\,$ & GDVAE\PP & STD-S & 0.51 & 0.89 & 0.65 & -116.62 & 1.83 \\
		$\,$ & (PCLBO)  & FTL-S & 1.05 & 1.09 & 0.65 & -111.12 & 1.11 \\
		\noalign{\smallskip}\cline{2-8} \noalign{\smallskip}
		$\,$ & GDVAE\PP & STD-S & 0.44 & 1.20 & 0.47 & -151.32 & 2.14  \\
		$\,$ & (NCNJ)   & FTL-S & 0.34 & 1.09 & 0.57 & -196.15 & 2.06 \\
		\noalign{\smallskip}\hline\noalign{\smallskip}
		\multirow{12}{*}{SMPL} &
		\multirow{4}{*}{GDVAE\PP} &
					  STD-S &  0.50 & 1.21 & 0.73 & -131.82 & 2.58 \\
		$\,$ & $\,$ & FTL-S &  0.38 & 1.38 & 0.80 & -137.06 & 2.46 \\
		$\,$ & $\,$ & STD-U &  0.36 & 1.11 & 0.67 & -151.74 & 2.46 \\ 
		$\,$ & $\,$ & FTL-U &  0.43 & 1.38 & 0.80 & -135.45 & 1.93 \\ 
		\noalign{\smallskip}\cline{2-8} \noalign{\smallskip}
		$\,$ & \multirow{2}{*}{GDVAE-FO} &
					  STD-S & 0.43 & 1.21 & 0.65 & -232.73 & 0.32  \\
		$\,$ & $\,$ & FTL-S & 0.26 & 1.33 & 0.79 & -94.65  & 0.64 \\
		\noalign{\smallskip}\cline{2-8} \noalign{\smallskip}
		$\,$ & GDVAE\PP & STD-S & 0.53 & 1.16 & 0.80 & -164.64 & 1.43 \\
		$\,$ & (PCLBO) & FTL-S  & 0.37 & 1.47 & 0.77 & -145.43 & 1.35 \\
		\noalign{\smallskip}\cline{2-8} \noalign{\smallskip}
		$\,$ & GDVAE\PP & STD-S & 0.43 & 1.36 & 0.54 & -178.26 & 2.01 \\
		$\,$ & (NCNJ) & FTL-S & 0.36 & 1.44 & 0.58 & -123.40 & 2.24 \\
		\noalign{\smallskip}\hline\noalign{\smallskip}
		\multirow{2}{*}{HA} &
		\multirow{2}{*}{GDVAE\PP} &
					  STD-S & 0.54/0.64 & 0.89/1.32 & 0.63/0.68 & -130.13/-125.44 & 1.86/1.92 \\
		$\,$ & $\,$ & FTL-S & 0.40/0.65 & 1.06/1.54 & 0.74/0.59 & -117.82/-108.69 & 1.72/1.83 \\ 
		\noalign{\smallskip}\hline
	\end{tabular}
\end{table*}

\begin{table}
	\setlength{\tabcolsep}{4.97pt}
	\centering
	\caption{Retrieval scores. 
		All models use the GDVAE\PPP\ training regime.
		$\uparrow (\downarrow)$ means the higher (lower) the better.
		See \S\ref{sec:ret} for additional details.
	}
	\label{tab:ret}       %
	\begin{tabular}{ccc|cccc}
		\hline\noalign{\smallskip}
		Data & Model & $\hat{s}_\psi$ & $z \uparrow$ & $z_E$ & $z_I$ & $\tilde{z}_I$\\
		\noalign{\smallskip}\hline\noalign{\smallskip}
		\multirow{21}{*}{SMAL} &
		\multirow{2}{*}{STD-S} & 
					  $\widehat{s}_\beta$  & 1.46 & 0.23 $\downarrow$ & 1.63 $\uparrow$   & 2.04 $\uparrow$  \\
		$\,$ & $\,$ & $\widehat{s}_\theta$ & 0.69 & 0.83 $\uparrow$   & 0.16 $\downarrow$ & 0.08 $\downarrow$   \\
		\noalign{\smallskip}\cline{2-7} \noalign{\smallskip}
		$\,$ & \multirow{2}{*}{FTL-S} & 
					  $\widehat{s}_\beta$  & 1.52 & 0.13 $\downarrow$ & 1.66 $\uparrow$ & 1.77 $\uparrow$ \\
		$\,$ & $\,$ & $\widehat{s}_\theta$ & 0.61 & 0.68 $\uparrow$ & 0.11 $\downarrow$ & 0.12 $\downarrow$ \\
				\noalign{\smallskip}\cline{2-7} \noalign{\smallskip}
		$\,$ & \multirow{2}{*}{STD-U} & 
					  $\widehat{s}_\beta$   & 1.35 & 0.19 $\downarrow$ & 1.56 $\uparrow$ & 1.90 $\uparrow$  \\
		$\,$ & $\,$ & $\widehat{s}_\theta$  & 0.73 & 0.79 $\uparrow$ & 0.15 $\downarrow$ & 0.06 $\downarrow$  \\ 
			\noalign{\smallskip}\cline{2-7} \noalign{\smallskip}
		$\,$ & \multirow{2}{*}{FTL-U} & 
					  $\widehat{s}_\beta$  &  1.54 & 0.15 $\downarrow$ & 1.77 $\uparrow$ & 1.99 $\uparrow$   \\
		$\,$ & $\,$ & $\widehat{s}_\theta$ &  0.63 & 0.66 $\uparrow$ & 0.21 $\downarrow$ & 0.18 $\downarrow$  \\
				\noalign{\smallskip}\cline{2-7} \noalign{\smallskip}
		$\,$ & STD-S   & $\widehat{s}_\beta$  &  1.15 & 0.16 $\downarrow$ & 1.30 $\uparrow$ & 2.00 $\uparrow$ \\
		$\,$ & (PCLBO) & $\widehat{s}_\theta$ &  0.74 & 0.89 $\uparrow$ & 0.20 $\downarrow$ & 0.04 $\downarrow$ \\
				\noalign{\smallskip}\cline{2-7} \noalign{\smallskip}
		$\,$ & FTL-S & $\widehat{s}_\beta$    & 0.74 & 0.25 $\downarrow$ & 0.96 $\uparrow$   & 1.84 $\uparrow$  \\
		$\,$ & (PCLBO) & $\widehat{s}_\theta$ & 0.56 & 0.62 $\uparrow$   & 0.21 $\downarrow$ & 0.09 $\downarrow$   \\
				\noalign{\smallskip}\cline{2-7} \noalign{\smallskip}
		$\,$ & STD-S & $\widehat{s}_\beta$   & 1.38 & 0.20 $\downarrow$ & 1.60 $\uparrow$ & 1.93 $\uparrow$  \\
		$\,$ & (NCNJ) & $\widehat{s}_\theta$ & 0.73 & 0.91 $\uparrow$ & 0.17 $\downarrow$ & 0.07 $\downarrow$   \\
				\noalign{\smallskip}\cline{2-7} \noalign{\smallskip}
		$\,$ & FTL-S & $\widehat{s}_\beta$ &  1.55 & 0.19 $\downarrow$ & 1.69 $\uparrow$ & 1.91 $\uparrow$  \\
		$\,$ & (NCNJ) & $\widehat{s}_\theta$ &  0.73 & 0.70 $\uparrow$ & 0.13 $\downarrow$ & 0.07 $\downarrow$ \\
		\noalign{\smallskip}\hline\noalign{\smallskip}
		\multirow{21}{*}{SMPL} &
		\multirow{2}{*}{STD-S} & 
						 $\widehat{s}_\beta$  & 0.72  & 0.12 $\downarrow$ & 1.97 $\uparrow$   & 1.73 $\uparrow$   \\
		$\,$ & $\,$ &    $\widehat{s}_\theta$ & 0.94  & 0.93 $\uparrow$   & 0.20 $\downarrow$ & 0.34 $\downarrow$ \\
				\noalign{\smallskip}\cline{2-7} \noalign{\smallskip}
		$\,$ & \multirow{2}{*}{FTL-S} & $\widehat{s}_\beta$  & 0.85 & 0.12 $\downarrow$ & 1.78 $\uparrow$   & 2.11 $\uparrow$  \\
		&	  						  & $\widehat{s}_\theta$ & 0.90 & 0.95 $\uparrow$   & 0.15 $\downarrow$ & 0.25 $\downarrow$ \\
				\noalign{\smallskip}\cline{2-7} \noalign{\smallskip}
		$\,$ & \multirow{2}{*}{STD-U} & $\widehat{s}_\beta$  & 0.65 & 0.06 $\downarrow$ & 1.95 $\uparrow$   & 1.80 $\uparrow$ \\
						  $\,$ & $\,$ & $\widehat{s}_\theta$ & 0.90 & 0.87 $\uparrow$   & 0.30 $\downarrow$ & 0.29 $\downarrow$  \\ 
						  		\noalign{\smallskip}\cline{2-7} \noalign{\smallskip}
		$\,$ & \multirow{2}{*}{FTL-U} & $\widehat{s}_\beta$  & 0.97 & 0.30 $\downarrow$ & 1.58 $\uparrow$   & 2.03 $\uparrow$  \\
		   $\,$ & $\,$ & $\widehat{s}_\theta$ & 0.87 & 0.89 $\uparrow$   & 0.24 $\downarrow$ & 0.28 $\downarrow$   \\
		   		\noalign{\smallskip}\cline{2-7} \noalign{\smallskip}
		$\,$ & STD-S   & $\widehat{s}_\beta$  & 0.42 & 0.23 $\downarrow$ & 0.90 $\uparrow$   & 1.74 $\uparrow$   \\
		$\,$ & (PCLBO) & $\widehat{s}_\theta$ & 0.93 & 0.91 $\uparrow$   & 0.14 $\downarrow$ & 0.31 $\downarrow$ \\
				\noalign{\smallskip}\cline{2-7} \noalign{\smallskip}
		$\,$ & FTL-S   & $\widehat{s}_\beta$  & 0.71 & 0.43 $\downarrow$ & 1.09 $\uparrow$   & 2.01 $\uparrow$   \\
		$\,$ & (PCLBO) & $\widehat{s}_\theta$ & 0.91 & 0.91 $\uparrow$   & 0.21 $\downarrow$ & 0.30 $\downarrow$ \\
				\noalign{\smallskip}\cline{2-7} \noalign{\smallskip}
		$\,$ & STD-S   & $\widehat{s}_\beta$  &  1.26 & 0.24 $\downarrow$ & 1.63 $\uparrow$   & 2.23 $\uparrow$ \\
		$\,$ & (NCNJ)  & $\widehat{s}_\theta$ &  0.74 & 0.96 $\uparrow$   & 0.34 $\downarrow$ & 0.38 $\downarrow$  \\
				\noalign{\smallskip}\cline{2-7} \noalign{\smallskip}
		$\,$ & FTL-S   & $\widehat{s}_\beta$  &  1.35 & 0.22 $\downarrow$ & 1.66 $\uparrow$   & 2.24 $\uparrow$   \\
		$\,$ & (NCNJ)  & $\widehat{s}_\theta$ &  0.77 & 0.96 $\uparrow$   & 0.15 $\downarrow$ & 0.32 $\downarrow$  \\
		\noalign{\smallskip}\hline
	\end{tabular}
\end{table}

\section{Results Tables}

\begin{figure*}[t]
    \centering
    \includegraphics[width=0.244\textwidth]{fig-table/Intrinsic-Scores-SMAL-FTL.pdf}
    \includegraphics[width=0.244\textwidth]{fig-table/Extrinsic-Scores-SMAL-FTL.pdf}
    \includegraphics[width=0.244\textwidth]{fig-table/Intrinsic-Scores-SMPL-FTL.pdf}
    \includegraphics[width=0.244\textwidth]{fig-table/Extrinsic-Scores-SMPL-FTL.pdf}
    \includegraphics[width=0.244\textwidth]{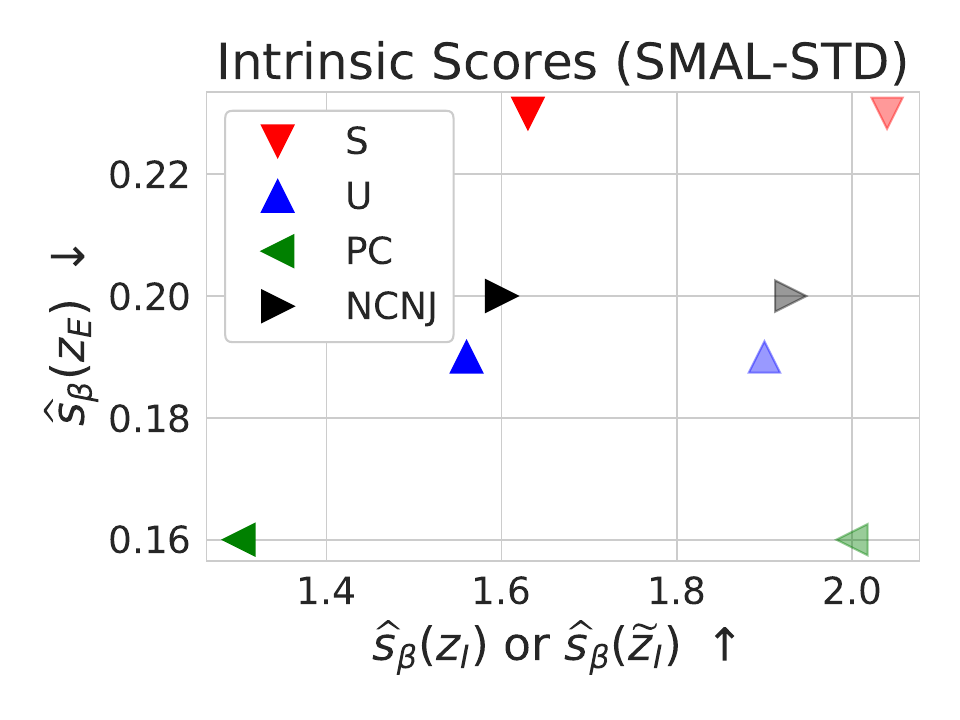}
    \includegraphics[width=0.244\textwidth]{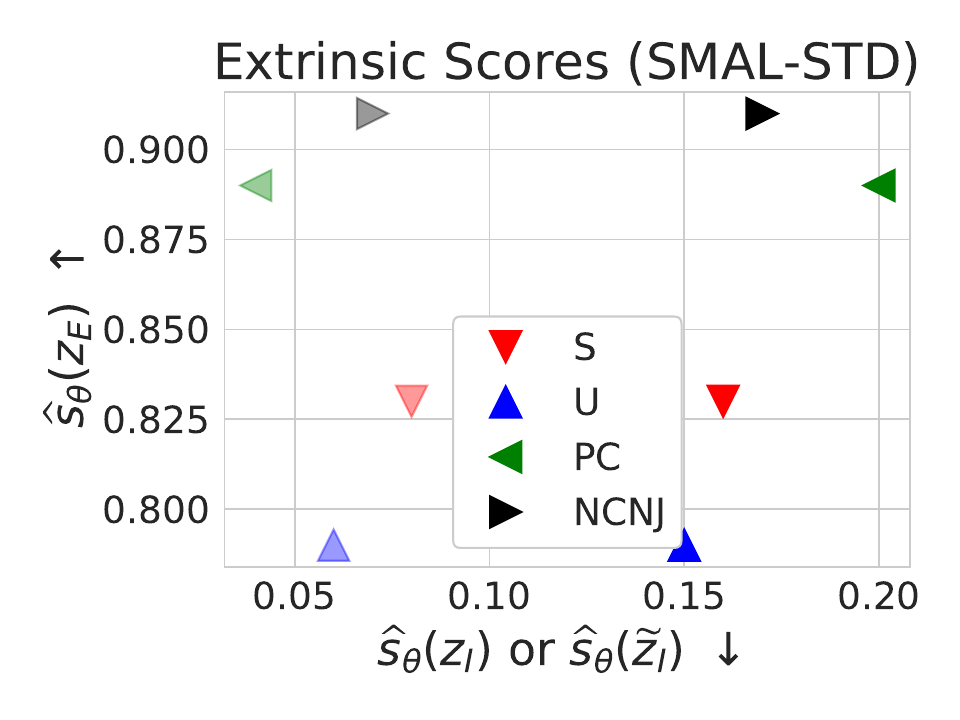}
    \includegraphics[width=0.244\textwidth]{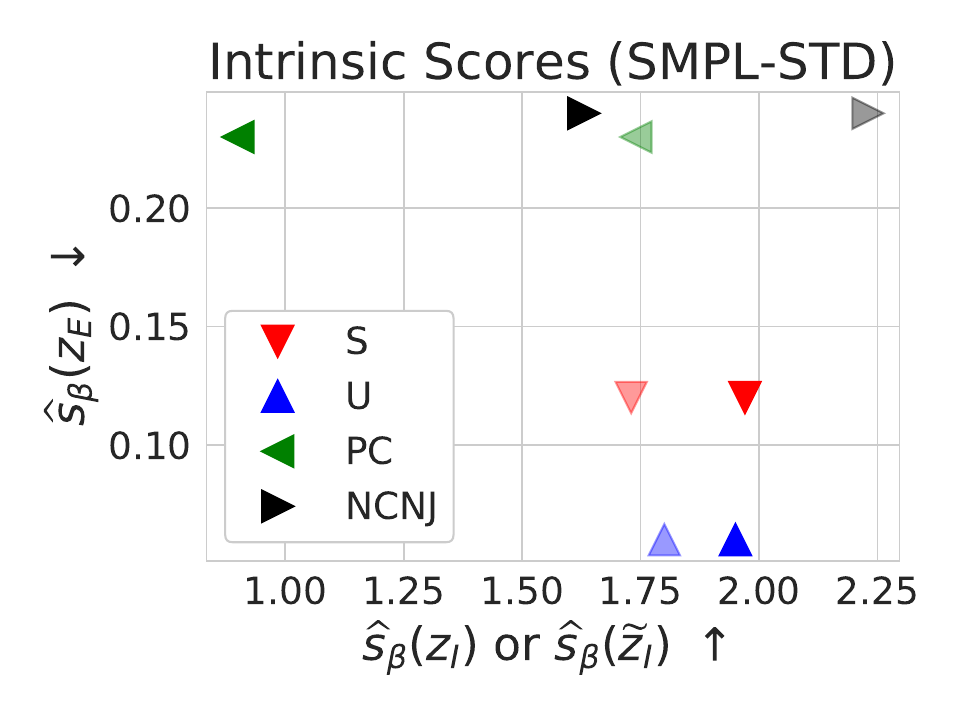}
    \includegraphics[width=0.244\textwidth]{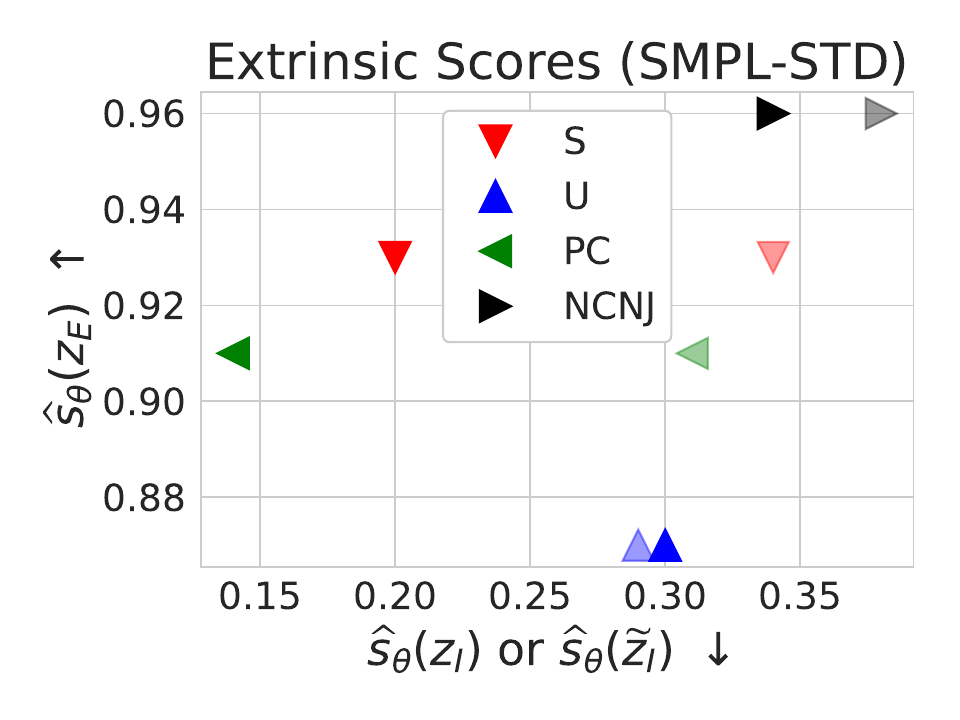}
    \caption{
        \new{
            Pose-aware retrieval scores with the STD AE model.
            Model notation refers to the GDVAE\PP\ model with (S) or without (U) rotation supervision, use of the PC-derived LBOS (PC; see \S\ref{sec:srobust}), and the partial disentanglement loss ablation (NCNJ; see \S\ref{sec:ablations}).
            The lighter (partially transparent) counterparts of each point corresponds to using $\widetilde{z}_I$ instead of $z_I$ for retrieval.
            We reproduce the FTL AE plots 
            (from Fig.\ \ref{table:t2f-retftl})
            to aid in comparison.
            See also Appendix Table \ref{tab:ret} for detailed values.
            Compared to the FTL case, for SMPL,
                U performs relatively better on intrinsic scores, 
                while S and U are relatively similar for extrinsic scores.
            For SMAL, we see that the extrinsic scores are generally better with the STD AE, compared to the FTL one.
            We also see that, in the STD case, %
                the PCLBO scenario performs relatively better on SMAL than its S/U counterparts.
            Finally, we note that using the spectrum-derived latents $\widetilde{z}_I$ are generally better, but not always (e.g., on SMPL-STD).
        }
    }
    \label{table:t2f-retstd}
\end{figure*}

\new{
In this section, we provide the detailed results tables for the experiments discussed in \S\ref{sec:vaeresults}. 
See Table \ref{tab:vaemass} for measurements of VAE quality, including reconstruction, generative modelling, and disentanglement metrics.
See Table \ref{tab:ret} for pose-aware retrieval scores, with various choices of latent vector, and Fig.\ \ref{table:t2f-retstd} for plots of those scores for the STD AE (as well as Fig.\ \ref{table:t2f-retftl} for the FTL AE case).
}

\section{Implementation Details}

All models were implemented in Pytorch \citep{pytorch}.
Notationally, let 
	$n = \text{dim}(x_c)$, 
	$n_E = \text{dim}(z_E)$,
	$n_I = \text{dim}(z_I)$,
	and
	$n_R = \text{dim}(z_R)$.
For this section, 
	we assume that the number of points $N_p$ in a point cloud (PC)
	is the same for inputs and outputs 
	(though the architectures themselves do not require this).
Validation sets of size 40 (or 250 for MNIST) were set aside 
	from the training set to observe generalization error estimates.
Hyper-parameters were largely set based on qualitative examination 
	of training outputs.

\subsection{Autoencoder Details}
\label{app:aearch}

\subsubsection{AE Network Architectures}

Both the STD and FTL architectures used the same network components,
with slight hyper-parameter alterations.
Our encoders 
$E_r : \mathbb{R}^{3N_p} \rightarrow \mathbb{R}^{4}$ and 
$E_x : \mathbb{R}^{3N_p} \rightarrow \mathbb{R}^{n}$ 
were implemented as PointNets \citep{qi2017pointnet} without spatial transformers, with hidden channel sizes 
\verb*|(64,128,256,512,128)| and \verb|(128,256,512,836,1024)|.
The inputs are only the point coordinates (i.e., three channels) 
and 
the output is a four-dimensional quaternion 
for $E_r$ and an $n$-dimensional vector for $E_x$.
The decoder $D : \mathbb{R}^n \rightarrow \mathbb{R}^{3N_p}$ is implemented as a fully connected network,
with hidden layer sizes \verb|(K,2K,4K)|, 
	where $K = 1200$ for STD and $K = 1250$ for FTL.
Within $D$, each layer consisted of a linear layer, layer normalization \citep{ba2016layer}, and ReLU (except for the last, which had only a linear layer).
For the MNIST dataset only,
	we changed the hidden layer sizes of  the decoder $D$ to be
	\verb|(512,1024,1536)|.

\subsubsection{AE Hyper-Parameters and Loss Weights}

For architectural parameters, 
	in the FTL case, we set $N_s = 333$, and hence $n = 999$.
For STD, we let $n = 600$ and did not notice improvements when increasing it.
For MNIST, we set $N_s = 32$ and, for the STD case, $n = 150$.
Regarding loss parameters,
	we set the reconstruction loss weights to 
	$\alpha_C = 200$,
	$\alpha_H = 1$,
	$\widetilde{\gamma}_P = 100$,
	and
	$\gamma_P = 20$, in the FTL case,
	altering only $\gamma_P = 250$ in the STD case.
Rotational consistency and prediction loss weights were set to
$\gamma_c = 1$
and
$\gamma_r = 10$.
Regularization loss weights were $\gamma_w = \gamma_d = 2\times 10^{-5}$.
For MNIST, we altered $\gamma_c = 50$ for STD, 
	while we let $\gamma_d = 10^{-6}$, 
				 $\gamma_w = 5\times 10^{-5}$, and 
				 $\gamma_c = 100$ for FTL.

\subsubsection{AE Training Details}

We train all AEs with Adam \citep{kingma2014adam}, 
	using an initial learning rate of 0.0005.
For supervised AEs, we pretrain the rotation predictor for 2000 iterations before the rest of the network.
We use a scheduler that decreases the learning rate by 5\% upon hitting a loss plateau, until it reaches 0.0001.
We trained MNIST, SMAL, SMPL, and HA for 200, 1250, 350, and 400 epochs, 
	respectively, and batch sizes of 64/100 (FTL/STD) and 36/40 for MNIST and non-MNIST datasets.
We set the number of rotated copies 
    (which expands the batch sizes above)
    to $N_R = 3$, 
    except in the case of MNIST 
    (for which we used 
    $N_R = 6$ in the FTL case
    and
    $N_R = 4$ in the STD case).
Finally, note that, during training, 
    for the supervised case only, 
    we replace the predicted rotation $\widehat{R}$ 
    with the real one $R$ in all operations.

\subsection{Variational Autoencoder Details}
\label{app:vaearch}

For implementation of the HFVAE, we use ProbTorch \citep{siddharth2017learning}.
Our normalizing flow subnetwork used nflows \citep{nflows}.

\subsubsection{VAE Network Architectures}

For the VAE, 
	all networks except for the flow mapping $f_\lambda$ 
	are implemented as fully connected networks
	(linear-layernorm-ReLU, as above).
Approximate variational posteriors have diagonal covariances.
Thus, we have the following mappings with their hidden sizes:
\begin{itemize}
	\item 
	The rotation distribution parameter encoders, 
	$\mu_R : \mathbb{R}^4 \rightarrow \mathbb{R}^{n_R}$ 
	and 
	$\Sigma_R : \mathbb{R}^4 \rightarrow \mathbb{R}^{n_R}$,	
	are implemented with an initial shared network,
	with hidden sizes \verb|(256, 128)| 
	into an intermediate dimensionality of 64,
	followed by single linear layer each.
	
	\item  
	The quaternion decoder 
	$D_q : \mathbb{R}^{n_R} \rightarrow \mathbb{R}^4$ 
	is structured as \verb|(64, 128, 256)|. 
	
	\item
	The intrinsic and extrinsic parameter encoders,
	$\mu_\xi : \mathbb{R}^n \rightarrow \mathbb{R}^{n_\xi}$ 
	and 
	$\Sigma_\xi : \mathbb{R}^n \rightarrow \mathbb{R}^{n_\xi}$,	
	for $\xi \in \{E, I\}$,
	have identical network architectures across latent group types:
		\verb*|(2000,1600,1200,400)|
		and
		\verb*|(2000,1200,400)|,
		for $\mu_\xi$ and $\Sigma_\xi$, respectively.

	\item 
	The only mapping that is not a fully connected network is the bijective flow $f_\lambda$ (and its inverse, $g_\lambda$).
	Recall that we use $f_\lambda$ as $\widetilde{\mu}$.
	Hence, 
	$f_\lambda : \mathbb{R}^{N_\lambda} \rightarrow \mathbb{R}^{n_I} $ 
	and $N_\lambda = n_I$.
	This is implemented as a normalizing flow with nine layers,
		where each layer consists of 
		an affine coupling transform \citep{dinh2016density},
		an activation normalization (actnorm) \citep{kingma2018glow},
		and a random feature ordering permutation. 
	The last layer does not have normalization or permutation.
	Each affine coupling uses an internal FC network 
		with one hidden layer of size 400.
	
	\item
	For the GDVAE\PP \  training regime, 
	we require a covariance parameter estimator 
	for inference during training:
	$\Sigma_\xi : \mathbb{R}^{N_\lambda} \rightarrow \mathbb{R}^{n_I}$.
	This is implemented via hidden layers \verb|(2n_I, 2n_I)|.
	
	\item 
	The shape decoder 
	$D_x : \mathbb{R}^{n_I + n_E} \rightarrow \mathbb{R}^{n} $
	is an FC network with hidden layers
	\verb|(600,1200,1600,2000)|.
	
\end{itemize}

\subsubsection{VAE Hyper-Parameters and Loss Weights}
\label{app:vaehyploss}

Recall that the loss hyper-parameters control the following terms:
the 
intra-group total correlation (TC) $\beta_1$,
dimension-wise KL divergence $\beta_2$,
mutual information $\beta_3$,
inter-group TC $\beta_4$,
log-likelihood reconstruction $\omega_R$,
relative quaternion reconstruction $\omega_q$,
flow likelihood $\omega_p$,
intrinsics consistency in $z_I$-space $\omega_I$,
intrinsics consistency in $\lambda$-space $\omega_\lambda$,
covariance disentanglement $\omega_\Sigma$,
and
Jacobian disentanglement $\omega_\mathcal{J}$.

\begin{table}
	\centering
	\setlength{\tabcolsep}{5pt}
	\caption{VAE hyper-parameters across datasets. These values are for the {GDVAE\PPP} \  training method. See text in Appendix \S\ref{app:vaehyploss} for details. }
	\label{tab:vaeparams} %
	\begin{tabular}{c|cccccccc}
		\hline\noalign{\smallskip}
		$\,$ & $\beta_2$  & $\beta_4$ & $\omega_R$ &  $\omega_\Sigma$
			 & $\omega_I$ & $\omega_\lambda$ & $n_E$ & $n_I$ \\
		\noalign{\smallskip}\hline\noalign{\smallskip}
		MNIST & 50 & 100 & 350 & 80 & 600 & 0   & 4  & 24 \\
		SMAL  & 10 & 50  & 400 & 40 & 300 & 200 & 8  & 24 \\
		SMPL  & 20 & 100 & 350 & 80 & 600 & 0   & 12 & 20 \\
		HA    & 20 & 80  & 360 & 60 & 450 & 100 & 10 & 22 \\
		\noalign{\smallskip}\hline
	\end{tabular}
\end{table}

In all cases, we set 
$n_R = 3$, $\beta_1 = 1$, $\beta_3 = 1$, $\omega_p = 1$, $\omega_\mathcal{J} = 200$, and $\omega_q = 10$. 
See Table \ref{tab:vaeparams} for dataset-dependent parameters.
An additional $L_2$ weight decay was applied to all networks, 
	with a strength of $10^{-4}$.
For the flow-only (GDVAE-FO) approach, 
	the parameters are the same per dataset, 
	except for $\omega_I$ and $\omega_\lambda$ 
	(which were tuned more in line with original GDVAE model, 
	in an effort to improve disentanglement).
For the FO case, we set $\omega_\lambda = 800$ 
	and $\omega_I$ to 0, 250, and 0,
	for SMAL, SMPL and HA, respectively.

\subsubsection{VAE Training Details}

As in the AE case, optimization is done with Adam, 
	using a reduce-on-plateau scheduler.
The initial learning rate was set to 0.0001,
	with a minimum of 0.00001.
A batch size of 264 was used, except for MNIST, for which we used 512.
The networks were trained for 25000 iterations for MNIST and 40000 iterations for all other datasets.
We note that for the GDVAE\PP \  mode only, 
we also cut the gradient of the $\omega_I$ loss term 
from flowing through $\widetilde{\mu}_I$  
(preventing extrinsic information in $\mu_I$ from contaminating $\widetilde{\mu}_I$).

\section{Full Rotation-Space Experiments}
\label{app:fullrot}

\begin{table}
	\centering
\caption{
\new{
AE evaluation on held-out test data with full 3D rotations. 
	     Metrics (left to right) refer to the Chamfer distance in reconstructions and the rotational consistency measures 
	     (in 3D and $x_c$-space, respectively); see \S\ref{sec:aeresults}. 
	     We place the scores obtained by the corresponding single-axis models in square brackets beside each value (from Table \ref{tab:ae}), for ease of comparison.
	     Notice the deterioration in both reconstruction and rotation invariance, compared to the single axis case.
	     Nevertheless, note that (1) $\mathcal{C}_{\text{3D}}$ is of a smaller magnitude than $d_C(P,\widehat{P})$, suggesting the presence of some rotation invariance, and (2) $\mathcal{C}_X$ are larger than zero (the expected value if there were no latent structure in the space). 
	     }
	 }
\label{app:tab:fullrot}       %
\setlength{\tabcolsep}{4pt}
\begin{tabular}{cc|ccc}
\hline\noalign{\smallskip}
Dataset & Model & $d_C(P,\widehat{P}) \downarrow$ & $\mathcal{C}_{\text{3D}} \downarrow$  & $\mathcal{C}_X \uparrow$ \\
\noalign{\smallskip}\hline\noalign{\smallskip}
\multirow{2}{*}{SMAL} &
        FTL-S & 0.46 [0.10] & 0.41 [0.14] & 0.11 [0.93] \\
$\,$ &  FTL-U & 0.29 [0.10] & 1.32 [0.21] & 0.17 [0.88] \\
\noalign{\smallskip}\hline
\end{tabular}
\end{table}

\new{
We also provide some limited tests our method on full 3D rotations, rather than single-axis rotations. We find that the invariance properties are severely reduced in this more difficult scenario.
In particular, we train two AEs on SMAL, both using the FTL architecture and allowing arbitrary rotations.
We try both the supervised (S) and unsupervised (U) cases. 

Results are shown in Table \ref{app:tab:fullrot}.
We see that both reconstruction and rotation invariance are worsened;
however, note that (1) $\mathcal{C}_{\text{3D}}$ is of a smaller magnitude than $d_C(P,\widehat{P})$ (for S), suggesting the presence of some rotation invariance, and (2) $\mathcal{C}_X$ are larger than zero (the expected value if there were no latent structure in the space).
Corroborating this latter point, in Fig.\ \ref{fig:fullrottsne}, we can qualitatively see that the tight latent clustering of rotated objects (as in Fig.\ \ref{fig:aetsne}) is no longer present, but that there is still some structure in the space, by which same-identity objects stay nearby under rotation.

\begin{figure}
    \centering
    \includegraphics[width=0.23\textwidth]{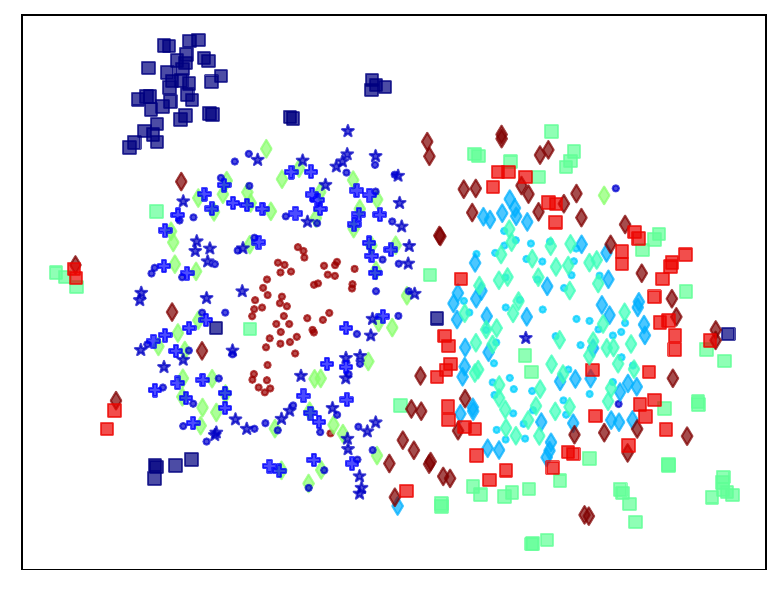}
    \includegraphics[width=0.23\textwidth]{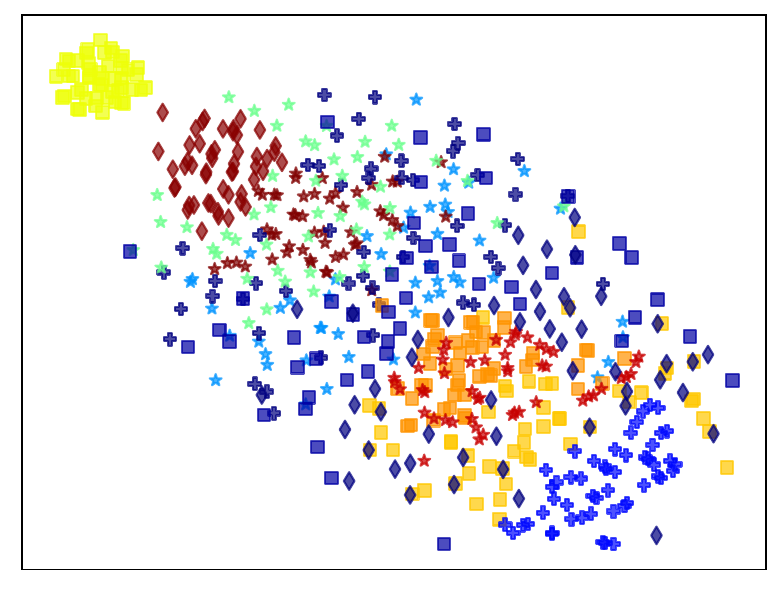}
    \caption{
    \new{
    Qualitative visualization of latent AE space with respect to full 3D rotation. Similar to Fig.\ \ref{fig:aetsne}, we show a t-SNE of the latent embeddings of random shapes under random rotations (we show more rotations as the set of rotations is now much larger). 
    Heuristically, we can see there is some clustering structure in the space, but it does not have the tight invariance of the single-axis case. 
    We show the supervised (S) case on the left and unsupervised (U) case on the right.
    See \S\ref{app:fullrot} for additional details.
    }
    }
    \label{fig:fullrottsne}
\end{figure}

We utilized slightly different hyper-parameters compared to the standard AE case.
In particular, for S, we set the batch size to $B=3$ and $N_R = 48$ trained for 24 epochs; for U, we set $B=2$ and $N_R = 60$. Decoder layers were set via $K=1400$ (see \S\ref{app:aearch}).
Loss weights were modified to
$\gamma_c = 8000$, $\widetilde{\gamma}_P = 50/7$, $\gamma_P = 10/7$ (S case)
 and
$\gamma_c = 500$, $\widetilde{\gamma}_P = 400/7$, $\gamma_P = 80/7$ (U case), to compensate for the larger rotational space.
}

\section{Mesh Experiments}

\subsection{AMASS Experiments}
\label{app:meshes:amass}

\subsubsection{AE Settings}
\newer{
Following other works (e.g., \citet{marin2021spectral,tan2018variational}),
    we use a fully connected AE for the AMASS meshes.
In particular, the encoder and decoder have hidden layers}
\verb|(1024,512)| and \verb|(512,1024)|. 
\newer{
A latent dimension of $\dim(x) = 128$ was used.
Both networks use the SELU activation \citep{klambauer2017self} and no normalization.
Note that AMASS shapes use the SMPL mesh, with $N_\text{SMPL} = 6890$ nodes.
}

\newer{
The loss utilized for training modifies only $\mathcal{L}_P$, which (in the FTL case) is given by
\begin{equation}
    \mathcal{L}_P = \gamma_P D_V(V,\widehat{V}),
\end{equation}
where 
$D_V(V_1,V_2)$ is the vertex-to-vertex mean squared error between the input nodal coordinate sets.
Other loss terms remain the same as in the PC case.
We set $\gamma_P = 5$ and use weight decay with $\gamma_w = 10^{-3}$.
For simplicity, following \citet{zhou2020unsupervised}, we include the global rotation in $z_E$ rather than using a separate latent variable.
Note that the input and output size is much larger for AMASS than for the PC case (6890 vs.\ 1600 points).
The same learning setup was used as in the PC case, 
    except we apply AdamW \citep{loshchilov2017decoupled} with a learning rate of $10^{-4}$ and batch size of 100.
We run for 250 epochs, using the same train, validation, and test splits as USPD.
Notice that, while the AE uses the identical meshing of the input for the reconstruction loss, it does not perform any disentanglement.
The VAE, which does perform disentanglement, uses only the raw $x$ values (and does not update the AE), without correspondence or label information.
}

\subsubsection{VAE Settings}
\newer{
We slightly modify the architecture of the VAE, removing batch normalization and replacing ReLU with SELU (as in the AE).
We also increase the layer sizes: the flow network is given 10 layers, the encoders that predict 
$\mu_E$ and $\mu_I$ use hidden layers }
\verb|(2400,2000,1600,800)|, and the decoder uses \verb|(2000,1800, 1600, 800)| 
\newer{
for hidden layers; other networks are unchanged.
We then use following hyper-parameters, with the SMPL (PC) settings as the default unless otherwise mentioned
$ \omega_R = 750 $, %
$ \beta_2 = 5 $, %
$ \beta_4 = 500 $, %
$ \omega_\Sigma = 5 $
$ \omega_\mathcal{J} = 50 $, %
$ \omega_I = \omega_\lambda = 1000$, %
$ n_E = 18 $, %
and
$ n_I = 9 $. %
No weight decay was used.
We trained with a batch size of 2200 for 40K iterations, starting from initial learning rate $5\times 10^{-5}$.
}

\subsubsection{Empirical Variation}
\label{app:meshvaria}
\newer{
We also compute variabilities on our mesh experiments (see \S\ref{sec:meshexps:amass}), 
to give an indication of the variability in the results for our method.
For pose transfers, we obtain a standard deviation of 8.08.
Table \ref{tab:amassvaria} shows the standard errors of the mean for the pose-aware retrieval task.
In general, the standard deviations are fairly high.
However, following USPD, the held-out data sets from AMASS are of size 10,733 for pose transfer and 11,738 for retrieval, meaning the standard \textit{error} of the mean is relatively small.
}

\begin{table}[t] %
	\setlength{\tabcolsep}{5.1pt}
	\centering
	\caption{
	\new{
	    Pose-aware retrieval \textit{standard errors of the mean} on mesh data from AMASS. 
	    Data is shown as mean plus/minus standard error 
	    (with means from Table \ref{tab:amassretrievals}).
        Note that our latent intrinsics and extrinsics nomenclature refers to the latent ``shape'' and ``pose'' (or articulation) vectors in other works. 
		See \S\ref{app:meshvaria} for additional details.
		}
	}
	\label{tab:amassvaria}    
	{
	\setlength{\tabcolsep}{3.5pt}
	\small
	\begin{tabular}{cc|c|c}
		\hline\noalign{\smallskip}
        \multicolumn{2}{c}{Retrieval with latent:} & Intrinsics & Extrinsics \\
		\noalign{\smallskip}\hline\noalign{\smallskip}
		\multirow{2}{*}{GDVAE\PP\ } 
          & $\widetilde{E}_\beta \pm \sigma(\widetilde{E}_\beta)$  
            & 0.41 $\pm$ 0.0162 & 1.36 $\pm$ 0.0272  \\
          & $\widetilde{E}_\theta \pm \sigma(\widetilde{E}_\theta)$ 
            & 1.15 $\pm$ 0.0065 & 0.80 $\pm$ 0.0066 \\ \noalign{\smallskip}\hline\noalign{\smallskip}
		GDVAE\PP\ 
          & $\widetilde{E}_\beta \pm \sigma(\widetilde{E}_\beta)$  
            & 0.50 $\pm$ 0.0179 & 1.49 $\pm$ 0.0282  \\
        (PCA)
          & $\widetilde{E}_\theta \pm \sigma(\widetilde{E}_\theta)$ 
            & 1.21 $\pm$ 0.0065 & 0.82 $\pm$ 0.0067 \\
		\noalign{\smallskip}\hline
	\end{tabular}
	}
\end{table}

\subsection{CoMA Experiments}
\label{app:meshes:coma}

\subsubsection{Dataset}
\newer{
We use the same data as in \citet{marin2021spectral}, namely 1853 training meshes with 100 faces from an unseen subject for the shape-from-spectrum recovery test set. We also use their data and dimensionality for the LBOS eigenvalues, so we set $\dim(\lambda) = 30$, and treat the meshes at full resolution (3931 vertices and 7800 faces).
}

\subsubsection{AE settings}
\newer{
Following \citet{marin2021spectral}, we use the same fully connected AE to derive the initial latent representation $x$: tanh was used as the non-linearity, no normalization was applied, and the hidden layers were given by }
\verb|(300,200,30,200)| \newer{
(with input and output in $\mathbb{R}^{3|V|}$), with $\dim(x) = 30$.
The reconstruction loss was the vertex-to-vertex MSE, with weight $\gamma_P = 5$.
We set the weight decay to $ \gamma_w = 0.01 $, the radial regularization to $ \gamma_d = 0 $, and the batch size to 16.
Since this dataset has no orientation changes, we fix our rotation prediction to be identity.
}

\subsubsection{VAE settings}
\newer{
We use the same VAE architecture as the PC experiments.
Only the hyper-parameters and training settings are altered, which we leave at the SMPL settings by default, except for the following changes
(see also \S\ref{app:vaehyploss} for details):
$ \omega_R = 250 $, %
$ \beta_2 = 5 $, %
$ \beta_4 = 250 $, %
$ \omega_\Sigma = 100 $
$ \omega_\mathcal{J} = 250 $, %
$ \omega_I = \omega_\lambda = 1000$, %
$ n_E = 1 $, %
and
$ n_I = 30 $. %
A lighter weight decay of $10^{-6}$ was used.
We trained with a batch size of 720 for 30K iterations, starting from initial learning rate $5\times 10^{-5}$.
While this setup works well for the shape-from-spectrum task 
(see Table \ref{tab:coma.shapefromspectrum}), 
and it mimics the $\dim(\lambda) = 30$ setting from \citet{marin2021spectral}, we found qualitatively that disentangled interpolations could be improved by altering these settings to $n_E = 4$, $n_I = 12$, $\omega_R = 50$, and $\beta_4 = \omega_J = 500$, which we use for Fig.\ \ref{fig:coma.interps}.
This is likely due to facial deformations not being exactly isometric; hence, using too high LBOS dimensionality (and too low $\dim(z_E)$) leads to $z_I$ capturing information we might not expect to be intrinsic (but improving shape-from-spectrum performance).
}

\end{document}